\documentclass{article}
\usepackage[nonatbib,final]{neurips_2024}
\usepackage{fancyhdr}

\usepackage[utf8]{inputenc} 
\usepackage[T1]{fontenc}
\usepackage[numbers]{natbib}
\usepackage{adjustbox}
\usepackage[hyphens]{url}
\usepackage{breakurl}    
\usepackage{hyperref}
\usepackage[most]{tcolorbox}
\usepackage[edges]{forest}
\usepackage[framemethod=tikz]{mdframed}
\usepackage{subcaption}
\usepackage{soul}
\usepackage{adjustbox}
\usepackage{multirow}
\usepackage[utf8]{inputenc} 
\usepackage[T1]{fontenc}    
\usepackage{booktabs}       
\usepackage{amsfonts}       
\usepackage{nicefrac}      
\usepackage{microtype}     
\usepackage{xcolor}        
\usepackage{colortbl}
\usepackage{enumitem}
\usepackage{amssymb}
\usepackage{wrapfig}
\usepackage{courier}

\definecolor{hidden-red}{RGB}{205, 44, 36}
\definecolor{hidden-blue}{RGB}{194,232,247}
\definecolor{hidden-orange}{RGB}{243,202,120}
\definecolor{hidden-green}{RGB}{34,139,34}
\definecolor{hidden-pink}{RGB}{255,245,247}
\definecolor{hidden-black}{RGB}{20,68,106}
\definecolor{purple}{RGB}{144,153,196}
\definecolor{yellow}{RGB}{255,228,123}
\definecolor{hidden-yellow}{RGB}{255,248,203}
\definecolor{tkcolor}{RGB}{224,223,255}
\definecolor{darkblue}{rgb}{0, 0.40, 0.75}
\newcommand{\eg}{\textit{e.g.,}}
\hypersetup{colorlinks=true, citecolor=darkblue, linkcolor=darkblue, urlcolor=darkblue}
\tcbset{
  aibox/.style={
    width=\linewidth,
    top=8pt,
    bottom=4pt,
    colback=blue!6!white,
    colframe=black,
    colbacktitle=black,
    enhanced,
    center,
    attach boxed title to top left={yshift=-0.1in,xshift=0.15in},
    boxed title style={boxrule=0pt,colframe=white,},
  }
}

\fancypagestyle{headstyle}{
    \fancyhead[L]{
        \includegraphics[width=80pt]{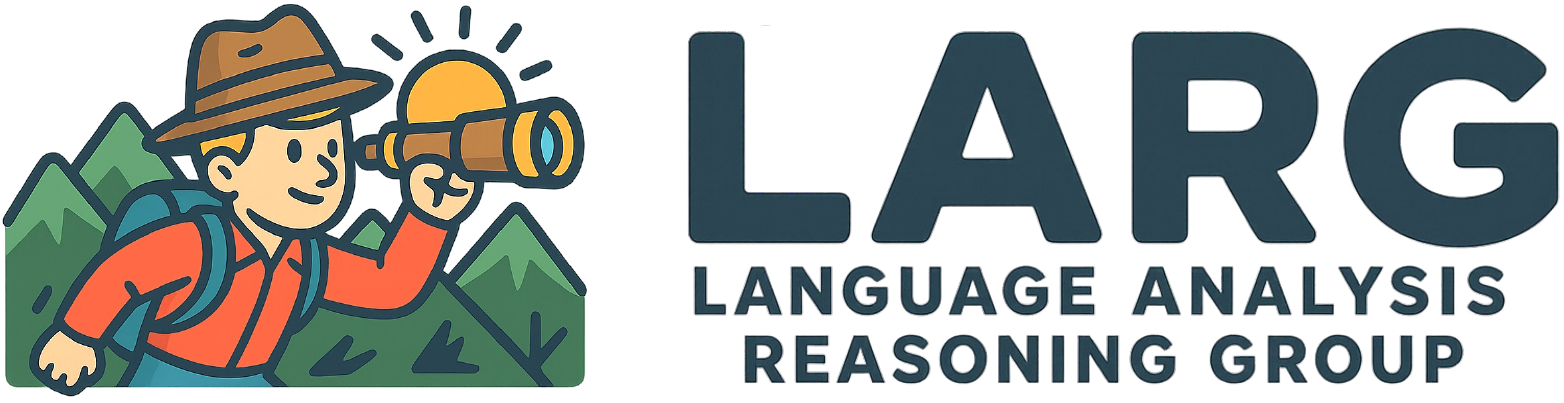}
    }
    \fancyhead[C]{}
    \fancyhead[R]{}

}

\newtcolorbox{AIbox}[2][]{aibox,title=#2,#1}

\tcbset{
  takeawaybox/.style={
    width=\linewidth,
    top=8pt,
    bottom=4pt,
    colback=hidden-yellow,
    colframe=black,
    colbacktitle=black,
    enhanced,
    center,
    attach boxed title to top left={yshift=-0.1in,xshift=0.15in},
    boxed title style={boxrule=0pt,colframe=white,},
  }
}

\newtcolorbox{TakeawayBox}[2][]{takeawaybox,title=#2,#1}

\title{\vspace{-5pt}\raisebox{-0.5em}{\includegraphics[height=2em]{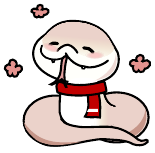}}\ \  Towards Reasoning Era: A Survey of \\ Long Chain-of-Thought for Reasoning Large Language Models }

\author{
  Qiguang Chen$^{\dagger}$ \quad Libo Qin$^{\ddagger}$ \quad Jinhao Liu$^{\dagger}$ \quad Dengyun Peng$^{\dagger}$ \quad Jiannan Guan$^{\dagger}$   \\
  \textbf{Peng Wang$^{\ddagger}$ \quad Mengkang Hu$^{\diamondsuit}$ \quad Yuhang Zhou$^{\heartsuit}$ \quad Te Gao$^{\ddagger}$ \quad Wanxiang Che}$^{\dagger}$ \\
  $^{\dagger}$ LARG, \\
  $^{\dagger}$ Research Center for Social Computing and Interactive Robotics,\\
	$^\dagger$ Harbin Institute of Technology\\
	$^{\ddagger}$ School of Computer Science and Engineering, Central South University \\
	$^\diamondsuit$ The University of Hong Kong \\
  $^{\heartsuit}$ Fudan University\\
	\texttt{\{qgchen,car\}@ir.hit.edu.cn},  \texttt{lbqin@csu.edu.cn} \\
  \\
  Project: \url{https://long-cot.github.io/}\\
  \\
  Github: \href{https://github.com/LightChen233/Awesome-Long-Chain-of-Thought-Reasoning}{\texttt{LightChen233/Awesome-Long-Chain-of-Thought-Reasoning}} \\
  \\
  \\
  \\
  \includegraphics[width=\textwidth]{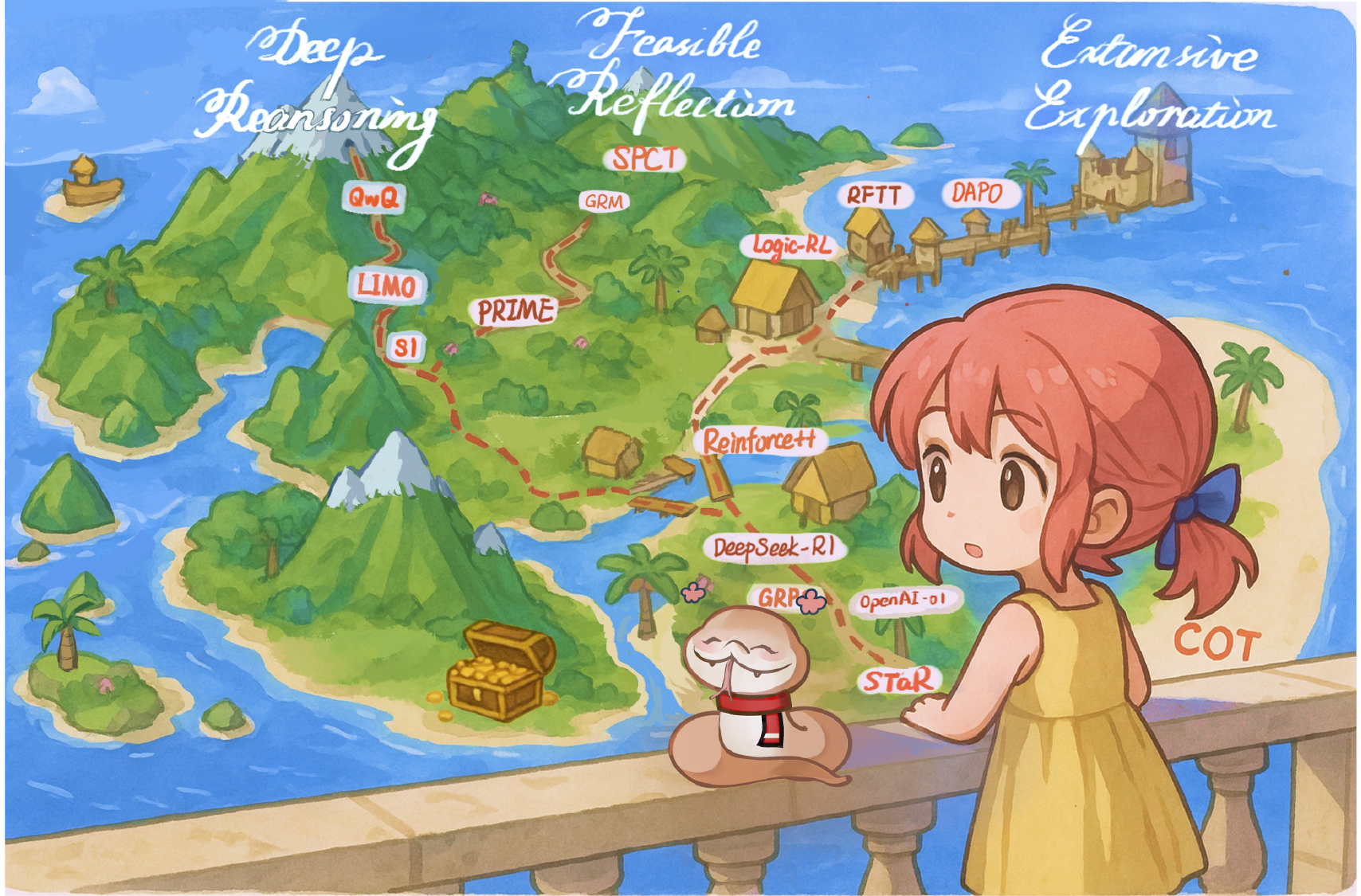}
  \\
  \\
  \\
}

\begin{document}

	\maketitle

  \newpage
	\begin{abstract}
    Recent advancements in reasoning with large language models (RLLMs), such as OpenAI-o1 and DeepSeek-R1, have demonstrated their impressive capabilities in complex domains like mathematics and coding. A central factor in their success lies in the application of long chain-of-thought (Long CoT) characteristics, which enhance reasoning abilities and enable the solution of intricate problems. However, despite these developments, a comprehensive survey on Long CoT is still lacking, limiting our understanding of its distinctions from traditional short chain-of-thought (Short CoT) and complicating ongoing debates on issues like ``overthinking'' and ``inference-time scaling''.
    This survey seeks to fill this gap by offering a unified perspective on Long CoT. Specifically, (1) We first distinguish Long CoT from Short CoT and introduce a novel taxonomy to categorize current reasoning paradigms. (2) Next, we explore the key characteristics of Long CoT: deep reasoning, extensive exploration, and feasible reflection, which enable models to handle more complex tasks and produce more efficient, coherent outcomes compared to the shallower Short CoT. (3) We then investigate key phenomena such as the emergence of Long CoT with these characteristics, including overthinking, and inference-time scaling, offering insights into how these processes manifest in practice.
    (4) Finally, we identify significant research gaps and highlight promising future directions, including the integration of multi-modal reasoning, efficiency improvements, and enhanced knowledge frameworks.
    By providing a structured overview, this survey aims to inspire future research and further the development of reasoning large language models~\footnote{Our logo refers to a cute cartoon image - Snake Puppy. Header Image is inspired by \citet{wang2025multimodal}}.

	\end{abstract}
  \pagestyle{headstyle}
	
\section{Introduction}
\begin{figure*}[t]
    \centering
    \includegraphics[width=0.98\textwidth]{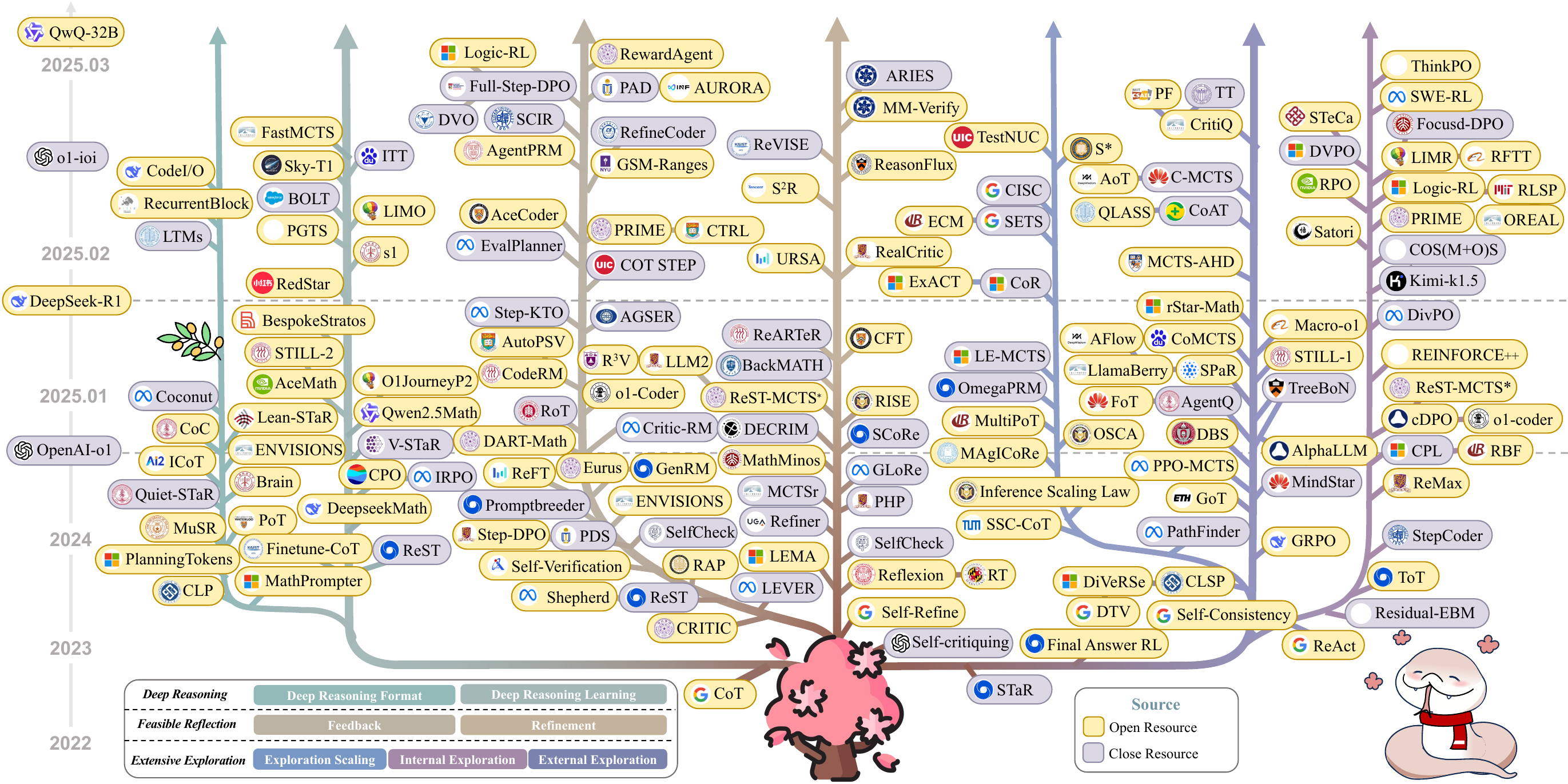}
    \caption{Evolution of selected Long CoT over the past three years, where colored branches represent different characteristics: deep reasoning, feasible reflection, and extensive exploration. Each characteristic is further divided into key areas: Deep reasoning includes its format and learning methods. Feasible reflection focuses on feedback and refinement techniques during reflection process as optimization strategies. Extensive exploration addresses scaling, internal, and external exploration as key improvements to Long CoT.}
    \label{fig:develop}
  \end{figure*}

In recent years, as shown in Figure~\ref{fig:develop}, the emergence of reasoning large language models (RLLMs) such as OpenAI o1~\citep{jaech2024openai} and DeepSeek R1~\citep{guo2025deepseek} has sparked a growing body of research into Long Chain-of-Thought (Long CoT) reasoning, greatly improving their mathematical reasoning, programming tasks, and multidisciplinary knowledge reasoning capabilities~\citep{sun2023survey,yu2024natural,team2025kimi,chen2025chatgpt,yax2024studying,gao2025comparison,zhong2024evaluation,wang2025factors}, even passing Turing Test~\citep{jones2025large}.
This shift marks a significant departure from traditional approaches to task handling in large language models (LLMs)~\citep{zhuang2023lens,qin2024large,qin2025survey,pfister2025understanding}. Unlike the shorter chain-of-thought (Short CoT) used in traditional LLMs, Long CoT reasoning entails a more detailed, iterative process of exploration and reflection within a given problem space by inference-time scaling~\cite{li2025survey,teng2025atom,pmlr-v124-lyzhov20a}. This process has led to notable advancements in mathematical and logical reasoning, as well as in exploring how supervised fine-tuning (SFT) and reinforcement learning (RL) techniques can enhance the learning and exploration of extended reasoning chains~\citep{qin2024o1,min2024imitate}.

However, there is no comprehensive survey to systematically understand the main factors and recent efforts of Long CoT for RLLMs, which hinders the development of RLLMs. As a result, there are ongoing debates about the effectiveness of simple ``inference-time scaling'' for Longer CoT~\citep{wu2024inference,liu2025can} versus the argument that ``over-thinking'' from excessively long scaling can harm LLMs and introduce unnecessary complexity~\citep{chen2024not,cuadron2025danger,kumar2025overthink}. Moreover, some researchers argue that, when solving specific problems, there is no clear relationship between length and accuracy~\citep{xie2025logic}.

\begin{figure*}[t]
	\centering
	\includegraphics[width=0.98\textwidth]{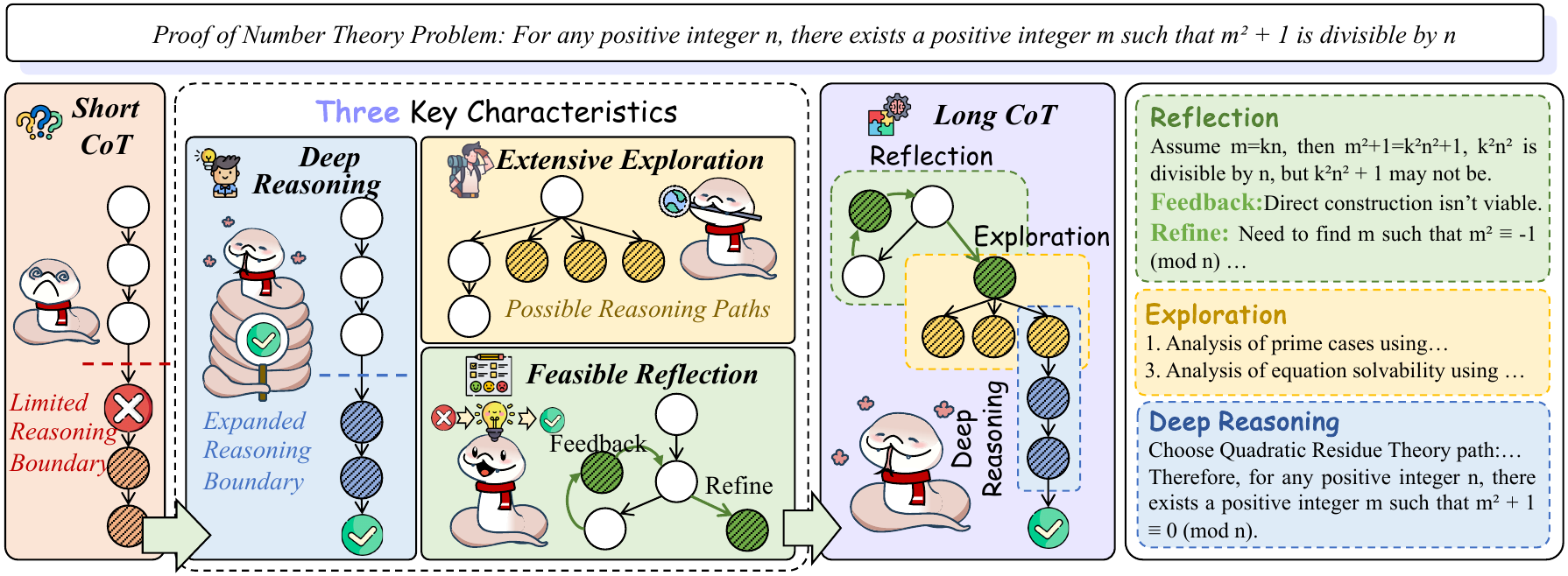}
	\caption{The differences between advanced Long CoT and traditional Short CoT are characterized by three key characteristics: deep reasoning, feasible reflection, and extensive exploration. Moreover, Long CoT integrates all these characteristics to achieve substantial logical efficacy.}
	\label{fig:intro}
\end{figure*}

To address this gap, we provide an extensive and comprehensive survey of Long CoT. Specifically, as illustrated in Figure~\ref{fig:intro}, we first \textbf{define and examine the distinctions} between Long CoT and traditional Short CoT, focusing on the following key aspects: (1) \textit{Deep Reasoning}, which requires a sufficient depth of logical processing to manage an extensive set of logical nodes; (2) \textit{Extensive Exploration}, which involves generating parallel uncertain nodes and transitioning from known to unknown logic; and (3) \textit{Feasible Reflection}, which involves feedback and refinement of logical connections. These characteristics enable Long CoT paradigms to integrate more intricate reasoning and accommodate a broader range of logical structures, ultimately leading to more efficient and coherent outcomes. Subsequently, we systematically \textbf{explore the underlying explanations for key phenomena associated with Long CoT}, such as its emergence, the overthinking phenomenon, inference-time scaling during testing, and the "Aha Moment," among others. To our knowledge, This is the first comprehensive survey dedicated to these specific topics.
Finally, considering the extensive body of literature, we \textbf{highlight promising areas for future research} and suggest valuable open-resource frameworks and datasets that can serve as a foundation for future investigations.

The main contributions of this work are as follows: 
\begin{itemize}[left=2pt,topsep=1pt,itemsep=2pt, parsep=1pt]
    \item \textbf{Systematic Distinction:} In this work, we first introduce the concept of Long CoT reasoning and distinguish it from the traditional Short CoT, thereby providing a clear framework for understanding both paradigms and their respective characteristics.
    \item \textbf{Explanation of Hot Phenomena:} We systematically investigate the notable phenomena associated with Long CoT reasoning, such as overthinking, inference-time scaling, and the ``Aha Moment'', offering valuable insights into the cognitive processes involved in complex reasoning.
    \item \textbf{Emerging Challenges and Frontiers:} We explore the emerging challenges within the field of Long CoT reasoning and identify key research frontiers. Given the vast body of literature, we highlight areas where further inquiry could significantly advance the development of Long CoT methodologies.
\end{itemize}
\section{Discussion of Long CoT v.s. Short CoT}
This section formalizes the key differences between Long Chain-of-Thought (Long CoT) and Short Chain-of-Thought (Short CoT), emphasizing reasoning depth, revisiting connections, and logical node exploration~\citep{wu2024comparative}. These distinctions are clearly separate from System 1 and System 2 thinking. The comparison between Long CoT and Short CoT is framed within System 2, with Long CoT involving more thorough reasoning, reflection, and exploration, while Short CoT generally prioritizes shallow and efficient logic over exhaustive reasoning.

\tikzstyle{my-box}=[
rectangle,
draw=hidden-black,
rounded corners,
text opacity=1,
minimum height=1.5em,
minimum width=5em,
inner sep=2pt,
align=center,
fill opacity=.5,
]
\tikzstyle{leaf}=[
my-box, 
minimum height=1.5em,
fill=yellow!32, 
text=black,
align=left,
font=\normalsize,
inner xsep=5pt,
inner ysep=4pt,
align=left,
text width=45em,
]
\tikzstyle{leaf2}=[
my-box, 
minimum height=1.5em,
fill=purple!27, 
text=black,
align=left,
font=\normalsize,
inner xsep=5pt,
inner ysep=4pt,
]
\tikzstyle{leaf3}=[
my-box, 
minimum height=1.5em,
fill=hidden-blue!57, 
text=black,
align=left,
font=\normalsize,
inner xsep=5pt,
inner ysep=4pt,
]
\begin{figure*}[t]
\vspace{-2mm}
\centering
\resizebox{\textwidth}{!}{
	\begin{forest}
		forked edges,
		for tree={
			grow=east,
			reversed=true,
			anchor=base west,
			parent anchor=east,
			child anchor=west,
			base=left,
			font=\large,
			rectangle,
			draw=hidden-black,
			rounded corners,
			align=left,
			minimum width=4em,
			edge+={darkgray, line width=1pt},
			s sep=3pt,
			inner xsep=2pt,
			inner ysep=4pt,
			line width=1.1pt,
			ver/.style={rotate=90, child anchor=north, parent anchor=south, anchor=center},
		},
		where level=1{text width=9em,font=\normalsize,}{},
        where level=2{text width=9em,font=\normalsize,}{},
        where level=3{text width=9.5em,font=\normalsize,}{},
        where level=4{text width=52em,font=\normalsize,}{},
[Taxonomy, ver
	[\ \ \ Deep Reasoning \\ \ \ for Long CoT~(\S\ref{sec:deep-reasoning}),ver
		[\ \ \ Deep Reasoning \\\ \ \ Formation~(\S\ref{sec:deep-reasoning-excution})
			[\ \ \ Natural Language \\\ \ \ \ Deep Reasoning
				[\eg~CoT~\citep{wei2022chain}{,} Reflection of thought~\citep{zhou2022reflection}{,} MathPrompter~\citep{imani2023mathprompter}{,} CLP~\citep{qin2023cross}{,} AutoCAP~\citep{zhang2024autocap}{,} Plan-\\
				Search~\citep{wang2024planning}{,} NaturalProgram~\citep{ling2023deductive}{,} CodeI/O~\citep{li2025codei}{,} \citet{wang2025thoughts}{,}  \textit{etc.}, leaf, text width=42em]
			]
			[\ Structured Language \\\ \ \ \ Deep Reasoning
				[\eg PoT~\citep{chen2023program}{,} CoC~\citep{li2024chain}{,} Brain~\citep{chen2024brain}{,} SIaM~\citep{yu2024siam}{,} ENVISIONS~\citep{xu2024interactive}{,} SKIntern~\citep{liao2025skintern}{,} QuaSAR~\citep{ranaldi2025improving}{,}\\
				TinyGSM~\citep{liu2023tinygsm}{,} MathDivide~\citep{srivastava2024mathdivide}{,} \citet{payoungkhamdee2025towards}{,} GPT-f~\citep{polu2020generative}{,} STP~\citep{dong2025stp}{,} \textit{etc.}, leaf, text width=42em]
			]
			[\ \ \ \ \ \ Latent Space \\\ \ \ \ Deep Reasoning
				[\eg Quiet-STaR~\citep{zelikman2024quiet}{,} 
				PlaningToken~\citep{wang2023guiding}{,} Coconut~\citep{hao2024training}{,}
				RecurrentBlock\citep{geiping2025scaling}{,} MuSR~\citep{sprague2024musr}{,} SERT\\
				\citep{zhang2025self}{,} Heima~\citep{shen2025efficient}{,} LTMs~\citep{kong2025scalable}{,} ITT~\citep{chen2025inner}{,} \citet{deng2024explicit}{,} \textit{etc.}, leaf, text width=42em]
			]
		]
		[\ \ \ Deep Reasoning \\ \ \ \ Learning~(\S\ref{sec:deep-reasoning-learning})
			[\ \ \ \ Deep Reasoning \\ \ \ \ \ \ \ \ Imitation
				[\eg~GSM8K~\citep{cobbe2021training}{,} AceMath~\citep{liu2024acemath}{,} DART-Math~\citep{tong2024dart}{,} O1-Journey-P2\citep{huang2024o1}{,} STILL-2~\citep{min2024imitate}{,} LIMO\\
				\citep{ye2025limo}{,} s1~\citep{muennighoff2025s1}{,} RedSTaR~\citep{xu2025redstar}{,} Fine-tune-CoT~\citep{ho-etal-2023-large}{,} CoT-Collection~\citep{kim-etal-2023-cot}{,} FastMCTS~\citep{li2025fastmcts}{,} \textit{etc.}
				, leaf, text width=42em]
			]
			[\ \ \ \ Deep Reasoning \\ \ \ \ \ \ Self-Learning
				[\eg~STaR~\citep{zelikman2022star}{,} ReST~\cite{gulcehre2023reinforced}{,}
				DynaThink~\cite{pan-etal-2024-dynathink}{,} V-STaR~\citep{hosseini2024v}{,} PGTS~\citep{li2025policy}{,} CPO~\citep{zhang2024chain}{,} TPO~\citep{liao2024tpo}{,} \\
				OpenRFT~\citep{zhang2024openrft}{,} MCMC-EM~\citep{hoffman2023training}{,} BOLT~\citep{pang2025bolt}{,} Weak2Strong~\citep{yang-etal-2024-weak}{,} Iterative RPO~\citep{pang2024iterative}{,} \textit{etc.}
				, leaf, text width=42em]
			]
		]
	]
	[\ \ Feasible Reflection \\ \ \ for Long CoT~(\S\ref{sec:reflection}), ver
		[\ \ \ Feedback~(\S\ref{sec:reflection-feedback})
			[\ \ \ Overall Feedback
				[\eg Self-Critique~\citep{saunders2022self}{,} SelfCheck~\citep{miao2024selfcheck}{,} Self-Verification~\citep{weng-etal-2023-large}{,} Critic-CoT~\citep{zheng2024critic}{,} Verifier~\citep{cobbe2021training}{,} STaR\\
				\citep{zelikman2022star}{,}ReST~\citep{gulcehre2023reinforced}{,}Critic~\citep{gou2023critic}{,}
				ReFT~\citep{trung-etal-2024-reft}{,}  AceCoder~\citep{zeng2025acecoder}{,} DeepSeek-R1~\citep{guo2025deepseek}{,} Critic-RM~\citep{yu2024self}{,} o1-\\
				coder~\citep{zhang2024o1}{,} 
				RM~\citep{ma2025dynamic}{,}
				Logic-RL~\citep{xie2025logic}{,}  Self-Contrast~\citep{zhang-etal-2024-self-contrast}{,}AGSER~\citep{liu2025attention}{,}DeepSeek-Math~\citep{shao2024deepseekmath}{,} \textit{etc.}
				, leaf3, text width=42em]
			]
			[\ \ \ Process Feedback
				[\eg ReAct~\citep{yao2023react}{,}Reflexion~\citep{shinn2023reflexion}{,}Math-Minos~\citep{gao2024llm}{,}Math-Shepherd~\citep{wang2024math}{,} ER-PRM~\citep{zhang2024entropy}{,}Eurus~\citep{yuan2025advancing}{,} \\
				PAD~\citep{gu2025capturing}{,} PAVs~\citep{setlur2025rewarding}{,} CTRL~\citep{xie2025teaching}{,} QwQ~\citep{team2025qwq}{,} Skywork-o1\citep{team2025skywork}{,} AceMath~\citep{liu2024acemath}{,} PRIME~\citep{cui2025process}{,} \\
				AURORA~\citep{tan2025aurora}{,} RewardAgent~\citep{peng2025agentic}{,} PDS~\citep{xu2024can}{,} COT STEP~\citep{chowdhury2025zero}{,} Step-DPO~\citep{lai2024step}{,} ORPS~\citep{yu2024outcome}{,} \textit{etc.}
				, leaf3, text width=42em]
			]
			[\ \ \ Hybrid Feedback
				[\eg Step-KTO~\citep{lin2025step}{,} \citet{zhang2025lessons}{,} \textit{etc.}
				, leaf3, text width=42em]
			]
		]
		[\ \ Refinement~(\S\ref{sec:reflection-refinement})
			[\ Prompt-based Refine-\\\ \ \ \ ment Generation
				[\eg Reflexion~\citep{shinn2023reflexion}{,} SelfCheck~\citep{miao2024selfcheck}{,} Self-Critique~\citep{saunders2022self}{,} Self-Refine~\citep{madaan2023self}{,} Refiner~\citep{paul-etal-2024-refiner}{,} MCTSr\\
				\citep{zhang2024accessing}{,} ReST-MCTS*~\citep{zhang2024restmcts}{,}LLM2~\citep{yang2024llm2}{,} GLoRe~\citep{havrilla2024glore}{,}\citet{yang2024confidence}{,}RR-MP~\citep{he2024enhancing}{,} PAD~\citep{pan2023automatically}{,} De-\\
				CRIM~\citep{ferraz2024llm}{,}StepCo~\citep{wu2024enhancing}{,} BackMATH~\citep{zhang-xiong-2025-backmath}{,} ReARTeR~\citep{sun2025rearter}{,} START~\citep{huang-etal-2024-advancing}{,} PHP~\citep{zheng2024progressivehint}{,} \textit{etc.}
				, leaf3, text width=42em]
			]
			[SFT-based Refinement \\ \ \ \ \ \ \ \ \ \ Imitation
				[\eg RealCritic~\citep{tang2025realcritic}{,} Self-Debugging~\citep{chen2024teaching}{,} ProgCo~\citep{song2025progco}{,} S3c-Math~\citep{yan2024s}{,} Math-Minos~\citep{gao2024llm}{,} LEMA\\
				\citep{an2023learning}{,} Reflection-tuning~\citep{li2023reflection}{,} CFT~\citep{wang2025critique}{,} \citet{chen2025iterative}{,}\citet{xi2024enhancing}{,} rStar~\citep{qi2024mutual}{,} RISE~\citep{qu2024recursive}{,}R3V\\
				\citep{cheng2024vision}{,} MM-Verify~\citep{sun2025mm}{,} URSA~\citep{luo2025ursa}{,} SIMA~\citep{wang2024enhancingvisual}{,}  CoT-based Synthesizer~\citep{zhang2025cot}{,} \citet{qin2024o1}{,} \textit{etc.}
				, leaf3, text width=42em]
			]
			[\ RL-based Refinement \\\ \ \ \ \ \ \ \ \ Learning
				[\eg SCoRe~\citep{kumar2024training}{,} DeepSeek-r1~\citep{guo2025deepseek}{,} \citet{zeng2025simplerl}{,} S$^{2}$R~\citep{ma2025s}{,} ReasonFlux~\citep{yang2025reasonflux}{,} ReVISE~\citep{lee2025revise}{,} \\
				ARIES~\citep{zeng2025aries}{,} \textit{etc.}
				, leaf3, text width=42em]
			]
		]
	]
	[Extensive Exploration \\ \ \ for Long CoT~(\S\ref{sec:exploration}), ver
    [\ Exploration Scaling\\\ \ \ \ \ \ \ \ \ \ (\S\ref{sec:exploration-scaling})
		[\ \ \ \ Vertical Scaling
			[\eg 
			OpenAI~\cite{jaech2024openai}{,} \citet{fu2023complexitybased}{,} \citet{jaech2024openai}{,} s1~\citep{muennighoff2025s1}{,} \citet{geiping2025scaling}{,} ITT~\citep{chen2025inner}{,} METAL\\
			\citep{li2025metal}{,} RaLU~\citep{li2025reasoning}{,} \citet{ji2025test}{,} \textit{etc.}, leaf2, text width=42em]
		]
		[\ \ \ \ \ Parallel Scaling
			[\eg Self-Consistency~\citep{wang2023selfconsistency}{,}Inference Scaling Law~\citep{wu2024inference}{,}\citet{liu2025can}{,} \citet{zhao2025sample}{,}WoT~\citep{zhang-etal-2024-wrong}{,} \\
			DIVERSE~\citep{li-etal-2023-making}{,} DnA-Eval~\citep{li2025dna}{,}  SSC-CoT~\citep{zhao2024stepwise}{,} ExACT~\citep{yu2025exact}{,}
			\citet{kim2025fostering}{,} S*~\citep{li2025s}{,} 
			CISC\\
			\citep{taubenfeld2025confidence}{,} 
			OmegaPRM~\citep{luo2024improve}{,} 
			AFT~\citep{li2025drafts}{,}Seed-CTS~\citep{wang2024seed}{,}CLSP~\citep{qin2023cross}{,} MultiPoT~\citep{luo-etal-2024-python}{,} ECM~\citep{chen2025ecm}{,} \textit{etc.}
			, leaf2, text width=42em]
		]
	]
	[\ Internal Exploration\\\ \ \ \ \ \ \ \ \ \ (\S\ref{sec:exploration-internal})
		[\ \ \ \ \ \ RL Strategies
		[\eg PPO~\citep{schulman2017proximal}{,} GRPO~\citep{shao2024deepseekmath}{,} REINFORCE++~\citep{hu2025reinforce++}{,} OREO~\citep{wang2024offline}{,} DAPO~\cite{liu2024improving}{,} LIMR~\citep{li2025limr} DAPO\\~\citep{liu2024improving}{,} LIMR~\citep{li2025limr}{,} TRPO~\citep{schulman2017proximal}{,} DVPO~\citep{huang2025lean}{,} RPO~\citep{sun2025reward}{,}PRIME~\citep{cui2025process}{,} DivPO~\citep{lanchantin2025diverse}{,}COS(M+O)S\\
		\citep{materzok2025cos}{,}CPL~\citep{wang2024cpl}{,}  Focused-DPO~\citep{zhang2025focused}{,}RFTT~\citep{zhang2025reasoning}{,}OREO~\citep{wang2024offline}{,} DeepSeekMath~\citep{shao2024deepseekmath}{,}TPO~\citep{yang2025thinking}{,} \textit{etc.}
			, leaf2, text width=42em]
		]
		[\ \ \ Reward Strategies
			[\eg  DeepSeek-R1~\citep{guo2025deepseek}{,}Kimi-k1.5~\citep{team2025kimi}{,}
			T1~\citep{hou2025advancing}{,}ReST-EM~\citep{singh2024beyond}{,}SWE-RL~\citep{wei2025swerl}{,}
			DeepScaleR~\citep{deepscaler2025}{,}\\
			ReST-MCTS*~\citep{zhang2024restmcts}{,} rSTaR-Math~\citep{guan2025rstar}{,} Logic-RL~\citep{xie2025logic}{,} OREAL~\citep{lyu2025exploring}{,} StepCoder~\citep{dou2024stepcoder}{,} RLSP~\citep{ye2025emergence}{,} \\
			Verifier~\citep{cobbe2021training}{,} TS-LLM~\citep{wan2024alphazerolike}{,} STeCa~\citep{wang2025steca}{,} OREO\citep{wang2024offline}{,} \citet{chu2025sft,shen2025satori}{,} \textit{etc.} 
			, leaf2, text width=42em]
		]
    ]
	[\ External Exploration\\\ \ \ \ \ \ \ \ \ \ (\S\ref{sec:exploration-external})
		[\ \ \ \ \ Human-driven \\\ \ \ \ \ \ \ Exploration
			[\eg SPaR~\citep{cheng2024spar}{,} Forest-of-thought~\citep{bi2024forest}{,}Scattered ForestSearch~\citep{light2024scattered}{,}\citet{kang-etal-2024-empirical}{,}AlphaLLM~\citep{ye2024advances}{,}\\
			PATHFINDER~\citep{golovneva2023pathfinder}{,}Least-to-Most~\citep{zhou2023leasttomost}{,} ToT~\citep{yao2023tree}{,} TreeBoN~\citep{qiu2024treebon}{,} CodeTree~\citep{li2024codetree}{,} Tree-of-Code\\
			\citep{ni2024tree} TouT~\citep{mo2024tree}{,} GoT~\citep{besta2024graph}{,} GraphReason~\citep{cao-2024-graphreason}{,} \citet{besta2024demystifying}{,} AoT~\citep{teng2025atom}{,} \citet{chen2024understanding}{,} \textit{etc.}
			, leaf2, text width=42em]
		]
		[\ \ \ \ \ \ Model-driven\\\ \ \ \ \ \ \ Exploration
			[\eg 
			DBS~\citep{zhu2024deductive}{,} \citet{lehnert2024beyond}{,} MindSTaR~\citep{kang2024mindstar}{,} Residual-EBM~\citep{xu2023no}{,} Mulberry~\citep{yao2024mulberry}{,} C-MCTS\\
			\citep{lin2025leveraging}{,} PPO-MCTS~\citep{liu2024making}{,} Llama-Berry~\citep{zhang2024llama}{,}  Marco-o1~\citep{zhao2024marco}{,} AtomThink~\citep{xiang2024atomthink}{,} \citet{puri2025probabilistic}{,} LE-\\
			MCTS~\citep{park2024ensembling}{,} rStar-Math~\citep{guan2025rstar}{,} MC-NEST~\citep{rabby2024mc}{,} CoAT~\citep{pan2025coat}{,} CoPlanner~\citep{wang2024cooperative}{,} CritiQ~\citep{guo2025critiq}{,} \textit{etc.}
			, leaf2, text width=42em]
		]
	]
]
]
	\end{forest}
}
\caption{Taxonomy of Long CoT, which includes deep reasoning, feasible reflection, and extensive exploration methodologies.}
\label{fig:long-cot-survey}

\end{figure*}
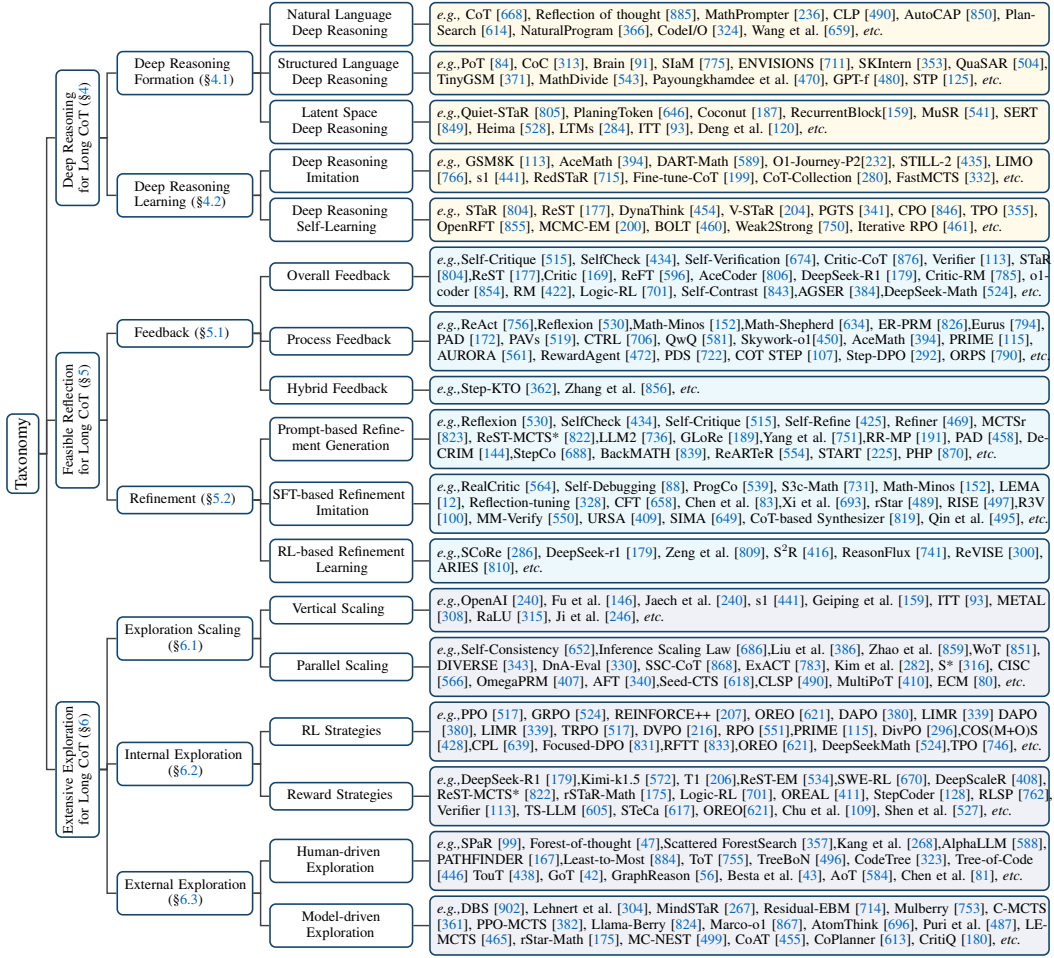

\subsection{Overview of Short CoT}

As illustrated by Figure~\ref{fig:intro}, Short CoT is typically characterized by a shallow, linear reasoning process, where conclusions are drawn sequentially, often relying on a limited number of logical nodes~\citep{mirzadeh2025gsmsymbolic}. This reasoning is usually rapid and straightforward, with simple, surface-level transitions and minimal exploration of alternative paths, which restricts its generalizability~\citep{sprague2024cot}.
Formally, given a reasoning model $\mathcal{R}$, we can define the rationale of Short CoT ($\texttt{CoT}_{S}$) as follows:
\begin{equation}
    \texttt{CoT}_{S} = \mathcal{R}(\{n_i\}^k_{i=1}| (k\le\mathcal{B}_s) \wedge  (j\! =\! 1 \Leftrightarrow  \forall i\!\le\! k, n_{i} \!\rightarrow\! n_{i+j})
    \wedge  (\forall i\!\neq\! j \!\le\! k,  n_i\! \neq\! n_j)),
    \label{eq:short-cot}
\end{equation}
where \( n_1 \) to \( n_k \) represent a sequence of logical nodes, which naturally satisfy that $\forall i, n_{i} \rightarrow n_{i+1}$. Here, $\mathcal{B}_s$ denotes the upper boundary on the number of logical nodes, as defined by \citet{chen2024unlocking}. In this paradigm, the reasoning progresses sequentially from one node to the next, with minimal revisitation of previous nodes and little exploration of alternative logical paths.

\subsection{Overview of Long CoT}
In contrast, Long CoT involves deeper reasoning, reflective analysis, and a broader exploration of logical structures. It facilitates reasoning across a wider range of logical steps, addressing both known and unknown elements of a problem~\citep{gandhi2025cognitive,wu2024comparative}. Building on this, Long CoT expands upon the constraints presented in Equation~\ref{eq:short-cot} based on tree structures by incorporating three critical components: deep reasoning, exploration, and reflection.

These components play distinct yet complementary roles in enhancing cognitive processes. Deep reasoning ensures each logical step is executed rigorously, even within complex structures, fostering robust logic across intricate relationships. Exploration encourages the identification of new pathways, revealing potential avenues that may not be immediately obvious. Reflection enables iterative analysis and reassessment of conclusions, allowing reasoning to evolve throughout problem-solving. By distinguishing these three categories, Long CoT enhances its ability to address a broader range of problems with precision and depth.
As shown in Figure~\ref{fig:long-cot-survey}, we will now discuss these key differences in detail.

\subsubsection{Deep Reasoning for Long CoT}
As shown by Figure~\ref{fig:intro}, deep reasoning refers to the capability to perform deep and thorough logical analysis across multiple interconnected logical nodes, where Short CoT generally can never achieve. This capability is essential when tackling complex problems that require a massive number of logical deductions to arrive at a valid conclusion.
To better define and understand deep reasoning, we frame it as a capability that primarily relaxes the first constraint in Equation~\ref{eq:short-cot}, as expressed by the following:
\begin{equation}
    k\le \mathcal{B}_s \rightarrowtail k\le \mathcal{B}_l \wedge \mathcal{B}_s \ll \mathcal{B}_l,
    \label{eq:deep-reasoning}
\end{equation}
where $\mathcal{B}_l$ represents the upper boundary for Long CoT reasoning, which can accommodate much more intricate logical nodes compared to the smaller boundary $\mathcal{B}_s$ for Short CoT. The larger boundary $\mathcal{B}_l$ alleviates issues related to insufficient depth in reasoning, thereby reducing the risk of generating unresolved answers or hallucinated responses in short-form reasoning.

\begin{AIbox}{Key Difference: Reasoning Depth}
    \begin{itemize}[left=2pt,topsep=1pt,itemsep=2pt, parsep=1pt]
        \item Short CoT typically addresses a limited set of logical nodes, involving shallow reasoning, and struggles with problems requiring complex or intricate logical structures.
    
        \item Long CoT is designed to accommodate a significantly larger set of logical nodes, allowing for deeper logic and more thorough analysis during the reasoning process.
    \end{itemize}
\end{AIbox}

\subsubsection{Extensive Exploration for Long CoT}

As shown by Figure~\ref{fig:intro}, Long CoT encourages branching out to extensively explore uncertain or unknown logical nodes, thereby expanding the potential set of reasoning paths. This exploration is particularly critical when solving problems characterized by ambiguity, incomplete information, or multiple possible solutions~\citep{bashir2025requirements,zeng2024b,wu2025depth}. More specifically, we describe how extensive exploration primarily addresses the relaxation of the second constraint in Equation~\ref{eq:short-cot}, which can be formalized as follows:
\begin{equation}
    j = 1 \Leftrightarrow  \forall i \le k, n_{i} \rightarrow n_{i+j} \rightarrowtail  \exists m, \forall i, \forall j \le m, n_{i} \rightarrow n_{i+j},
    \label{eq:exploration}
\end{equation}
where the condition indicates that for a logical node $n_i$, there are $m$ nodes that are explored in parallel. The acceptability of parallel exploration allows for a more systematic approach, enabling the exploration of previously unconsidered logical paths. This, in turn, helps maximize the understanding of all possible solutions, ultimately leading to the correct final answer.

\begin{AIbox}{Key Difference: Exploration of Logical Nodes}
    \begin{itemize}[left=2pt,topsep=1pt,itemsep=2pt, parsep=1pt]
        \item Short CoT generally restricts exploration to a fixed set of logical nodes, often resulting in oversimplified reasoning and limited exploration.
        \item Long CoT explores more various paths, including uncertain or uncharted areas, fostering more nuanced and comprehensive problem-solving.
    \end{itemize}
\end{AIbox}

\subsubsection{Feasible Reflection for Long CoT}

As shown by Figure~\ref{fig:intro}, Long CoT involves revisiting previous logical nodes to verify their connections are valid and accurate, and then correcting them or selecting an alternative logical path.
Formally, feasible reflection relaxes the third constraint in Equation~\ref{eq:short-cot}, which originally requires acyclic reasoning such that $n_i \ne n_j$ for all $i \ne j \le k$. In contrast, feasible reflection permits the reasoning path to return to a previously visited node, captured as:
\begin{equation}
    \forall i\neq j \le k,  n_i \neq n_j \rightarrowtail \exists i< j \le k, n_{i} = n_{j}, 
\end{equation}
where this condition indicates that, for a logical node $n_{j-1}$, the subsequent node is not limited to the original next node $\hat{n}_{j}$. Instead, it may transition to $n_{i}$ (i.e., the next logical node becomes $n_{j}$, where $n_{j}=n_{i}$).
Practically, reflection implementation consists of two components:\vspace{-1mm}
\paragraph{Feedback} refers to evaluating both overall and intermediate outputs for correctness and quality, also known as critique or verification. It can be derived from external sources, validation checks, or by reflecting on prior conclusions within the reasoning process.
Formally, at each step \( n_i \), a verification process \( \mathcal{V}_i \) ensures the correctness, feasibility, and consistency of the reasoning. If an issue is identified, the process redirects $n_i$ to the nearest correct node $n_{j}$, where $j<i$. This relationship is formalized as:
\begin{equation}
    \mathcal{F}_i, n_{j} \leftarrow \text{Feedback}(\texttt{CoT}_L^i)
    \label{eq:feedback}
\end{equation}
where \( \texttt{CoT}_L^i = \{n_1, \dots, n_i\} \) represents the current logical path up to the $i$-th logical node for Long CoT.\vspace{-1mm}

\paragraph{Refinement} involves adjusting intermediate steps or modifying the logical flow to correct inconsistencies or address gaps based on the given feedback. This process can be expressed mathematically as follows:
\begin{equation}
\widetilde{n}_{i+1} \leftarrow \text{Refine}(n_{i+1}|\texttt{CoT}_L^i, \mathcal{F}_i, n_{j}),
\label{eq:refinement}
\end{equation}
where \( \widetilde{n}_{i+1} \) represents the refined version of the subsequent logical node \( n_{i+1} \), according to the current logic $\texttt{CoT}_L^i$, feedback result $\mathcal{F}_i$, and previous logical node $n_{j}$.

Overall, incorporating reflection ensures that errors are identified and corrected promptly. This capability enables LLMs to quickly shift to alternative reasoning paths or correct their current trajectory. By doing so, error propagation is minimized, resulting in more accurate conclusions.
\begin{AIbox}{Key Difference: Feedback \& Refinement}
    \begin{itemize}[left=2pt,topsep=1pt,itemsep=2pt, parsep=1pt]
        \item Short CoT typically moves in a straightforward, non-repetitive manner from one node to the next, so that cannot correct their logic.
        \item Long CoT allows for revisiting and revising earlier decisions by feedback and refinement, ensuring that optimizable and prior logical conclusions during the reasoning progress.
    \end{itemize}
\end{AIbox}

\subsubsection{Unified Application and Development History of Three Capabilities}
The Long CoT discussed here represents a unified reasoning system that seamlessly integrates and applies three key capabilities: deep reasoning, reflective mechanisms, and exploration capabilities. In contrast, during the Short CoT era, these capabilities developed independently, each evolving in isolation.

As shown in Figure~\ref{fig:intro}, early efforts primarily focused on enhancing deep reasoning within traditional CoT paradigms. This was followed by the gradual introduction of reflective mechanisms, which were initially based on human-designed pipelines. Over time, exploration capabilities were added, and these components were ultimately merged, giving rise to the modern concept of Long CoT, a unified approach to reasoning that seeks to enhance all three capabilities in harmony.

The progression of Long CoT is gradual, rather than a sudden emergence through isolated models like o1~\citep{jaech2024openai} and R1~\citep{guo2025deepseek}. Instead, it develops gradually. For example, earlier systems, such as ToT~\citep{yao2023tree}, enhance exploration but lack reflective mechanisms, disqualifying them as Long CoT~\citep{chen2024understanding}. While GoT~\citep{besta2024graph} incorporates self-reflection based on ToT, its original model still lacked robust deep reasoning, preventing it from qualifying as Long CoT at that time. It is also notable that modern Long CoT systems, often neglect earlier technologies. This article addresses this gap by tracing the evolution of each capability, with the final section offering a comprehensive analysis of the integrated Long CoT system.

In summary, Long CoT and Short CoT represent distinct paradigms. Long CoT features a deeper, broader, and more reflective reasoning process, enhancing both accuracy and coherence. Short CoT, by contrast, is better suited to simpler, well-defined problems. This distinction highlights the scalability and adaptability of Long CoT, making it particularly effective for more complex reasoning.

\begin{AIbox}{Key Difference: Unified Application of Three Capabilities}
    It is important to highlight that Long CoT integrates these three distinct capabilities to perform complex reasoning. In contrast, traditional Short CoT optimization typically focuses on only one of these characteristics.
\end{AIbox}

  \section{Long CoT Analysis \& Evaluation}
\subsection{Analysis \& Explanation for Long CoT}
Research on Long CoT has significantly enhanced RLLMs by improving reasoning accuracy, reducing errors, and supporting dynamic decision-making. However, several phenomena and their corresponding mechanisms remain inadequately summarized. This section addresses key topics, including the mechanisms of Long CoT and their underlying principles~\citep{sauhandikaa2024explainable, cambria2024xai, mccoy2024language, sadr2025think}. Methodologically, two main perspectives have emerged to explain Long CoT: (1) External Behavior Analysis (\S~\ref{sec:external-behavior-analysis}) and (2) Internal Mechanism Analysis (\S~\ref{sec:internal-mechanism-analysis}).

\subsubsection{Long CoT External Behavior Analysis}
\label{sec:external-behavior-analysis}
\begin{figure*}[t]
	\centering
	\includegraphics[width=0.98\textwidth]{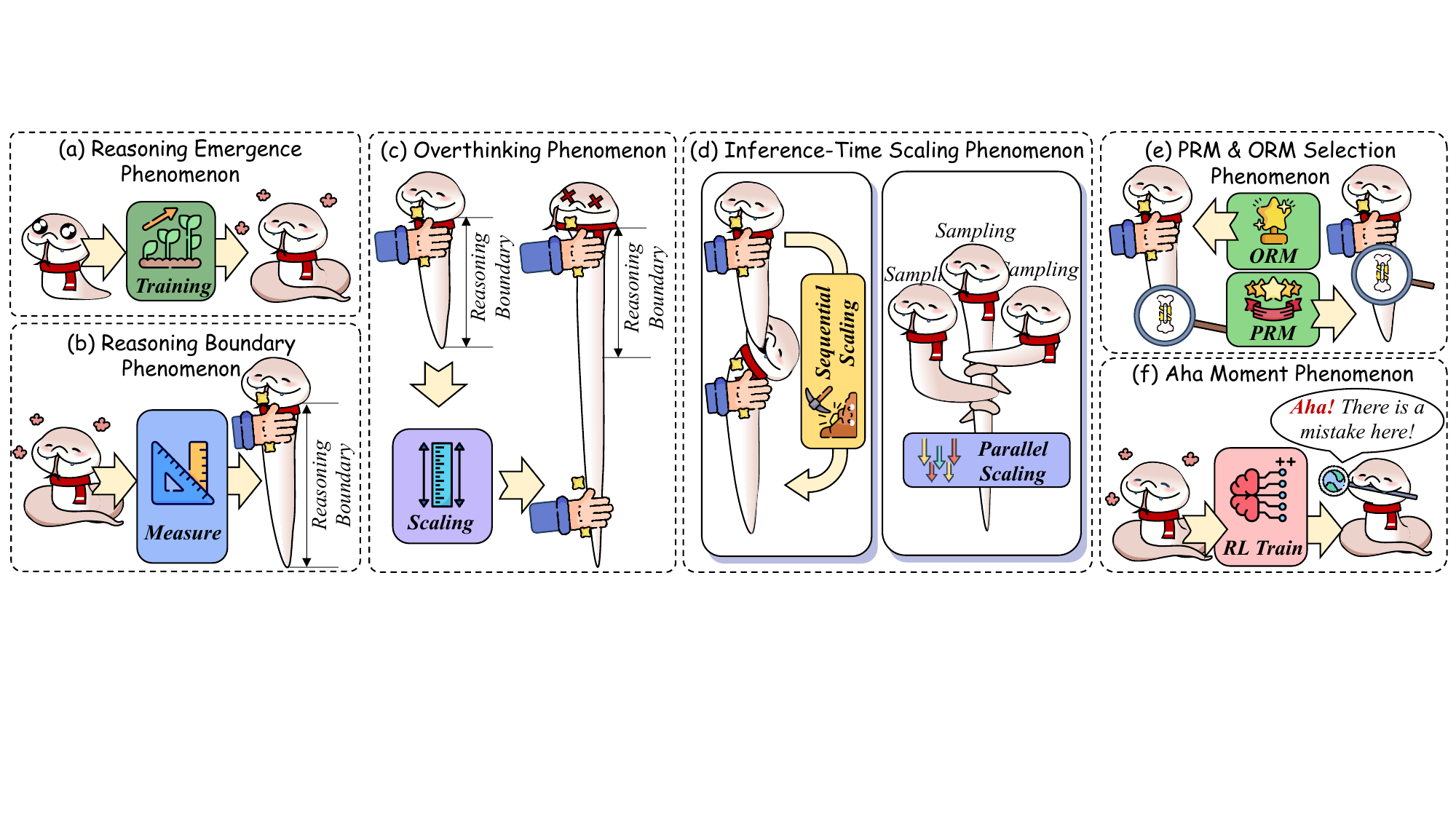}
	\caption{Analysis of the six classic phenomena of Long CoT external behavior: (a) emergence of Long CoT in current RLLMs; (b) reasoning boundaries and limitations of current Long CoT systems; (c) overthinking caused by scaling beyond RLLMs' reasoning boundaries, leading to performance decay; (d) inference-time scaling, discussing mainstream scaling methods, corresponding scaling laws and their limitations; (e) use of process reward model (PRM) or outcome reward model (ORM); (f) exploration of the "aha" moment and its underlying causes.}
	\label{fig:behavior}
\end{figure*}

The primary research stream focuses on explaining RLLM behaviors for Long CoT~\citep{ashok2025language}. As illustrated in Figure~\ref{fig:behavior}, six key phenomena are identified and discussed for Long CoT in this part.\vspace{-1mm}

\paragraph{Long CoT Emergence Phenomenon}
Research shows that contextual examples improve large models' generative abilities by guiding the formation of reasoning chains~\citep{zelikman2022star, shum-etal-2023-automatic, li-qiu-2023-mot, kazemi-etal-2023-lambada, ma2025problem,wen2025from,zeng2025simplerl,zhu2025can}. \citet{wang-etal-2023-towards} and \citet{lippmann2025style} demonstrate that these examples standardize reasoning chain generation relevant to the answers both in in-context-learning and supervised-finetuning. In an experiment by \citet{madaan-etal-2023-makes}, removing problem-specific entities from contextual examples, while retaining only the logical structure, led to similar performance as using complete examples, highlighting the logical structure imitation of Long CoT during inference. From a learning perspective, \citet{ye2025does} analyzes and reveals the three-stage developmental trajectory of Long CoT: early memorization, followed by in-distribution generalization, and ultimately cross-distribution generalization, thereby enabling the model to exhibit Long CoT capabilities.

More recently, \citet{stechly2024chain} and \citet{wang2024chainofthought} have shown that modifying the decoding process or designing specific prompts can activate the Long CoT within pre-trained models. They propose that CoT is embedded during pre-training and requires specific activation~\citep{yang-etal-2024-large-language-models}. Further, \citet{sadr2025think} focus the Long CoT source from the training data, and build on this with the notion of ``model attribution'', to specifically identify the training data most influential for specific outputs. Building on this, \citet{guo2025deepseek} and \citet{xie2025logic} investigate using rule-based reinforcement learning to directly activate Long CoT during pre-training, aiming to enhance performance~\citep{xiang2025towards}.
Furthermore, \citet{gandhi2025cognitive} identify four key cognitive behaviors, including verification, backtracking, sub-target setting, and backward chaining, which successfully facilitate Long CoT. Qwen series models~\citep{yang2024qwen25} inherently demonstrate these behaviors, which can be easily triggered by rule-based reinforcement. In contrast, the models of Llama series~\citep{dubey2024llama} lack these capabilities and thus requires example-based reinforcement learning to improve significantly~\citep{cen2025behavior}. Moreover, \citet{wang2025largerlanguagemodelsimply} identify a pretraining scaling law that explains how increasing calculation size in RLLMs enhances their reasoning capabilities. \citet{wang2025beyond} further explore the scaling law of Long CoT, showing that more fine-grained Long CoT granularity leads to more efficient and effective generalization performance.\vspace{-1mm}

\paragraph{Reasoning Boundary Phenomenon}
Recent research has highlighted the upper bounds and limitations of RLLMs across various reasoning tasks~\citep{imani2023mathprompter, huang2024far, sprague2024musr, hosseini2024not,ferrag2025reasoning,he2025mmboundary}. Specifically, \citet{bi2024program} investigate these bounds in code generation, showing that RLLMs struggle with tasks that exceed certain complexity thresholds~\citep{pham2025clozemath}, especially when imitating Long CoT samples of varying complexity. In the context of upper-bound performance, \citet{merrill2023expressive} and \citet{li2023chain} focus on single-step arithmetic tasks, concluding that model performance is constrained by input length. Moreover, \citet{feng2023towards} proposes a mathematical model indicating that fixed-size models cannot produce accurate numerical answers beyond specific limits. However, increasing the number of reasoning steps improves a model's capability requirements to solve more complex problems.

Inspired by these explorations, \citet{chen2024unlocking} first define the ``reasoning boundary'' phenomenon and quantify these limits, showing that surpassing an RLLM's reasoning capacity leads to performance decline~\citep{chen2025rbf++}. Similarly, \citet{zhou2025gsm} introduce GSM-Infinite, linking different upper limits to accuracy levels. \citet{chen2024unlocking} also examine the interaction between these boundaries across tasks of varying complexity, providing insights into the effectiveness of Long CoT strategies~\citep{zhao2024exploring}. Moreover, \citet{amiri2025lower} propose a ``tight lower bound'' fo
r Long CoT further guiding reasoning error reductions. Further, \citet{baeumel2025lookahead} suggest that due to its reliance on a single-digit lookahead heuristic, there are inherent boundaries in performing addition with multiple operands, which thus hinders the fundamental limitation of LLMs in scaling to more complex numerical reasoning.
\citet{liu2025prorl} further investigate the role of reinforcement learning in expanding these reasoning boundaries instead of relying solely on pretraining capabilities.

\paragraph{Overthinking Phenomenon} Research has highlighted the overthinking phenomenon~\citep{chen2024not, jin2024impact, pan-etal-2024-dynathink, cuadron2025danger, kumar2025overthink,peng2025revisiting}, where performance improves with longer reasoning chains up to a threshold, after which it declines. In contrast, \citet{xie2025logic} and \citet{ma2024step} find no significant correlation between reasoning length and accuracy.
To explain this, one line of research suggests that Long CoT strategies~\citep{arbuzov2025beyond,liao2025lost}, like avoiding ``snowball errors''~\citep{gan2025rethinking}. Alternatively, \citet{chen2024unlocking,wolf2024compositional}  highlight a performance drop when the reasoning boundaries are exceeded, providing an explanation for the overthinking phenomenon. This suggests that reasoning length and logical complexity should be kept below a certain boundary~\citep{zhang2025complexity}. Building on this, \citet{wu2025more} mathematically determine the feasible reasoning length for Long CoT. Finally, \citet{chen2025ecm} introduces Ohm's law of Long CoT, which accurately predicts and controls performance.\vspace{-1mm}

\paragraph{Inference-Time Scaling Phenomenon}
Recent advances in inference-time scaling algorithms~\citep{pmlr-v124-lyzhov20a,welleck2024from} have garnered significant attention, particularly for their ability to extend reasoning length and improve performance~\citep{pmlr-v124-lyzhov20a,lin2025investigating,xia2025rethinking}. 
Specifically, \citet{brown2024large} identify a phenomenon called ``Large Language Monkeys'', in which a series of reasoning tasks show that with enough trials, a correct result can be achieved. Additionally, o1~\citep{jaech2024openai} and R1~\citep{guo2025deepseek} demonstrated that directly scaling the length of model inference improves final performance. 

To understand inference-time scaling, we will discuss these two paradigms:
(1) \textit{\textbf{Sequential Scaling:}} Sequential scaling involves increasing the reasoning path length. While this can enhance performance, studies by \citet{jin2024impact} show that, beyond a certain point, longer reasoning paths can degrade performance due to error accumulation. They suggest an optimal path length that depends on the model’s capabilities and task complexity~\citep{anderson2025phd,setlur2025scaling,balachandran2025inference}.
Furthermore, \citet{chen2024unlocking}  and \citet{wu2025more} explain that excessive exploration lengths beyond the RLLM's inherent reasoning boundary lead to performance decay, which guides RLLMs for deeper reasoning capabilities~\citep{ballon2025relationship}.
(2) \textit{\textbf{Parallel Scaling:}} Parallel scaling involves performing multiple reasoning steps and verifying the results. While it shows promise, \citet{parashar2025inference} and \citet{wang2025examining} argue that simply increasing inference time does not guarantee improved performance. \citet{wu2024inference} show that the computational FLOPs  $N$ of inference are correlated with the lower bound of performance error, which scales with $\log N$. Additionally, \citet{chen2025ecm} establish an upper bound for parallel scaling, showing that RLLMs cannot exceed Pass@k verification through various verifiers. They further argue that sampling optimization cannot exceed the model’s internal reasoning limitations, demonstrating that for $N$ samples, accuracy is proportional to $\frac{m}{(k/\log N+b)^2}$, where $m$, $n$, and $b$ are model-dependent constants.\vspace{-1mm}

\paragraph{PRM \& ORM Selection Phenomenon}
As RLLMs evolve, it is crucial to navigate the debate around the selection between process supervision and outcome supervision, two key reinforcement learning paradigms for complex reasoning tasks. The phenomenon of choosing between these two approaches has become a pivotal issue, as it is essential to differentiate and decide which supervision strategy is more suitable for specific tasks~\citep{xu2025towards,fu2025unveiling,zhang2025process}. While process supervision is intuitively advantageous for long-term reward assignments, the exact relationship between the two approaches remains unclear. It is commonly believed that process supervision is more challenging due to the trajectory-level coverage problem, which demands significant effort to collect fine-grained supervision data~\citep{zheng2024processbench,song2025prmbench}. Additionally, PRM faces the issue of reward hacking~\citep{amodei2016concrete,di2022goal,pan2022effects,baker2025monitoring,li2025rewarding}, where agents exploit flaws in the reward function to produce unintended behaviors~\citep{guo2025deepseek}. Addressing this to surpass rule-based reward systems has become an important research area~\citep{guo2025deepseek,xie2025logic,peng2025agentic}. Furthermore, \citet{lampinen-etal-2022-language} and \citet{tan-2023-causal} establish a causal link between intermediate steps and final answers in qualitative experiments. Building on this, \citet{jia2025we} demonstrate that, under the standard data coverage assumption, reinforcement learning with outcome supervision is not statistically more challenging than process supervision, aside from polynomial factors. More strictly, \citet{he2025response} mathematically demonstrate that outcome-level rewards suffice for online reinforcement learning in RLLMs.\vspace{-1mm}

\paragraph{Aha Moment Phenomenon} Earlier, \citet{guo2025deepseek} demonstrated that direct RL using rule-based rewards can trigger the aha moment, fostering natural self-reflection without supervision~\citep{fan2025cyclicreflex}. Following this, \citet{team2025openr1,xie2025logic} replicate this phenomenon. Further, \citet{zhou2025r1} and \citet{meng2025mmeureka} further extend this phenomenon to multimodal scenarios. However, \citet{liu2025oatzero} argue that the aha moment may not emerge in R1-Zero-like training. Instead, they observe that self-reflection patterns, such as superficial self-reflection (SSR), appear at epoch 0, the stage of base models. In this case, self-reflections do not necessarily lead to correct answers. Upon closer examination of R1-Zero training via RL, they find that the increasing response length results not from self-reflection, but from RL optimizing well-designed rule-based rewards. 
Moreover, \citet{yang2025understandingahamomentsexternal} demonstrate that the ``aha moment'' is externally marked by increased use of anthropomorphic language during self-reflection and a dynamic adjustment of uncertainty in response to problem difficulty. This process enables the model to maintain reasoning without succumbing to "Reasoning Collapse." Internally, it is characterized by a clear distinction between anthropomorphic traits and logical reasoning, with anthropomorphic language intensifying as the problem becomes more complex.

\paragraph{Reinforcement Learning Entropy Phenomenon}
In reinforcement learning for Long CoT, the entropy mechanism is a crucial factor influencing the performance of RLLMs. Policy entropy measures the diversity and exploratory strength of a model’s outputs. By managing this entropy effectively, a model preserves exploration and thus excels on complex reasoning tasks.
Earlier, \citet{jang2022entropy} investigate how initial entropy affects exploration in deep RL and proposed an entropy‐aware initialization to encourage effective exploration. Building on this, \citet{zhang2024entropy} developed an Entropy‐Regularized PRM that balances policy updates against large deviations from the starting distribution, thereby improving reasoning. \citet{cheng2025reasoning} found that high‐entropy regions correlate positively with three exploratory reasoning behaviors: (1) key tokens linking logical steps, (2) self‐verification and correction, and (3) rare behaviors underrepresented in the base model. Most recently, \citet{agarwal2025unreasonable} introduced an Entropy Minimization method and demonstrated its strong impact on LLM performance in mathematical, physical, and coding tasks.

However, recent research indicates that, during early training, policy entropy declines sharply, causing the model to converge prematurely on specific output patterns and limiting further reasoning improvement~\citep{cui2025entropy}. In reinforcement learning, policy entropy ($H$) and downstream task performance ($R$) follow an exponential relation:
$R = -a \cdot e^{H} + b,$
so a drop in entropy produces a rapid performance decline until saturation. This “policy entropy collapse” is common without entropy control, as reduced entropy constrains exploration and stalls reasoning gains~\citep{cui2025entropy}.
To counter this collapse, two methods, Clip-Cov and KL-Cov, regulate entropy by constraining updates on high-covariance tokens. Clip-Cov clips their update magnitudes, whereas KL-Cov imposes a Kullback–Leibler penalty. Empirical results show both techniques prevent collapse and enhance reasoning performance~\citep{cui2025entropy}.

\subsubsection{Long CoT Internal Mechanism Analysis}
\label{sec:internal-mechanism-analysis}
The second stream of research investigates the internal mechanisms of Long CoT-related RLLMs.\vspace{-1mm}

\paragraph{Reasoning Interal Mechanism}
Recent studies have explored the internal mechanisms underlying the coherent rationale outputs of Long CoT, with particular emphasis on attention mechanisms~\citep{skean2025layer,rai-yao-2024-investigation}. These studies primarily examine neural substructures in RLLMs, framing CoT reasoning from a white-box perspective~\citep{wang2023large,yu2025back,hanna2023does,dutta2024think}.
\citet{weston2023system} introduces the concept of System 2 Attention (S2A), which demonstrates Long CoT generation by selectively focusing attention on relevant information. Additionally, \citet{li2024happened} explore gradient distributions between direct output and Long CoT layers, revealing that Long CoT layers help maintain stability by distinguishing relevant from irrelevant reasoning~\citep{wei2025modeling}. Finally, \citet{zhang2025finite} conceptualize RLLMs as finite state automata, offering further insight into how internal dynamics influence external behavior. Despite Short CoT's struggles with self-correction, \citet{bertolazzi2025validation} show that these models rely on \textit{consistency heads} (attention heads) to assess the alignment of numerical values in arithmetic solutions through internal shortcuts.\vspace{-1mm}

\paragraph{Knowledge Incorporating Mechanism} Current RLLMs primarily focus on mathematics and coding but have shown potential for generalization to other knowledge-rich domains, sparking growing interest in the mechanism for integrating domain-specific knowledge into Long CoT~\citep{wu2024thinking,xie2025logic,zheng2025enhancing}. \citet{prystawski2023think} suggest that generative models store entity knowledge learned during pre-training independently, with the reasoning process in Long CoT linking this knowledge across entities.
\citet{radha2025reasoning} recently introduced the Probabilistic Mixture Model (PMM), which categorizes model outputs into reasoning, memorization, and guessing. They also propose an Information-Theoretic Consistency (ITC) analysis to quantify the relationship between model confidence and strategy selection.
Additionally, \citet{jin-etal-2025-exploring} define "Concept Depth" as the lowest layers at which complex concepts are understood, demonstrating varying levels of knowledge integration in RLLMs. \citet{ou2025llms} examine RLLM knowledge internalization through knowledge loop evolution, arguing that new knowledge acquisition is shaped by its connection to existing knowledge, with the loop evolving from formation to optimization and from shallow to deep.

\subsection{Long CoT Evaluations}
\subsubsection{Metrics}

In benchmarking, various metrics assess model performance across reasoning tasks, each focusing on different aspects of reasoning ability. These metrics evaluate both RLLMs' effectiveness in achieving desired outcomes and their learning efficiency. As a result, metrics for RLLMs have gained increasing attention in recent research.
For mathematical or code-related tasks, three key metrics are commonly used: \texttt{Accuracy}, \texttt{Pass@k}, and \texttt{Cons@k} based on regex extraction:
\begin{itemize}[left=2pt,topsep=0pt,itemsep=2pt, parsep=0pt]
    \item \texttt{Accuracy} measures the proportion of correct outputs.
    \item \texttt{Pass@k} evaluates the likelihood of generating at least one correct solution within $k$ attempts.
    \item \texttt{Cons@k} assesses consistency by determining the model's ability to consistently produce correct or logically coherent solutions across multiple attempts.
\end{itemize}
In scientific or commonsense question-answering tasks, evaluation often uses \texttt{Exact Match (EM)} and  \texttt{Accuracy} based on regex extraction, where \texttt{EM}  determines whether the model’s output exactly matches the expected solution.

For feedback techniques like ORM or PRM, \texttt{Rank} and  \texttt{Best-of-N} metrics are often used:
\begin{itemize}[left=2pt,topsep=0pt,itemsep=2pt, parsep=0pt]
    \item \texttt{Rank} measures whether the reward model correctly prioritizes the best reasoning processes from the top $k$ candidates.
    \item \texttt{Best-of-N} selects the highest-scoring solution from $N$ generated reasoning trajectories, indirectly measuring the reward model’s effectiveness based on final outcomes.
\end{itemize}

\subsubsection{Decoding Strategies}
Decoding strategies are essential for controlling the inference process. Common approaches include \texttt{Greedy Decoding}, \texttt{Beam Search}, and \texttt{Major@k}. Both \texttt{Greedy Decoding} and \texttt{Beam Search} limit the sampling range to reduce randomness, guiding the model toward more consistent outputs. In contrast, \texttt{Major@k} identifies the most reliable solution by selecting the one with the highest consistency from a set of $k$ candidate solutions.

\subsubsection{Benchmarks}
In the realm of Benchmarks, the focus lies on assessing the reasoning capabilities of RLLMs across diverse domains. There are two primary categories: (1) Outcome Benchmarks, which focus on the holistic view of Long CoT reasoning, and (2) Process Benchmarks, which concentrate on the local view of the Long CoT process or individual capabilities.\vspace{-1mm}

\paragraph{Outcome Benchmarks}
In the realm of Outcome Benchmarks, the first focus lies on evaluating the logical reasoning capabilities: 
\begin{itemize}[left=2pt,topsep=1pt,itemsep=2pt, parsep=1pt]
    \item \textbf{Complex Mathematics:} A central focus in complex mathematics is evaluating benchmarks like GSM8K~\citep{cobbe2021training} and MATH~\citep{hendrycks2021measuring}, which assess basic mathematical problem-solving abilities~\citep{zhou2024ai,zhong2024achieving}. Recent additions, such as AIME 2024~\citep{aimo2024aime}, AIME 2025~\citep{opencompass2025aime}, MATH-500~\citep{lightman2023let}, AMC 2023~\citep{aimo2024amc}, USAMO~\citep{petrov2025proof}, OlympiadBench~\citep{he-etal-2024-olympiadbench}, and OlympiadArena~\citep{huang2024olympicarena}, expand the evaluation of LLM performance in mathematics. Moreover, Putnam-AXIOM~\citep{gulati2024putnamaxiom} and FrontierMath~\citep{glazer2024frontiermath} introduce more complex problems that challenge future reasoning systems. Additionally, ThinkBench~\citep{huang2025thinkbench} and MATH-Perturb~\citep{huang2025math} focus on robust evaluation for Long CoT~\citep{bao2023assessing,yu2025benchmarking}.
    \item \textbf{Complex Coding:} Complex coding benchmarks are also vital, with competitions like Codeforces, SWEbench~\citep{jimenez2024swebench}, CodeContests~\citep{li2022competition}, and LiveCodeBench~\citep{jain2025livecodebench} evaluating LLM coding and problem-solving skills. Notable additions such as MHPP~\citep{dai2024mhpp}, ProBench~\citep{yang2025probench}, HumanEval Pro, MBPP Pro~\citep{yu2024humaneval}, and EquiBench~\citep{wei2025equibench} enhance the scope and complexity of coding challenges. Moreover, some studies have explored applying these benchmarks in real-world code development scenarios for automatic code generation and evaluation~\citep{he2025code, treude2025interacting}.
    \item \textbf{Commonsense Puzzle:} 
    Commonsense puzzle benchmarks, including LiveBench~\citep{white2025livebench}, BIG-Bench Hard~\citep{suzgun-etal-2023-challenging} and ZebraLogic~\citep{lin2025zebralogic}, assess models’ ability to reason about commonsense situations. The ARC~\citep{chollet2019measure} and DRE-Bench~\citep{yang2025truly} is often viewed as a challenging commonsense-based AGI test. JustLogic~\citep{chen2025justlogic} further contributes to the evaluation of deductive reasoning and commonsense problem-solving. Moreover, \citet{li2025questbench} introduce QuestBench, a benchmark designed to evaluate the ability of RLLMs to generate insightful and meaningful questions.
\end{itemize}

The second focus area concerns Knowledge Benchmarks, essential for evaluating a model's capability in complex reasoning across various tasks for out of distribution evaluation~\citep{wang2025divil}:
\begin{itemize}[left=2pt,topsep=1pt,itemsep=2pt, parsep=1pt]
    \item \textbf{Scientific Reasoning:} Scientific Reasoning benchmarks, such as GPQA Diamond~\citep{rein2024gpqa}, MMLU-Pro~\citep{wang2024mmlupro}, and SuperGPQA~\citep{du2025supergpqa}, assess multi-domain reasoning in fields like chemistry, biology, and physics~\citep{dong2024clr}. These benchmarks test models' ability to not only accumulate knowledge but also integrate it for problem-solving. Humanity's Last Exam (HLE)~\citep{phan2025humanity} further challenges models by requiring deep interdisciplinary reasoning across scientific disciplines. Further, \citet{chung2025theoretical} propose TPBench to evaluate the effectiveness of RLLMs in solving theoretical physics problems.
    \item \textbf{Medical Reasoning:} In the realm of Medical Reasoning, the need for complex, domain-specific, and accurate reasoning is paramount~\citep{zhao2025can,zhan2025evaluation,xu2025deepseek,patil2025cognitive}. Benchmarks, such as MedQA~\citep{jin2021disease}, JAMA Clinical Challenge~\citep{chen2024benchmarking}, LLMEval-Med~\citep{zhang2025llmeval} and Medbullets~\citep{chen2024benchmarking}, simulate diagnostic and treatment decision-making processes, reflecting real-world medical practice. These benchmarks evaluate a model's handling of medical knowledge and reasoning, from diagnosis to treatment planning. Additionally, MedXpertQA~\citep{zuo2025medxpertqa} introduces a comprehensive evaluation framework combining text and multimodal data, specifically assessing AI's reasoning capabilities in healthcare.
\end{itemize}

\subsubsection{Process Evaluations}
\paragraph{Deep Reasoning Benchmarks}
Recent progress in RLLMs underscores the need for specialized benchmarks to evaluate their deep reasoning abilities in Long CoT~\citep{lee2025evaluating,zhou2025landscape}. Notably, \citet{lin2025zebralogic} introduces ZebraLogic, a framework for assessing logical reasoning, especially in complex non-monotonic scenarios. Similarly, BigGSM~\citep{chen2024unlocking} and GSM-Ranges~\citep{shrestha2025mathematical} focus on perturbing numerical values to test logical and arithmetic reasoning in edge cases beyond the models' training distribution. ROSCOE~\citep{golovneva2023roscoe}, ReCEval~\citep{prasad-etal-2023-receval}, DiVeRSe~\citep{li-etal-2023-making}, HLV~\citep{chen2025threading}, and CoT-Kinetics~\citep{bi2025cot} are designed to assess each step in the deep reasoning process during Long CoT tasks.\vspace{-1mm}

\paragraph{Exploration Benchmarks}
Several studies assess RLLMs' exploration capabilities in Long CoT tasks. Specifically, Sys2Bench~\citep{parashar2025inference} evaluates the exploration and scaling abilities of RLLMs, emphasizing generalization across diverse tasks. BanditBench~\citep{nie2024evolve} extends this by testing model performance in interactive environments, offering insights into practical applications. Additionally, \citet{heyman2025evaluating} introduce a graph coloring problem to assess reasoning and spatial exploration in complex problem-solving scenarios.\vspace{-1mm}

\paragraph{Reflection Benchmarks}
Reflection benchmarks measure RLLMs' ability to identify, reflect upon, and correct errors in Long CoT reasoning. These benchmarks fall into two categories: feedback and refinement.
(1) \textbf{\textit{Feedback Benchmark:}} These benchmarks assess the ability of LLMs to detect errors and respond to feedback for improvement. For example, \citet{lambert2024rewardbench} introduces RewardBench to evaluate RLLMs' reward capabilities. This framework is extended by Multimodal RewardBench\citep{yasunaga2025multimodal}, and CodeCriticBench~\citep{zhang2025codecriticbench} to include multimodal and code contexts, respectively. Benchmarks such as ProcessBench~\citep{zheng2024processbench}, PRMBench~\citep{song2025prmbench}, MR-Ben~\citep{zeng2024mrben}, and DeltaBench~\citep{he2025can} focus on error detection and correction across various tasks at the step level. Additionally, ReaLMistake~\citep{kamoi2024evaluating} and JudgeBench~\citep{tan2024judgebench} address more real-world error evaluation.
(2) \textbf{\textit{Refinement Benchmark:}} These benchmarks focus on error correction in complex tasks. CriticBench~\citep{lin-etal-2024-criticbench} assesses critique-correction capabilities, while MLDebugging~\citep{huang2025mldebugging}, and ErrorRadar~\citep{yan2024errorradar} specializes in coding or multimodal reasoning error detection and refinement. FinerReason \citep{chen2025finereason} introduces a commonsense puzzle for broader feedback and refinement evaluations. Medec~\citep{abacha2024medec} adapts error correction to healthcare, addressing medical issues.\vspace{-1mm}

\subsubsection{Advanced Evaluation}
\paragraph{Agentic \& Embodied Reasoning}
Agentic and Embodied reasoning requires models to demonstrate an understanding of real-world interactions, tool use, and adaptive reasoning in response to change. To assess real-world understanding, \citet{wang-etal-2022-scienceworld} introduce a benchmark that evaluates agents' ability to reason about physical concepts. \citet{zhang2025physreason} extend this by assessing agents' interactions with real-world physics. Additionally, realistic tasks often demand complex planning and tool usage, necessitating benchmarks to evaluate agent reasoning. These benchmarks assess agents' abilities to navigate and complete tasks in digital environments. Building on this, \citet{huang2024far} propose a framework for evaluating decision-making in multi-agent, competitive settings. \citet{nath2025toolcomp} introduce ToolComp, a benchmark designed to evaluate multi-step tool-use reasoning. To analyze adaptive reasoning in the face of real-world change, OSWorld~\citep{xie2024osworld}, CogAgent~\citep{hong2024cogagent}, Mobile-Agent-E~\citep{wang2025mobilee}, WebShop~\citep{yao2022webshop}, WebArena~\citep{zhou2024webarena}, WGSR-Bench~\citep{yin2025wgsr}, and WebGames~\citep{thomas2025webgames} assess AI systems across domains such as operating systems, mobile GUIs, browser tasks, and interactive entertainment~\citep{zheng2025vem, wang2025mobile,lu2025ui,mishra2025ttt}. \citet{hu2025text2world} present Text2World, which evaluates agents' ability to generate interactive environments from text to test agent adaptability~\citep{yu2025generating}.\vspace{-1mm}

\paragraph{Multimodal Reasoning}
Multimodal reasoning refers to a system’s ability to integrate and reason across diverse input types, including text, images~\citep{jia2025exploring}. This capability is crucial for solving complex problems that require information from diverse formats.
\begin{itemize}[left=2pt,topsep=1pt,itemsep=2pt, parsep=1pt]
    \item \textbf{Complex Mathematics:} Mathematical reasoning often integrates both textual and visual components, such as equations, graphs, or diagrams~\citep{yan2024survey}. Specifically, challenges like MathVista~\citep{lu2024mathvista}, MathVision~\citep{wang2024measuring}, MathVerse~\citep{zhang2024mathverse}, M3CoT-Math~\citep{chen-etal-2024-m3cot}, CMMaTH~\citep{li-etal-2025-cmmath}, EnigmaEval~\citep{wang2025enigmaeval}, CoMT-Geometry~\citep{cheng2024comt}, and PGPS9K~\citep{zhang2023multimodal} aim to advance multimodal reasoning in mathematics, improving the evaluation of multimodal Long CoT logic.
    \item \textbf{Complex Code:} The second area of focus involves code-related reasoning, where systems interpret textual descriptions and code snippets. Benchmarks like HumanEval-V~\citep{zhang2024humaneval}, Code-Vision~\citep{wang2025code}, Plot2Code~\citep{wu2024plot2code}, and ChartMimic~\citep{yang2025chartmimic} evaluate systems’ capabilities to generate or interpret code from natural language and multimodal inputs for assessing systems that integrate natural language processing with programming tasks.
    \item \textbf{Complex Science:} This area involves integrating scientific texts with related diagrams or experimental data. Benchmarks like ScienceQA~\citep{lu2022learn},  M3CoT-Science~\citep{chen-etal-2024-m3cot}, BMMR~\citep{xi2025bmmr}, and ScienceBoard~\citep{sun2025scienceboard} evaluate how well models combine science information with Long CoT reasoning across various scientific domains~\citep{ye2025mmscibench}. Further, \citet{guo2024can} propose MolPuzzle for the evaluation of molecular structure elucidation.
    \item \textbf{Commonsense Puzzle:} This area focuses on commonsense reasoning, where systems combine reasoning cues and images to make deeper conclusions. \citet{chen-etal-2024-m3cot} introduce M3CoT-Commensense, which incorporates commonsense Long CoT reasoning for complex multimodal interactions. Further, PuzzleVQA~\citep{chia-etal-2024-puzzlevqa}, MMReason~\citep{yao2025mmreason} and LEGO-Puzzles~\citep{tang2025lego} focus more on abstract and spatial puzzle reasoning, respectively. Additionally, \citet{wang2025can} propose two benchmarks: Clue-Visual Question Answering (CVQA), which tests visual comprehension through three task types, and Clue of Password-Visual Question Answering (CPVQA), which features two task types focusing on the interpretation and application of visual data.\vspace{-1mm}
\end{itemize}

\paragraph{AI for Research}
Recent advancements in AI have significantly advanced scientific research~\citep{chen2025ai4research,zhou2025large,wang2025enabling,gottweis2025towards}, with platforms like SciWorld~\citep{wang-etal-2022-scienceworld} improving the research process. Simultaneously, \citet{pricope2025hardml} and \citet{chan2024mle} introduce a machine-learning platform to evaluate the potential of RLLMs in automating experiments. Several studies also examine RLLMs' ability to generate innovative research ideas. For instance, \citet{si2024can} conduct evaluations with over 100 NLP researchers to assess RLLMs' creativity, revealing notable limitations~\citep{li2024chainofideas,wu2025agentic,team2025open}. Additionally, \citet{li2025deepsolution} introduce SolutionBench, a benchmark for assessing systems' ability to generate feasible solutions for complex engineering problems.\vspace{-1mm}

	\section{Deep Reasoning for Long CoT}
\label{sec:deep-reasoning}
Deep reasoning capabilities primarily require profound depth and comprehensiveness in cognitive and reasoning processes. In the absence of such capabilities, RLLMs suffer significant performance declines~\citep{wang2025don, wang2025thoughts}.
Current methods for enhancing deep reasoning can be categorized into two main approaches: (1) \textit{\textbf{Deep Reasoning Format}} (\S~\ref{sec:deep-reasoning-format}), which involves utilizing various reasoning execution formats to maximize the reasoning step length $k$ within reasoning boundary $\mathcal{B}_l$ in Equation~\eqref{eq:deep-reasoning}, by selecting the most suitable reasoning format; and (2) \textit{\textbf{Deep Reasoning Learning}} (\S~\ref{sec:deep-reasoning-learning}), which focuses on improving the model's internal capabilities to enhance its deep reasoning abilities, thereby extending the reasoning boundary $\mathcal{B}_l$ in Equation~\eqref{eq:deep-reasoning} intrinsically.

\begin{figure*}[t]
	\centering
	\includegraphics[width=0.98\textwidth]{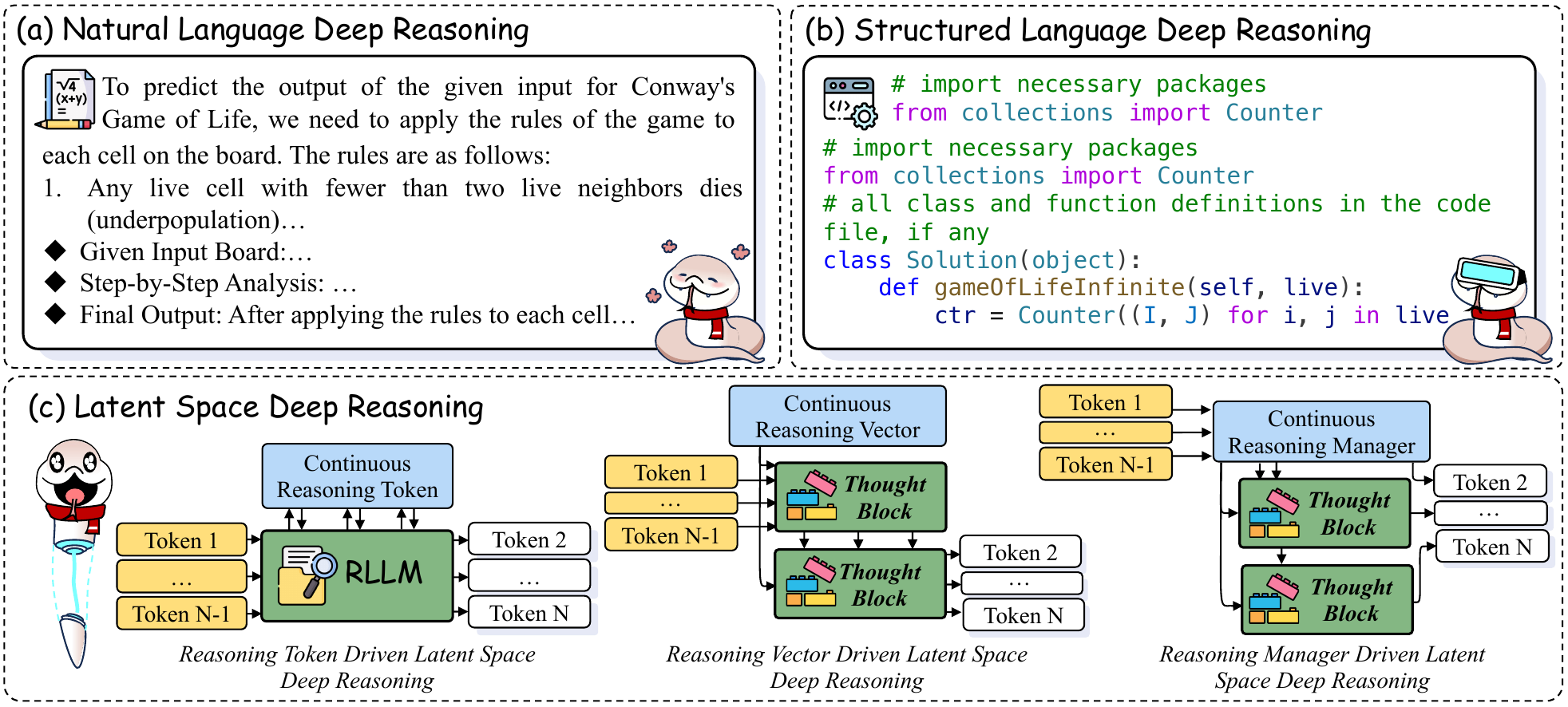}
	\caption{Three main categories of deep reasoning formats: natural language, structured language, and latent-space reasoning (subdivided into token-, vector-, and manager-driven latent reasoning), with examples drawn from \citet{li2025codei}.}
	\label{fig:reasoning-excution}
\end{figure*}

\subsection{Deep Reasoning Format}
\label{sec:deep-reasoning-format}
As illustrated in Figure~\ref{fig:reasoning-excution}, deep reasoning formats can be categorized into three main types: natural language (\S~\ref{sec:nl-deep-reasoning}), structured language (\S~\ref{sec:sl-deep-reasoning}), and latent-space reasoning (\S~\ref{sec:ls-deep-reasoning}), the latter of which is further subdivided into token-, vector-, and manager-driven latent reasoning. The reasoning performance across these formats is presented in Table~\ref{tab:deep-reasoning-format}.

\label{sec:deep-reasoning-excution}
\begin{table*}[t]
	\centering
	\resizebox{0.98\textwidth}{!}{
	\begin{tabular}{l|l|ccccc}
	\toprule
	\textbf{Model} & \textbf{Base Model} & \textbf{GSM8k} & \textbf{MATH} & \textbf{GPQA} & \textbf{OlympiadBench} & \textbf{LiveCodeBench} \\
	\midrule
		\rowcolor{gray!8}\multicolumn{7}{c}{\textit{Latent Space Deep Reasoning}}\\
	\midrule
	No-CoT~\citep{deng2024explicit} & Mistral-7B~\citep{jiang2023mistral7b} & 38.0 & - & - & - & - \\
	SQ-VAE~\citep{wang2023guiding} & Llama-2-7B~\citep{touvron2023llama} & 40.0 & 7.0 & - & - & - \\
	RecurrentBlock-3.5B~\citep{geiping2025scaling} & - & 42.1 & - & - & - & - \\
	ICoT-SI~\citep{deng2024explicit} & Mistral-7B~\citep{jiang2023mistral7b} & 51.0 & - & - & - & - \\
	\midrule
		\rowcolor{gray!8}\multicolumn{7}{c}{\textit{Natural Language Deep Reasoning}}\\
	\midrule
	Self-Rewarding~\citep{pmlr-v235-chen24j} & Llama-2-7B~\citep{touvron2023llama} & 40.0 & 10.7 & - & - & - \\
	Llama-3.1-8B~\citep{dubey2024llama} & - & 56.7 & 20.3 & - & - \\
	MetaMath~\citep{yu2024metamath} & Llama-2-7B~\citep{touvron2023llama} & 66.5 & - & - & - & - \\
	OVM~\citep{yu2024ovm} & Llama-2-7B~\citep{touvron2023llama} & 73.7 & - & - & - & - \\
	NuminaMath-7B-CoT~\citep{li2024numina} & - & 75.4 & 55.2 & - & 19.9 & - \\
	Qwen2-7B~\citep{yang2024qwen2} & - & 79.9 & 44.2 & - & 21.3 & - \\
	Qwen2-Math-7B~\citep{yang2024qwen25math} & - & 80.4 & 50.4 & - & 38.2 & - \\
	Internlm2-math-plus-7B~\citep{ying2024internlmmath} & - & 84.0 & 54.4 & - & 18.8 & - \\
	OMI2~\citep{li2025codei} & Qwen2.5-Coder-7B~\citep{hui2024qwen25coder} & 84.1 & 72.3 & 36.2 & - & 27.2 \\
	Llama-3.1-70B~\citep{dubey2024llama} & - & 85.5 & 41.4 & - & - & - \\
	CODEI/O++~\citep{li2025codei} & Qwen2.5-Coder-7B~\citep{hui2024qwen25coder} & 85.7 & 72.1 & 40.6 & - & 29.1 \\
	CODEI/O~\citep{li2025codei} & Qwen2.5-Coder-7B~\citep{hui2024qwen25coder} & 86.4 & 71.9 & 43.3 & - & 28.5 \\
	WI~\citep{li2025codei} & Qwen2.5-Coder-7B~\citep{hui2024qwen25coder} & 87.0 & 71.4 & 39.1 & - & 26.0 \\
	WI (Full)~\citep{li2025codei} & Qwen2.5-Coder-7B~\citep{hui2024qwen25coder} & 87.0 & 71.1 & 42.9 & - & 27.6 \\
	OMI2 (Full)~\citep{li2025codei} & Qwen2.5-Coder-7B~\citep{hui2024qwen25coder} & 88.5 & 73.2 & 40.9 & - & 28.4 \\
	DeepSeekMath-7B-RL~\citep{shao2024deepseekmath} & - & 88.2 & 51.7 & - & 19.0 & - \\
	Llama-3.1-405B~\citep{dubey2024llama} & - & 89.0 & 53.8 & - & - & - \\
	CoMAT~\citep{leang2024comat} & GPT-4~\citep{achiam2023gpt} & 93.7 & - & 40.4 & - & - \\
	CoT~\citep{ranaldi2025improving} & GPT-4~\citep{achiam2023gpt} & 94.5 & - & 41.8 & 50.2 & - \\
	FCoT~\citep{lyu-etal-2023-faithful} & GPT-4~\citep{achiam2023gpt} & 95.0 & - & - & - & - \\
	Qwen2.5-Math-7B-Instruct~\citep{yang2024qwen25math} & - & 95.2 & 83.6 & - & 41.6 & - \\
	MathPrompter~\citep{imani2023mathprompter} & GPT-4~\citep{achiam2023gpt} & 95.6 & - & - & - & - \\
	Qwen2.5-Math-72B-Instruct~\citep{yang2024qwen25math} & - & 95.9 & 85.9 & - & 49.0 & - \\
	DeepSeek-R1-Distill-Qwen-7B~\citep{guo2025deepseek} & - & - & 92.8 & - & 49.1 & 37.6 \\
	DeepSeek-R1-Distill-Qwen-32B~\citep{guo2025deepseek} & - & - & 94.3 & - & 62.1 & 57.2 \\
	\midrule
		\rowcolor{gray!8}\multicolumn{7}{c}{\textit{Structured Language Deep Reasoning}}\\
	\midrule
	STaR~\citep{zelikman2022star} & Llama-2-7B~\citep{touvron2023llama} & 58.2 & 16.0 & - & - & - \\
	ENVISIONS~\citep{xu2024interactive} & Llama-2-7B~\citep{touvron2023llama} & 59.0 & 19.0 & - & - & - \\
	MAmmoTH~\citep{yue2023mammoth} & Code-Llama-7B~\citep{roziere2023code} & 59.4 & - & - & - & - \\
	MathCoder-CL~\citep{wang2024mathcoder} & Code-Llama-7B~\citep{roziere2023code} & 67.8 & 30.2 & - & - & - \\
	ToRA-Code~\citep{gou2023tora} & Llama-2-7B~\citep{touvron2023llama} & 72.6 & - & - & - & - \\
	Brain~\citep{chen2024brain} & Code-Llama-7B~\citep{roziere2023code} & 74.0 & - & - & - & - \\
	DeepSeek-Coder-7B~\citep{guo2024deepseek} & - & 77.4 & 44.4 & - & - & - \\
	SIaM~\citep{yu2024siam} & Qwen-2-Math-Base & 81.5 & 50 & - & - & - \\
	OC-SFT-1~\citep{li2025codei} & Qwen2.5-Coder-7B~\citep{hui2024qwen25coder} & 86.7 & 70.9 & 37.7 & - & 27.5 \\
	PyEdu~\citep{li2025codei} & Qwen2.5-Coder-7B~\citep{hui2024qwen25coder} & 85.8 & 71.4 & 40.9 & - & 25.8 \\
	Qwen2.5-Math-7B-Instruct~\citep{yang2024qwen25math} & - & 94.6 & 85.2 & - & 55.6 & - \\
	Qwen2.5-Math-72B-Instruct~\citep{yang2024qwen25math} & - & 95.8 & 88.1 & - & 60.6 & - \\
	QuaSAR~\citep{ranaldi2025improving} & GPT-4~\citep{achiam2023gpt} & 96.5 & - & 55.4 & 44.6 & - \\
	MathDivide~\citep{srivastava2024mathdivide} & GPT-4~\citep{achiam2023gpt} & 96.8 & - & - & - \\
	\bottomrule
	\end{tabular}
	}
	\caption{Performance of various deep reasoning formats, sorted primarily by GSM8K scores. ``-'' indicates that the paper did not report this score.}
	\label{tab:deep-reasoning-format}
\end{table*}
\subsubsection{Natural Language Deep Reasoning}
\label{sec:nl-deep-reasoning}
Traditionally, researchers have sought to adapt natural language for intuitive and free-flowing deep reasoning~\citep{wei2022chain,zhou2022reflection,imani2023mathprompter,qin2023cross,zhang2024autocap,wang2024planning,geal2024large}. Early work by \citet{wei2022chain} demonstrated that the use of natural language Long CoT significantly enhances the reasoning capabilities of RLLMs. Further, the Natural Program framework~\citep{ling2023deductive} allows RLLMs to engage in deeper natural language reasoning by ensuring a more structured and rigorous logical analysis. More recently, CodeI/O~\citep{li2025codei} has introduced a technique that reorganizes code-based reasoning patterns into natural language formats, further boosting the reasoning potential of RLLMs~\citep{bao2025teaching}. Similarly, \citet{li2025cort} propose CoRT, which integrates code into reasoning to facilitate a mixture of formats, resulting in improved cognitive performance.

\subsubsection{Structured Language Deep Reasoning}
\label{sec:sl-deep-reasoning}
Structured language deep reasoning encompasses various approaches designed to program~\citep{chen2023program,liu2023tinygsm,srivastava2024mathdivide,payoungkhamdee2025towards,pmlr-v202-gao23f,wen2025codeplan,wang2023chatlogic,zhang2025computational} or symbolic language~\citep{polu2020generative,dong2025stp,lin2024lean,leang2025theorem,yang2024formal,poesia2024certified,bao2022multi,bao-etal-2024-abstract,wang2025ma,leng2025semi} format for enhanced deep reasoning. In this context, most studies focus on utilizing code to better enhance the mathematical reasoning capabilities~\citep{li2024chain,chen2024brain,yu2024siam,chen2025code}.
\citet{xu2024interactive} propose a neural-symbol self-training framework guided by the environment, addressing both the scarcity of symbolic data and the limitations of symbolic processing in LLMs.
Additionally, \citet{liao2025skintern} present SKIntern, which refines symbolic RLLMs through curriculum learning and linear attenuation, enabling the internalization of symbolic knowledge with fewer examples, reducing computational costs, and accelerating inference. Furthermore, \citet{ranaldi2025improving} introduce QuaSAR, a CoT variant that directs LLMs to operate at higher abstraction levels through quasi-symbolic reasoning, thus improving natural language reasoning and providing more precise structural representations.

\subsubsection{Latent Space Deep Reasoning}
\label{sec:ls-deep-reasoning}
Latent space deep reasoning encompasses techniques designed to enhance the reasoning abilities of LLMs by leveraging operations within continuous latent spaces~\citep{sprague2024musr,deng2024explicit,ruan2025reasoning,jiang2025dart}. These approaches can be categorized into three main paradigms:
(1) \textit{\textbf{Reasoning Token-Driven Latent Space Deep Reasoning:}} Early work~\citep{wang2023guiding,zelikman2024quiet} introduce the concept of ``planning tokens'' or ``thought tokens'' to guide reasoning within latent spaces~\citep{yang2025machine,yue2025hybrid}. Further, Coconut~\citep{hao2024training} expands on this through the maintenance of multiple alternative reasoning paths, increasing both complexity and efficiency~\citep{zhang2025self,tack2025llm}. At the extreme, Heima~\citep{shen2025efficient} condenses the entire Long CoT process into a single token, yielding substantial computational savings.
(2) \textit{\textbf{Reasoning Vector Driven Latent Space Deep Reasoning:}} Building on the previous paradigm, LTM~\citep{kong2025scalable} conceptualizes the layers of LLMs as ``thought blocks'' and introduces the concept of ``thought vectors'' for each layer. This approach allows for the scaling of inference-time computations by implicitly performing reasoning within the latent space through recurrent depth.
(3) \textit{\textbf{Reasoning Manager Driven Latent Space Deep Reasoning:}} Inspired by these, \citet{schone2025implicit}, \citet{geiping2025scaling}, and \citet{saunshi2025reasoning} propose a mechanism similar to a continuous reasoning manager, which iteratively governs a trained ``recurrent block'' as a recurrent ``thought block''~\citep{lu2025latent}. This method integrates deeper model layers during reasoning, enhancing performance without needing specialized training data, and even outperforming larger RLLMs. Additionally, ITT~\citep{chen2025inner} leverages the original transformer layer as a recurrent ``thought block'', selecting key tokens via adaptive token routing and controlling reasoning depth with residual thinking connections, enabling more efficient processing of critical tokens. Further, System-1.5 Reasoning~\citep{wang2025system} defines two dynamic shortcuts. The Model Depth Shortcut (DS) lets non-critical tokens exit early via lightweight adapter branches while routing critical tokens through deeper Transformer layers, thus supporting adaptive, vertical reasoning. The Step Shortcut (SS) reuses hidden states across decoding steps to bypass trivial iterations and enable horizontal reasoning in latent space.

\begin{figure*}[t]
	\centering
	\includegraphics[width=0.98\textwidth]{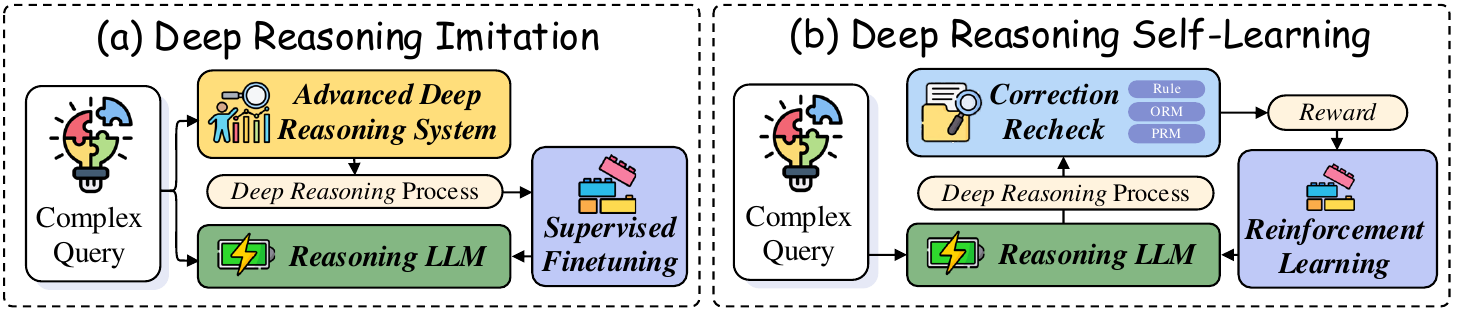}
	\caption{The different learning strategies of deep reasoning learning, including deep reasoning imitation of the data from advanced deep reasoning systems, like advanced RLLMs, MCTS, etc.; deep reasoning self-learning from preference-based RL by implicit reward.}
	\label{fig:deep-reasoning-learning}
\end{figure*}

\subsection{Deep Reasoning Learning}
\label{sec:deep-reasoning-learning}
Insufficient deep reasoning in RLLMs can significantly degrade performance~\citep{wang2025don, wang2025thoughts}. As a result, research has focused on improving reasoning through training. Supervised fine-tuning (SFT)~\citep{zhang2023instruction} stabilizes model outputs by serving as a memory process~\citep{xie2024memorization}, while reinforcement learning (RL) enables generalization and self-learning~\citep{guo2025deepseek,chu2025sft,huan2025does,xu2025genius}. Recent studies for deep reasoning learning have explored using SFT to imitate advanced reasoning in RLLMs and applying RL to enhance self-improvement in reasoning.
As illustrated in Figure~\ref{fig:deep-reasoning-learning}, this section outlines two key approaches to improve deep reasoning: (1) \textbf{\textit{Deep Reasoning Imitation}} (\S~\ref{sec:deep-reasoning-imitation}), which involves learning reasoning from human-annotated or distilled data through SFT, and (2) \textbf{\textit{Deep Reasoning Self-Learning}} (\S~\ref{sec:deep-reasoning-self-learning}), where models improve reasoning through preference-based RL with implicit rewards. The performance of these methods is shown in Table~\ref{tab:deep-reasoning-learning}.

\subsubsection{Deep Reasoning Imitation}
\label{sec:deep-reasoning-imitation}
Deep reasoning in RLLMs can be effectively achieved by mimicking advanced reasoning systems, such as human reasoning~\citep{morishita2024enhancing,cai2024system,chen2025advancing,li2025sos1}, advanced RLLMs~\citep{guo2025deepseek,busbridge2025distillation,yao2025unveiling,le2025cot2align,chen2025unveiling}, and scaling-augmented RLLMs~\citep{li2025fastmcts,yuan2025agent,peng2025rewarding,zhu2025chain,bao2024exploring}. This approach enables the model to learn complex reasoning patterns and generalize across tasks~\citep{yang2022chain,li2024unlocking}.
Specifically, (1) \textbf{\textit{Imitation from Human:}} Earlier, \citet{cobbe2021training} first propose the deep reasoning imitation paradigm using human examples. ALT~\citep{morishita2024enhancing} improves RLLM reasoning by generating larger datasets of human-annotated logical templates, which fosters deeper reasoning~\citep{he2025protoreasoning}. To enhance diversity, EIT~\citep{cai2024system} promotes simpler human-generated plans, while LLMs contribute more nuanced reasoning, facilitating collaboration between human input and AI.
(2) \textbf{\textit{Imitation from Advanced RLLMs:}}
A body of work utilizes zero-shot prompting to guide large teacher RLLMs in generating reasoning rationale, which is then used to fine-tune smaller RLLMs, marking the beginning of deep reasoning imitation~\citep{ho-etal-2023-large, kim-etal-2023-cot, yeo2025demystifying,luo2025deconstructing}. Additionally, AceMath~\citep{liu2024acemath} applies few-shot prompting to distill Long CoT samples from advanced LLMs, followed by multi-stage quality-guided SFT to enhance performance. \citet{chen2024brain} separate the data synthesis process into planning and reasoning stages, thereby improving reasoning quality. DART-Math~\citep{tong2024dart} effectively distills complex queries requiring deeper reasoning during synthesis, advancing deep reasoning capabilities. Further, \citet{ahmad2025opencodereasoning} propose OpenCodeReasoning, expanding this paradigm to the code scenarios.
(3) \textbf{\textit{Imitation from Scaling-augmented RLLMs:}} Earlier, \citet{bansal2024smaller} enhance data quality by scaling the sampling size and length, boosting imitation performance~\citep{liu2025synlogic,yuan2025incentivizing}. \citet{yang2024qwen25math} and \citet{zhao2025promptcot} further improve data quality by scaling sampling and selecting samples through sample feature or an additional reward model. Additionally, \citet{li2025fastmcts} identify optimal deep reasoning paths through MCTS, advancing imitation effectiveness.

Recent studies~\citep{huang2024o1, min2024imitate} show that distilling knowledge from advanced RLLM APIs like O1~\citep{jaech2024openai} and R1~\citep{guo2025deepseek} significantly enhances the performance of smaller LLMs~\citep{li2025naturalthoughts,guha2025openthoughts}. This method, employing supervised fine-tuning, boosts model performance on complex mathematical reasoning tasks, sometimes surpassing the teacher models' performance. Building on these findings, LIMO~\citep{ye2025limo}, S1~\citep{muennighoff2025s1}, and RedStar~\citep{xu2025redstar} argue that a large number of imitation samples is unnecessary. They demonstrate that even a minimal set of samples can activate deep reasoning capabilities in foundational LLMs. For practical applications, \citet{turtel2025llms} showcase how these techniques can predict future events beyond a model's knowledge cutoff. \citet{sun2025improving}, \citet{yang2025select2reason} and \citet{zhao2025ufo} further enhance deep reasoning imitation by selecting high-quality samples from large datasets, thereby improving the quality of the imitation data.

\begin{table*}[t]
	\centering
	\resizebox{\textwidth}{!}{
	\begin{tabular}{l|l|l|cccccc}
	\toprule
	\textbf{Model} & \textbf{Data Size} & \textbf{Base Model} & \textbf{GSM8K} & \textbf{MATH} & \textbf{MATH-500} & \textbf{AIME2024} & \textbf{GPQA} & \textbf{OlympiadBench} \\
\midrule
	\rowcolor{gray!8}\multicolumn{9}{c}{\textit{Deep Reasoning Imitation}}\\
	\midrule
	SFT~\citep{yeo2025demystifying} & 200K & Llama-3.1-8B~\citep{dubey2024llama} & - & - & - & 54.1 & 3.5 & - \\
	Retro-Enh~\citep{chen2025advancing} & 14M & Llama-3-8B~\citep{dubey2024llama} & 45.1 & 21.7 & - & - & - & - \\
	Query-Exp~\citep{chen2025advancing} & 24M & Llama-3-8B~\citep{dubey2024llama} & 51.3 & 23.1 & - & - & - & - \\
	Res-Div~\citep{chen2025advancing} & 14M & Llama-3-8B~\citep{dubey2024llama} & 53.0 & 23.2 & - & - & - & - \\
	MetaMath~\citep{tong2024dart} & 0.40M & Mistral-7B~\citep{jiang2023mistral7b} & 76.5 & 29.8 & - & - & - & 5.9 \\
	ALT-FLDx2~\citep{morishita2024enhancing} & 100K & Llama-3.1-70B~\citep{dubey2024llama} & 83.3 & 24.4 & - & - & - & - \\
	EIT~\citep{cai2024system} & 15K & Llama-2-70B~\citep{touvron2023llama} & 84.1 & 32.5 & - & - & - & - \\
	MathScale~\citep{tong2024dart} & 2.0M & Mistral-7B~\citep{jiang2023mistral7b} & 74.8 & 35.2 & - & - & - & - \\
	Tutor-Amp~\citep{chen2025advancing} & 11M & Llama-3-8B~\citep{dubey2024llama} & 64.4 & 35.9 & - & - & - & - \\
	
	MMIQC~\citep{tong2024dart} & 2.3M & Mistral-7B~\citep{jiang2023mistral7b} & 75.4 & 37.4 & - & - & - & 9.4 \\
	VRT~\citep{tong2024dart} & 0.59M & Mistral-7B~\citep{jiang2023mistral7b} & 82.3 & 38.7 & - & - & - & 8.7 \\
	KPMath-Plus~\citep{tong2024dart} & 1.6M & Mistral-7B~\citep{jiang2023mistral7b} & 82.1 & 46.8 & - & - & - & - \\
	Llama-2-70B-Xwin-Math-V1.1~\citep{li2024common} & 1.4M & Llama-2-70B~\citep{touvron2023llama} & 90.2 & 52.5 & - & - & - & 16.3 \\
	DART-Math-Mistral-7B~\citep{tong2024dart} & 591K & Mistral-7B~\citep{jiang2023mistral7b} & 81.1 & 45.5 & - & - & - & 14.7 \\
	DART-Math-Llama-3-70B~\citep{tong2024dart} & 591K & Llama-3-70B~\citep{dubey2024llama} & 89.6 & 56.1 & - & - & - & 20.0 \\
	
	Rejection Sampling~\citep{li2025fastmcts} & 197K & Qwen2.5-7B~\citep{yang2024qwen25} & 87.1 & 70.0 & - & 10.0 & - & 27.1 \\
	Evol-Instruct-7B~\citep{luo2023wizardmath} & 905K & Qwen2.5-Math-7B~\citep{yang2024qwen25math} & 88.5 & - & 77.4 & 16.7 & - & - \\
	FastMCTS~\citep{li2025fastmcts} & 288K & Qwen2.5-7B~\citep{yang2024qwen25} & 88.9 & 74.0 & - & 20.0 & - & 27.5 \\
	KPDDS-7B~\citep{huang2024key} & 800K & Qwen2.5-Math-7B~\citep{yang2024qwen25math} & 89.9  & - & 76.0  & 10.0 & - & - \\
	DeepSeek-R1-Distill-Qwen-7B~\citep{guo2025deepseek} & 800K & Qwen2.5-7B-Instruct~\citep{yang2024qwen25} & 91.7 & - & 91.6 & 43.3 & - & - \\
	Openmathinstruct-7B~\citep{toshniwal2024openmathinstruct} & 14M & Qwen2.5-Math-7B~\citep{yang2024qwen25math} & 92.0  & - & 79.6  & 10.0 & - & - \\
	NuminaMath~\citep{ye2025limo} & 100K & Qwen2.5-Math-7B~\citep{yang2024qwen25math} & 92.9 & - & 81.8 & 20.0 & - & - \\
	PromptCoT-DS-7B~\citep{zhao2025promptcot} & 115K & DeepSeek-R1-Distill-Qwen-7B~\citep{guo2025deepseek} & 92.6 & - & 93.0  & 60.0 & - & - \\
	PromptCoT-Qwen-7B~\citep{zhao2025promptcot} & 905K & Qwen2.5-Math-7B~\citep{yang2024qwen25math} & 93.3 & - & 84.0  & 26.7 & - & - \\
	AceMath-7B-Instruct~\citep{liu2024acemath} & 1.2M & Qwen2-Math-7B-Instruct~\citep{yang2024qwen25math} & 93.7 & 83.1 & - & - & - & 42.2 \\
	AceMath-72B-Instruct~\citep{liu2024acemath} & 1.2M & Qwen2.5-Math-72B-Instruct~\citep{yang2024qwen25math} & 96.4 & 86.1 & - & - & - & 48.4 \\
	NuminaMath~\citep{ye2025limo} & 100K & Qwen2.5-32B-Instruct~\citep{yang2024qwen25} & - & - & 59.2 & 6.5 & 25.8 & 36.7 \\
	OpenThoughts~\citep{ye2025limo} & 114K & Qwen2.5-32B-Instruct~\citep{yang2024qwen25} & - & - & 80.6 & 50.2 & 42.9 & 56.3 \\
	Sky-T1-32B-Preview~\citep{team2025novasky} & 17K & Qwen2.5-32B-Instruct~\citep{yang2024qwen25} & - & - & 82.4 & 43.3 & 56.8 & - \\
	Journey Learning~\citep{huang2024o1} & 5K & Qwen2.5-Math-72B~\citep{yang2024qwen25math} & - & - & 87.2 & 43.3 & - & - \\
	STILL-2~\citep{min2024imitate} & 3.9K & Qwen2.5-32B-Instruct~\citep{yang2024qwen25} & - & - & 90.2 & 46.7 & 55.1 & - \\
	Bespoke-32B~\citep{bespokestratos} & 17K & Qwen2.5-32B-Instruct~\citep{yang2024qwen25} & - & - & 93.0 & 63.3 & 58.1 & - \\
	s1~\citep{muennighoff2025s1} & 1K & Qwen2.5-32B-Instruct~\citep{yang2024qwen25} & - & - & 93.0 & 56.7 & 59.6 & - \\
	DeepSeek-R1-Distill-Qwen-32B~\citep{guo2025deepseek} & 800K & Qwen2.5-32B-Instruct~\citep{yang2024qwen25} & - & - & 94.3 & 72.6 & 62.1 & -  \\
	LIMO~\citep{ye2025limo} & 817 & Qwen2.5-32B-Instruct~\citep{yang2024qwen25} & - & - & 94.8 & 15.8 & 66.7 & 66.8 \\
	
	\midrule
	\rowcolor{gray!8}\multicolumn{9}{c}{\textit{Deep Reasoning Self-Learning}}\\
	\midrule
	DPO~\citep{hwang-etal-2024-self} & 40K & DeepSeek-Math-7B-Base~\citep{shao2024deepseekmath} & 74.8 & 34.9 & - & - & - & - \\
	ReFT~\citep{hwang-etal-2024-self} & 40K & DeepSeek-Math-7B-Base~\citep{shao2024deepseekmath} & 71.4 & 36.0 & - & - & - & - \\
	Self-Explore~\citep{hwang-etal-2024-self} & 40K & DeepSeek-Math-7B-Base~\citep{shao2024deepseekmath} & 78.6 & 37.7 & - & - & - & - \\
	SimPO~\citep{reduce_overthinking_2025} & 10K & Qwen2.5-Math-7B-Instruct~\citep{yang2024qwen25math} & 88.8 & 40.0 & 56.6 & - & - & - \\
	DPO~\citep{liao2024tpo} & 11K & DeepSeek-Math-7B-Instruct~\citep{shao2024deepseekmath}  & - & 48.7 & - & - & - & - \\
	TPO~\citep{liao2024tpo} & 11K & DeepSeek-Math-7B-Instruct~\citep{shao2024deepseekmath}  & - & 51.3 & - & - & - & - \\
	DPO~\citep{liao2024tpo} & 11K & Qwen2-7B-Instruct~\citep{yang2024qwen2}  & - & 54.3 & - & - & - & - \\
	TPO~\citep{liao2024tpo} & 11K & Qwen2-7B-Instruct~\citep{yang2024qwen2}  & - & 55.5 & - & - & - & - \\
	MCTS~\citep{chen2024alphamath} & 15K & DeepSeek-Math-7B-Base~\citep{shao2024deepseekmath} & 83.2 & 64.0 & - & - & - & - \\
	SBS~\citep{chen2024alphamath} & 15K & DeepSeek-Math-7B-Base~\citep{shao2024deepseekmath} & 84.1 & 66.3 & - & - & - & - \\
	FastMCTS+Branch-DPO~\citep{li2025fastmcts} & 152K & FastMCTS-7B~\citep{li2025fastmcts} & 89.9 & 75.4 & - & 20.0 & - & 29.6 \\
	
	\bottomrule
	\end{tabular}
	
	}

	\caption{Performance of various deep reasoning learning methods, sorted primarily by Math or Math-500 scores. ``-'' indicates that the paper did not report this score.}
	\label{tab:deep-reasoning-learning}
\end{table*}

\subsubsection{Deep Reasoning Self-Learning}
\label{sec:deep-reasoning-self-learning}
While simple imitation can yield strong performance, current models still rely heavily on human annotations or outputs from more advanced models for both imitation and distillation~\citep{lobo2024impact}. To address this limitation, recent research has focused on enabling more advanced reasoning through techniques like self-play and self-learning~\citep{yang-etal-2024-weak,zhang2024openrft,li2025self,qin2025incentivizing}. Specifically, self-learning methods can be classified into two paradigms, differentiated by their sampling strategies:

(1) \textbf{\textit{Self-Learning from Direct Sampling:}}
The earliest method, STaR~\citep{zelikman2022star}, utilizes In-Context Learning (ICL) to sample deep reasoning results~\citep{shao2023synthetic} and uses the correctness of the final answer as an implicit reward for self-learning~\citep{hoffman2023training,pang2025bolt,pang2024iterative,zhang2025process,wang2024cream,liu-etal-2024-direct}. 
Further, ReST~\citep{gulcehre2023reinforced} extends this by introducing a Grow-Improve paradigm, where self-generated reasoning is first annotated with rewards and then enhanced via offline RL algorithms. However, these approaches can be fragile, especially when the reward process lacks robustness. Inspired by the Expectation-Maximization (EM) algorithm, \citet{singh2024beyond} propose a method that generates rewards and iteratively optimizes LLMs to achieve the best performance on a validation set, significantly improving robustness. To further strengthen the reward process, a series of work introduce a method to adapt incorrect solutions, training a verifier~\citep{dong2023raft,hosseini2024v} or utilize entropy~\citep{wang2025entropy,zhang2025entropy} to select or refine the reward process and improve self-learning quality.
(2) \textbf{\textit{Self-Learning from Tree Search:}}
Early deep learning methods, such as EXIT~\citep{anthony2017thinking}, combined MCTS with deep neural networks for reinforcement learning, iteratively self-training the network to guide the tree search and enhance reasoning. Building on this, CPO~\citep{zhang2024chain} and TPO~\citep{liao2024tpo} align each step of Long CoT reasoning with the corresponding tree search path, using Tree of Thoughts (ToT)~\citep{yao2023tree} preference information to support deeper reasoning~\citep{yang2024react, hwang-etal-2024-self}. \citet{li2025policy} propose Policy-Guided Tree Search (PGTS), integrating RL with structured tree exploration for more efficient navigation of reasoning paths. Further developments, such as AlphaMath~\citep{chen2024alphamath}, AlphaLLM-CPL~\citep{wang2024towards}, and TongGeometry~\citep{zhang2024proposing}, refine MCTS behavior through stepwise trajectory pair extraction and curriculum preference learning, boosting LLM reasoning abilities~\citep{puerto2024fine, li2025enhancing,xi2024training}.

\begin{TakeawayBox}{Takeaways: Imitation \& Self-Learning}
    \begin{itemize}[left=2pt,topsep=1pt,itemsep=2pt, parsep=1pt]
        \item Imitating deep reasoning from advanced RLLMs, and scaling-augmented methods like MCTS can help models learn complex reasoning patterns with fewer samples.
        \item Self-learning techniques, including reinforcement learning and tree search, allow RLLMs to enhance their reasoning abilities over time.
        \item The combination of imitation from advanced RLLMs and self-learning techniques strengthens RLLM reasoning, leading to strong performance on complex tasks.
    \end{itemize}
\end{TakeawayBox}
  \section{Feasible Reflection for Long CoT}
\label{sec:reflection}
Feasible Reflection is a pivotal component of Long CoT reasoning, enabling LLMs to handle complex tasks through iterative feedback and refinement~\citep{li2023reflection,gan2025rethinking}. Specifically, it comprises two primary stages: (1) Feedback (\S~\ref{sec:reflection-feedback}), which generates feedback signals $\mathcal{F}_i$ to correct node $n_j$ in Equation~\eqref{eq:feedback}; and (2) Refinement (\S~\ref{sec:reflection-refinement}), which adjusts the subsequent node $n_{i+1}$ according to the feedback in Equation \eqref{eq:refinement}.

\begin{figure*}[b]
	\centering
	\includegraphics[width=0.76\textwidth]{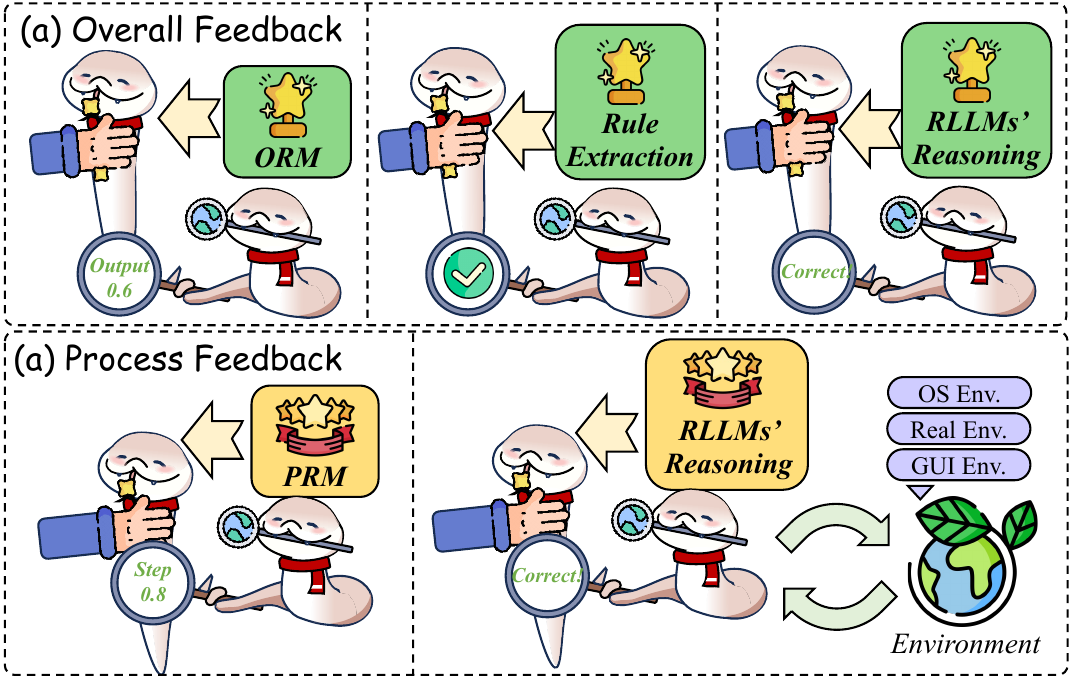}
	\caption{The feedback capabilities framework for feasible reflection consists of Overall Feedback and Process Feedback. Overall Feedback includes the Outcome Reward Model (ORM) in a value format, rule extraction for correctness judgment, and overall RLLMs based on RLLMs. Process Feedback includes the Process Reward Model (PRM) in a value format and step-level RLLMs, also based on RLLMs.	}
	\label{fig:feedback}
\end{figure*}

\subsection{Feedback}
\label{sec:reflection-feedback}
Feedback refers to the process of providing evaluations of both overall outputs and the processes that lead to them, with the goal of assessing their accuracy and quality~\citep{li2024generation,li2024llms,wei2025survey,guan2024search,wu2025sailing}. This process, also referred to as critique or verification, can be executed using either natural language or structured data formats, which serve as the foundation for tree-search methods~\citep{chen-etal-2024-tree}.
Specifically, as shown in Figure~\ref{fig:feedback}, feedback can be categorized into three distinct types:
(1) Overall Feedback (\S~\ref{sec:overall-feedback}); (2) Process Feedback (\S~\ref{sec:process-feedback}); (3) Hybrid Feedback (\S~\ref{sec:hybrid-feedback}).

\subsubsection{Overall Feedback}
\label{sec:overall-feedback}
The overall feedback focuses on providing a global view of the entire process and results, rather than assessing each step individually. This feedback significantly enhances reasoning skills and reward modeling in reinforcement learning for RLLMs. Specifically, as shown in Figure~\ref{fig:feedback} (a), the overall feedback can be categorized into three main sources: Outcome Reward Model, Rule Extraction, and RLLMs Feedback. The performance across these categories is summarized in Table~\ref{tab:feedback}.\vspace{-1mm}

\begin{table*}[t]
	\centering
	\resizebox{0.92\textwidth}{!}{
	\begin{tabular}{l|l|cccc|c}
    \toprule
	\textbf{Model} & \textbf{Base Model} & \textbf{Chat} & \textbf{Chat\_Hard} & \textbf{Safety} & \textbf{Reasoning} & \textbf{Overall} \\
    \midrule
	\rowcolor{gray!8}\multicolumn{7}{c}{\textit{RLLMs}}\\
	\midrule
GPT-4o-mini~\citep{achiam2023gpt} & - & 95.0 & 60.7 & 80.8 & 83.7 & 80.1 \\
Llama3.1-70B-Instruct~\citep{dubey2024llama} & - & 97.2 & 70.2 & 86.0 & 82.8 & 84.0 \\
Llama3.1-405B-Instruct~\citep{dubey2024llama} & - & 97.2 & 74.6 & 87.1 & 77.6 & 84.1 \\
GPT-4~\citep{achiam2023gpt} & - & 95.3 & 74.3 & 86.9 & 87.6 & 86.0 \\
GPT-4o~\citep{achiam2023gpt} & - & 96.1 & 76.1 & 86.6 & 88.1 & 86.7 \\
Gemini-1.5-pro~\citep{team2024gemini} & - & 92.3 & 80.6 & 87.9 & 92.0 & 88.2 \\
Self-taught Evaluator~\citep{wang2024selftaught} & Llama-3.1-70B-Instruct~\citep{dubey2024llama} & 96.6 & 84.2 & 81.0 & 91.5 & 88.3 \\
SFR-LLaMA-3.1-8B-Judge~\citep{wang2024direct} & Llama-3.1-70B-Instruct~\citep{dubey2024llama} & 95.5 & 77.7 & 86.2 & 95.1 & 88.7 \\
SFR-NeMo-12B-Judge~\citep{wang2024direct} & Mistral-NeMo-Instruct-12B~\citep{nvdia2024nemo} & 97.2 & 82.2 & 86.5 & 95.1 & 90.3 \\
SFR-LLaMA-3.1-70B-Judge~\citep{wang2024direct} & Llama-3.1-70B-Instruct~\citep{dubey2024llama} & 96.9 & 84.8 & 91.6 & \textbf{97.6} & 92.7 \\
Skywork-Critic-Llama-3.1-70B~\citep{wang2024direct} & Llama-3.1-70B-Instruct~\citep{dubey2024llama} & 96.6 & 87.9 & \textbf{93.1} & 95.5 & 93.3 \\
LMUnit~\citep{saad2024lmunit} & Llama-3.1-70B-Instruct~\citep{dubey2024llama} & - & - & - & - & 93.4 \\
EvalPlanner~\citep{saha2025learning} & Llama-3.1-70B-Instruct~\citep{dubey2024llama} & \textbf{97.5} & \textbf{89.4} & 93.0 & 95.5 & \textbf{93.9} \\
\midrule
	\rowcolor{gray!8}\multicolumn{7}{c}{\textit{Outcome Reward Models}}\\
\midrule
tulu-v2.5-13b-uf-rm~\citep{ivison2024unpacking} & TULU-2-13B~\citep{ivison2023camels} & 39.4 & 42.3 & 55.5 & 47.4 & 46.1 \\
Prometheus-2-7B~\citep{kim-etal-2024-prometheus} & Mistral-7B-Instruct-v0.2~\citep{jiang2023mistral7b} & 85.5 & 49.1 & 77.1 & 76.5 & 72.0 \\
Prometheus-8x7b-v2~\citep{kim-etal-2024-prometheus} & Mixtral-8x7B-Instruct~\citep{jiang2024mixtral} & 93.0 & 47.1 & 80.5 & 77.4 & 74.5 \\
Critic-RM-Rank~\citep{yu2024self} & Llama-3.1-70B-Instruct~\citep{dubey2024llama} & 97.0 & 58.0 & 84.0 & 92.0 & 82.8 \\
RM~\citep{stiennon2020learning} & Llama-3.1-70B-Instruct~\citep{dubey2024llama} & 98.3 & 74.5 & 83.8 & 88.0 & 86.4 \\
SynRM~\citep{ye2024improving} & Llama-3.1-70B-Instruct~\citep{dubey2024llama} & 97.5 & 76.8 & 86.3 & 88.5 & 87.3 \\
CLoud~\citep{ankner2024critiqueoutloud} & Llama-3-70B-Instruct~\citep{dubey2024llama} & 98.0 & 75.6 & 87.6 & 89.0 & 87.6 \\
FLAMe-RM-24B~\citep{vu-etal-2024-foundational} & PaLM-2-24B~\citep{anil2023palm} & 92.2 & 75.7 & 89.6 & 93.8 & 87.8 \\
SteerLM-RM 70B~\citep{wang2024helpsteer} & Llama-2-70B-chat~\cite{touvron2023llama} & 91.3 & 80.3 & 90.6 & 92.8 & 88.8 \\
Llama-3-OffsetBias-RM-8B~\citep{park2024offsetbias} & Llama-3-8B-Instruct~\citep{dubey2024llama} & 97.2 & 81.8 & 86.8 & 91.9 & 89.4 \\
InternLM-20B-Reward~\citep{cai2024internlm2} & InternLM2-8B-Instruct~\citep{cai2024internlm2} & \textbf{98.9} & 76.5 & 89.9 & 95.8 & 90.2 \\
ArmoRM-Llama3-8B-v0.1~\citep{wang-etal-2024-interpretable} & Llama-3-8B-Instruct~\citep{dubey2024llama} & 96.9 & 76.8 & 92.2 & \textbf{97.3} & 90.8 \\
Nemotron-4-340B-Reward~\citep{wang2024helpsteer} & Nemotron-4-340B~\citep{adler2024nemotron} & 95.8 & 87.1 & \textbf{92.2} & 93.6 & 92.2 \\
Skywork-Reward-Llama-3.1-8B~\citep{liu2024skywork} & Llama-3.1-70B-Instruct~\citep{dubey2024llama} & 95.8 & 87.3 & 90.6 & 96.2 & 92.5 \\
Skywork-Reward-Gemma-2-27B~\citep{liu2024skywork} & Gemma-2-27B-it~\citep{team2024gemma} & 95.8 & \textbf{91.4} & 92.0 & 96.1 & \textbf{93.8} \\
	\bottomrule
	\end{tabular}
	}

	\caption{Performance of various overall feedback methods, sorted primarily by Overall scores in RewardBench~\citep{lambert2024rewardbench}. ``-'' indicates that the paper did not report this score.}
	\label{tab:feedback}
\end{table*}

\paragraph{Overall Feedback from Outcome Reward Model} 
Since many tasks cannot be directly evaluated using accuracy or other standard metrics, research has increasingly focused on Outcome Reward Models (ORM), which provide value-based rewards for more general and quantifiable feedback~\citep{zhou2025reinforcing,yu2025rlpr,liu2025skywork}. 
In 2021, OpenAI~\citep{cobbe2021training} has proposed a ``Gen-Verifier'' paradigm, which uses a specialized ORM to evaluate the accuracy of generated rationales, showing significant progress in feedback capabilities~\citep{shao2024deepseekmath}.
\citet{ji-etal-2023-towards} introduce a trained knowledge scorer to analyze hallucinations in the reasoning process, providing feedback to RLLMs and improving the accuracy of their outputs over time. Moreover, Generative Reward Models~\citep{zhang2024generative} use next-token prediction for overall feedback, which seamlessly integrates with instruction adjustments, leveraging inference-time calculations to improve ORM feedback.

However, specifically trained ORMs are often costly and not sufficiently robust. Building on this, Self-Rewarding Language Models (SRLMs)~\citep{zhou2025self} incorporate a self-consistency framework, optimizing feedback to improve model alignment and consistency~\citep{zhang2025consistent}. \citet{yu2024self} introduce Critic-RM, combining RLLM-generated natural language criticism with corresponding feedback. This method filters high-quality feedback while jointly fine-tuning reward prediction and criticism generation, optimizing ORM performance.\vspace{-1mm}

\paragraph{Overall Feedback from Rule Extraction}
Although ORM has achieved significant improvements, its accuracy still falls short of 100\%, preventing it from outperforming rule-based answer correction feedback~\citep{yao2023tree,hao-etal-2023-reasoning,zhang2025reinforcement}. Previous studies, such as STaR~\citep{zelikman2022star}, ReST~\citep{gulcehre2023reinforced}, and ReFT~\citep{trung-etal-2024-reft}, have demonstrated that feedback based on final answer rewards is more effective than both PRM and ORM in mathematical scenarios~\citep{gao2024designing}.
Furthermore, \citet{guo2025deepseek} and \citet{xie2025logic} introduce a multi-stage RL framework that incorporates rule-based rewards, significantly enhancing both output accuracy and length while mitigating reward hacking through simple yet robust rules~\citep{baker2025monitoring}, such as format validation and result verification. In coding scenarios where direct rule-based feedback is difficult, OpenCodeInterpreter~\citep{zheng-etal-2024-opencodeinterpreter}, AceCoder~\citep{zeng2025acecoder}, O1-Coder~\citep{zhang2024o1}, and VerMCTS~\citep{brandfonbrener2024vermcts} address this challenge by implementing an automated test-case synthesis pipeline, deriving rewards based on program performance~\citep{ni2023lever,gou2023critic,zhou2024solving}. Additionally, \citet{ma2025dynamic} propose an automated approach to training a test case generator, which alleviates the scarcity of test cases and demonstrates that increasing the number of test cases correlates with improved reward quality. Moreover, \citet{ma2025sorft} decompose problem-solving into structured coding subtasks: file localization, function localization, line localization, and code editing generation, and applies multi-viewed rule-based rewards.\vspace{-1mm}

\paragraph{Overall Feedback from RLLMs}
Research on feedback from RLLMs centers on detecting errors and biases through natural language feedback, also known as LLM-as-Judge, self-reflection or self-critique~\citep{hu2025training,kadavath2022language,bai2022constitutional,renze2024self,miao2024selfcheck,wang2023shepherd,yuan2025study,xu2025j4r,ma2025s}. This method has led to significant improvements across various tasks, particularly in self-correction~\citep{weng-etal-2023-large,zheng2024critic,gero2023selfverification,fernando24promptbreeder,zhang-etal-2024-small}. \citet{huang2024large} contend that traditional LLMs struggle to generate effective feedback without external signals, requiring the development of RLLMs with enhanced feedback capabilities~\citep{saunders2022self,li2025extended}. As a result, many studies leverage RLLMs' error-identification strengths, often stemming from their pretraining phase, to improve feedback generation and correction~\citep{ye2025physics,bao2023contrastive,bao-etal-2024-abstract,huang2025think}.

Earlier, \citet{mcaleese2024llm} found that training RLLMs to learn self-critique and deep reasoning can further boost performance. \citet{zhang-etal-2024-self-contrast} propose a self-contrast mechanism that compares multiple perspectives, identifies differences, and summarizes insights to resolve inconsistencies. However, these methods often offer task-independent feedback. To address this, \citet{hao2024llm} introduce AutoRace, which tailors evaluation criteria for specific tasks. The Reversal of Thought (RoT) framework~\citep{yuan2024reversal} introduces a novel paradigm combining reverse reasoning with self-reflection, helping models identify the limits of their knowledge and enhance reasoning efficiency. Furthermore, ACR~\citep{zhou2025refinecoder} implements a scoring system for coding tasks, using LLM-as-a-Judge for quality assessment and LLM-as-a-Critic for critiquing low-quality code, improving consistency across benchmarks. \citet{zheng2025what} integrate code execution error data and feedback from RLLMs to improve code generation performance. \citet{liu2025attention} present AGSER, a method using attention-guided self-reflection to address hallucinations by splitting input queries into attentive and non-attentive components. Finally, \citet{saha2025learning} introduce EvalPlanner, which separates feedback into planning and reasoning components for more streamlined expression using existing RLLMs.
More comprehensively, \citet{hu2025training} outline the complete pipeline, key insights, and practical lessons for training RLLMs to function as judges.
\vspace{-1mm}

\subsubsection{Process Feedback}
\label{sec:process-feedback}
Techniques combine process feedback with MCTS or RL rewards to provide automated, step-by-step guidance, reducing the need for labor-intensive annotations while enhancing reasoning capabilities~\citep{uesato2022solving,kazemnejad2024vineppo}. These techniques can be categorized into two main types based on the source of feedback: process reward models (PRMs) and prompted LLMs. The performance comparison are mainly shown in Table~\ref{tab:process-feedback}.\vspace{-1mm}

\paragraph{Process Feedback from Process Rewarded Model}
Recent studies highlight the significance of feedback in developing effective PRMs for complex reasoning tasks, particularly in a step-level view~\citep{choudhury2025process,li2024hunyuanprover,ma2023let}. (1) \textbf{\textit{Process Annotated PRM Training:}} Earlier, \citet{lightman2023let} demonstrate that training process feedback with human-annotated data (PRM800K) surpasses outcome supervision in creating reliable reward models. However, this approach requires significant human effort. To address this, \citet{wang2024math} introduce Math-Shepherd, a dataset that generates step-by-step supervision using a Tree Search-inspired method~\citep{chen-etal-2024-step,yuan2025advancing}. Following this, methods like QwQ~\citep{team2025qwq}, Skywork-o1~\citep{team2025skywork}, AceMath~\citep{liu2024acemath}, and PRIME~\citep{cui2025process} adopt similar techniques to enhance PRM performance. Additionally, \citet{zhang2024entropy} propose entropy regularization to improve model convergence.
Rather than focusing solely on the first error step, Full-Step-DPO~\citep{xu2025full} assigns rewards for the entire reasoning chain, including error steps. VersaPRM~\citep{zeng2025versaprm} extends PRMs across multiple domains, broadening their applicability. Similarly, \citet{gu2025capturing} and \citet{zhang2025distill} suggest training models with student preferences aligned to teacher preferences, ensuring effective preference distillation. Further, \citet{wang2025visualprm} propose VisualPRM400K and expand this paradigm to multimodal scenarios.
(2) \textbf{\textit{Outcome Annotated PRM Training:}} Alternative approaches, such as ReST-MCTS*~\citep{zhang2024restmcts}, OVM~\citep{yu2024ovm}, Implicit PRM~\citep{yuan2024free}, AutoPSV~\citep{lu2024auto}, and DVO~\citep{zhang2025direct}, leverage outcome supervision or implicit feedback to train PRMs, reducing the need for extensive human-annotated data~\citep{xie2025teaching,saha2025learning}.
UAS~\citep{yu2025uncertainty} incorporates uncertainty-aware value models~\citep{hu2024uncertainty} into feedback predictions~\citep{liu2025adaptivestep,duan2025efficient,yang2025uncertainty,zhao2025learning}. Additionally, Aurora~\citep{tan2025aurora} utilizes ensemble prompting strategies and reference answers for reverse verification, training stronger PRMs that better align with the Long CoT data distribution. Furthermore, PAV~\citep{setlur2025rewarding} suggests that rewards should reflect reasoning progress, as measured by changes in the likelihood of producing a correct future response before and after each step. \citet{yang2024selective,lee2024token,yoon2024tlcr} extend these paradigms to the token level. Moreover, \citet{chen2025scaling} expand these into interactive agent scenarios, allowing for automatically learning reward models from the environment without additional manual annotation. \citet{wang2025test} equip a dual-layer MLP module to evaluate the reward at each step, successfully integrating the policy model and PRM into a unified interface without additional process annotations, reducing over 99\% of PRM parameters for efficient reasoning.

\begin{table*}[t]
	\centering
	\resizebox{\textwidth}{!}{
	\begin{tabular}{l|l|cccc|ccc}
    \toprule
	\multicolumn{1}{c}{} & & \multicolumn{4}{c}{\textit{ProcessBench}} & \multicolumn{3}{|c}{\textit{PRMBench}} \\
	\midrule
	\multicolumn{1}{c}{} & & GSM8K & MATH & OlympiadBench & OmniMATH & Simplicity & Soundness & Sensitivity \\
    \midrule
	\rowcolor{gray!8}\multicolumn{9}{c}{\textit{Process Reward Models}}\\
	\midrule
	Qwen2.5-Math-7B-PRM~\citep{zheng2024processbench} & Qwen2.5-Math-7B~\citep{yang2024qwen25math} & 39.4 & 52.2 & 39.4 & 33.1 & - & - & - \\
Math-Shepherd-PRM-7B~\citep{wang2024math} & Mistral-7B~\citep{jiang2023mistral7b} & 47.9 & 29.5 & 24.8 & 23.8 & 47.1 & 45.7 & 60.7 \\
RLHFlow-PRM-Mistral-8B~\citep{dong2024rlhf} & Mistral-7B~\citep{jiang2023mistral7b} & 50.4 & 33.4 & 13.8 & 15.8 & 46.7 & 57.5 & 68.5 \\
RLHFlow-PRM-DeepSeek-8B~\citep{dong2024rlhf} & DeepSeek-7B~\citep{bi2024deepseek} & 38.8 & 33.8 & 16.9 & 16.9 & 47.6 & 57.5 & 68.1 \\
Skywork-PRM-1.5B~\citep{liu2024skywork} & Qwen2.5-Math-1.5B-Instruct~\citep{yang2024qwen25} & 59.0 & 48.0 & 19.3 & 19.2 & 33.6 & 28.6 & 48.8 \\
Skywork-PRM-7B~\citep{liu2024skywork} & Qwen2.5-Math-7B-Instruct~\citep{yang2024qwen25} & 70.8 & 53.6 & 22.9 & 21.0 & 38.4 & 32.7 & 54.3 \\
Qwen2-1.5B-PRM800k~\citep{sun2025efficient} & Qwen2-Math-1.5B-Instruct~\citep{yang2024qwen25math} & 34.0 & 55.3 & 34.2 & 41.0 & - & - & - \\
Qwen2-1.5B-Math-Shepherd~\citep{sun2025efficient} & Qwen2-Math-1.5B-Instruct~\citep{yang2024qwen25math} & 48.9 & 34.1 & 9.8 & 13.7 & - & - & - \\
Qwen2-1.5B-Epic50k~\citep{sun2025efficient} & Qwen2-Math-1.5B-Instruct~\citep{yang2024qwen25math} & 55.6 & 36.1 & 20.2 & 30.0 & - & - & - \\
Qwen2.5-Math-7B-PRM800K & Qwen2.5-Math-7B-Instruct~\citep{yang2024qwen25math} & 68.2 & 62.6 & 50.7 & 44.3 & - & - & - \\
Qwen2.5-Math-PRM-7B~\citep{zheng2024processbench} & Qwen2.5-Math-7B-Instruct~\citep{yang2024qwen25math} & 82.4 & 77.6 & 67.5 & 66.3 & - & - & - \\
Universal-PRM-7B~\citep{tan2025aurora} &  Qwen2.5-Math-7B-Instruct~\citep{yang2024qwen25math} & 85.8 & 77.7 & 67.6 & 66.4 & - & - & - \\
\midrule
	\rowcolor{gray!8}\multicolumn{9}{c}{\textit{Critic Model}}\\
\midrule
Llama-3.1-8B-Instruct~\citep{dubey2024llama} & - & 27.5 & 26.7 & 18.5 & 19.2 & - & - & - \\
GPT-4o~\citep{achiam2023gpt} & - & 61.9 & 53.9 & 48.3 & 44.6 & 59.7 & 70.9 & 75.8 \\
QwQ-32B-Preview~\citep{team2025qwq} & Qwen2.5-32B-Instruct~\citep{yang2024qwen25} & 62.3 & 52.7 & 46.2 & 43.9 & - & - & - \\
DeepSeek-R1-Distill-Qwen-14B~\citep{guo2025deepseek} & Qwen2.5-14B-Instruct~\citep{yang2024qwen25} & 67.3 & 38.8 & 29.9 & 32.1 & - & - & - \\
Dyve-14B~\citep{zhong2025dyve} & DeepSeek-R1-Distill-Qwen-14B~\citep{guo2025deepseek} & 68.5 & 58.3 & 49.0 & 47.2 & - & - & - \\
Qwen2.5-72B-Instruct~\citep{yang2024qwen25} & - & 76.2 & 61.8 & 54.6 & 52.2 & - & - & - \\
SCRIT~\citep{tang2025enabling} & Qwen2.5-72B-Instruct~\citep{yang2024qwen25} & 80.2 & 60.0 & 32.5 & 27.8 & - & - & - \\
o1-mini~\citep{jaech2024openai} & - & 93.2 & 88.9 & 87.2 & 82.4 & 64.6 & 72.1 & 75.5 \\
Llemma-PRM800k-7B~\citep{song2025prmbench} & Llemma-7B~\citep{azerbayev2024llemma} & - & - & - & - & 51.4 & 50.9 & 66.0 \\
Llemma-MetaMath-7B~\citep{song2025prmbench} & Llemma-7B~\citep{azerbayev2024llemma} & - & - & - & - & 50.3 & 49.0 & 66.0 \\
Llemma-oprm-7B~\citep{song2025prmbench} & Llemma-7B~\citep{azerbayev2024llemma} & - & - & - & - & 49.0 & 49.8 & 64.1 \\
MATHMinos-Mistral-7B~\citep{gao2024reasongoodbadbetter} & Mistral-7B~\cite{jiang2023mistral7b} & - & - & - & - & 51.4 & 54.4 & 66.5 \\
ReasonEval-7B~\citep{xia2024evaluating} & Llemma-7B~\citep{azerbayev2024llemma} & - & - & - & - & 55.5 & 63.9 & 71.0 \\
ReasonEval-34B~\citep{xia2024evaluating} & Llemma-34B~\citep{azerbayev2024llemma} & - & - & - & - & 51.5 & 63.0 & 73.1 \\
Gemini-2.0-flash-exp~\citep{song2025prmbench} & - & - & - & - & - & 62.7 & 67.3 & 75.4 \\
Gemini-2.0-thinking-exp-1219~\citep{song2025prmbench} & - & - & - & - & - & 66.2 & 71.8 & 75.3 \\
	\bottomrule
	\end{tabular}
	}

	\caption{Performance of various process feedback methods on ProcessBench~\citep{zheng2024processbench} and PRMBench~\citep{song2025prmbench}. ``-'' indicates that the paper did not report this score.}
	\label{tab:process-feedback}
\end{table*}

\paragraph{Process Feedback from RLLMs}
As PRM training remains heavily dependent on manually annotated data, recent research has explored methods for enabling models to generate their natural language feedback to optimize performance~\citep{xu2024can}. These approaches fall into two primary categories:
(1) \textit{\textbf{Model-Driven Feedback Reasoning:}}
Earlier work such as React~\citep{yao2023react} and Reflexion~\citep{shinn2023reflexion} enhances RLLMs with natural language feedback at each action and reasoning step~\citep{gao2024llm,chowdhury2025zero,chen2025judgelrm}, improving decision-making in diverse tasks. Similarly, Step-DPO~\citep{lai2024step} uses RLLM to self-verify step-level positive and negative pairs for training through the DPO paradigm, achieving strong performance. Additionally, \citet{sun2025error} propose a dynamic error classification framework that adapts based on model outputs, improving performance in mathematical reasoning tasks by addressing specific error patterns in math word problems. Furthermore, \citet{xie2024monte} and \citet{he-etal-2024-advancing} iteratively apply MCTS to collect preference data, utilizing its forward-looking capabilities to decompose instance-level rewards into more precise step-level signals, thereby enhancing feedback accuracy. However, step-wise feedback often suffers from reliability issues, which can be mitigated by uncertainty quantification~\citep{yin-etal-2024-reasoning, ye2025uncertainty}, improving the reliability of step-wise verification in reward models for mathematical reasoning tasks. Moreover, \citet{fu2025unveiling} define the CoT Average Causal Effect (CACE) to capture causal relationships between steps, resulting in a causalized Long CoT where all steps are both correct and comprehensible.
(2) \textit{\textbf{Environment-Driven Feedback Reasoning:}}
Given the increasing complexity of large models, there is growing interest in combining prompt-based LLMs with external environments to generate more interpretable and controllable feedback~\citep{xie2025robotic,hu2024agentgen}. For example, ORPS~\citep{yu2024outcome} and \citet{drori2025diverse} minimize dependence on human annotations by using execution feedback, enabling models to autonomously refine their solutions. Additionally, \citet{shrestha2025mathematical} contribute by translating model outputs into Python code, helping to identify logical errors, gain insights into flawed reasoning processes, and guide improvements in mathematical reasoning. \citet{xu2024interactive} integrate reasoning models with an interactive environment, enabling learning in more dynamic scenarios and creating a more generalizable self-learning framework.

\subsubsection{Hybrid Feedbacks}
\label{sec:hybrid-feedback}
Given the respective advantages and limitations of Overall Feedback and Process Feedback, recent studies have sought to combine both for optimal feedback.
Specifically, \citet{zhang2025lessons} propose a consensus filtering mechanism that integrates Monte Carlo estimation with an LLM-as-judge to enhance both overall and stepwise feedback, thus improving reasoning accuracy.  In a similar vein, \citet{lin2025step} introduce Step-KTO, a framework combining stepwise process-level and outcome-level binary feedback, using PRM and ORM to guide language models toward coherent reasoning, with a focus on error correction through reflection mechanisms.

\begin{TakeawayBox}{Takeaways: Feedback}
    \begin{itemize}[left=2pt,topsep=1pt,itemsep=2pt, parsep=1pt]
        \item \textbf{Evolving Feedback Models:}  Feedback mechanisms, including overall, process, and hybrid feedback, are crucial for improving the reasoning capabilities of RLLMs.
        \item \textbf{Innovative Approaches in Process Feedback:} Process feedback using techniques like PRMs with MCTS enhances Long CoT, though challenges like reward hacking remain.
        \item \textbf{Self-Reflection and Model-Driven Feedback:} Self-reflection and model-driven feedback improve RLLM performance by enabling error detection, task-specific insights, and more autonomous learning.
    \end{itemize}
\end{TakeawayBox}

\subsection{Refinement}
\begin{figure*}[b]
	\centering
	\includegraphics[width=0.98\textwidth]{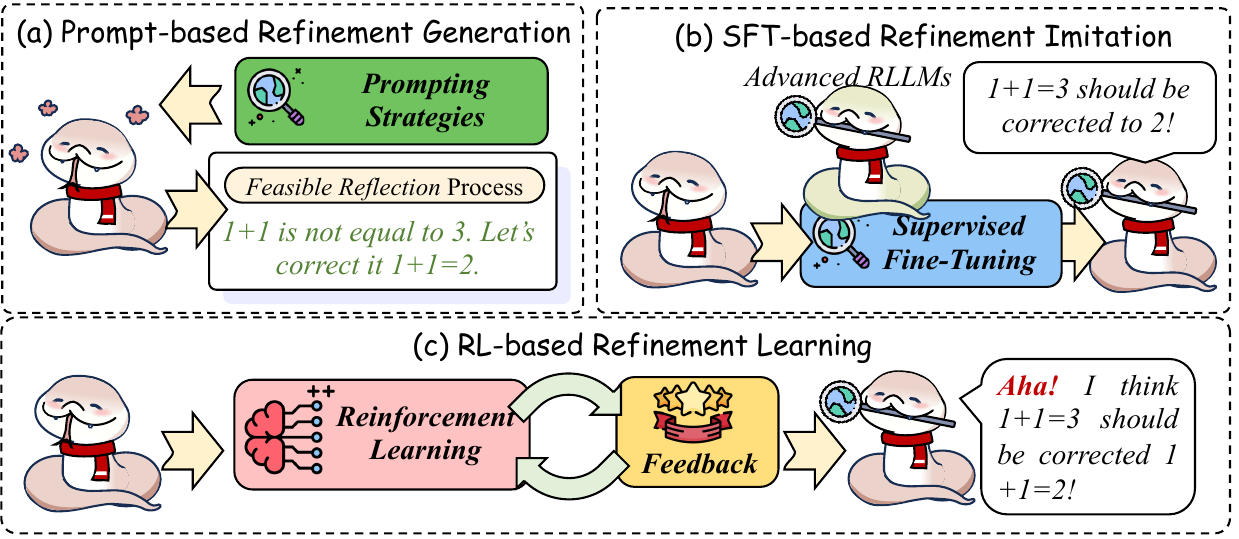}
	\caption{The three main categories of refinement methods, including Prompt-based Refinement Generation, SFT-based Refinement Imitation, and RL-based Refinement Learning.}
	\label{fig:refinement}
\end{figure*}

\vspace{-1mm}
\label{sec:reflection-refinement}
Refinement refers to the process of addressing errors in reasoning based on prior feedback. As shown in Figure~\ref{fig:refinement}, refinement methods can be grouped into three primary categories: prompt-based refinement generation (\S~\ref{sec:prompt-based-refinement-generation}), SFT-based refinement imitation (\S~\ref{sec:sft-based-refinement-imitation}), and RL-based refinement learning (\S~\ref{sec:rl-based-refinement-learning}).

\subsubsection{Prompt-based Refinement Generation}
\label{sec:prompt-based-refinement-generation}
Research on prompt-based refine generation focuses on enhancing the performance of LLMs through iterative self-refinement mechanisms~\citep{pan2023automatically,zhao-etal-2024-enhancing-zero,chen2024adaptive,liu2024large,zhang2024learning,wan2024cot,wang2024a,mei2025survey}.
A prominent approach involves prompting RLLMs to generate initial outputs, followed by self-feedback that iteratively refines and improves performance across tasks such as dialogue generation and mathematical reasoning~\citep{saunders2022self,madaan2023self,zheng2024progressivehint,shinn2023reflexion,miao2024selfcheck,khalifa2023grace,vacareanu2024general,liu2025instruct}, which even much reduce the hallucinations~\citep{huang-etal-2024-advancing,ji-etal-2023-towards}. Noteworthy methods, like Self-Backtracking~\citep{yang2025step}, Refiner~\citep{paul-etal-2024-refiner}, and BackMath~\citep{zhang-xiong-2025-backmath}, allow LLMs to adjust their reasoning autonomously, reducing unnecessary complexity in decision-making~\citep{wu2024enhancing}. Further, \citet{havrilla2024glore} extend the paradigm by integrating overall-level and step-level refinements, improving refinement performance. \citet{yang2024confidence} propose a method to decompose the self-correction capability of LLMs into "confidence" and "critique" capacities, designing probabilistic metrics to evaluate them and exploring the role of reflection mechanisms in model behavior. Additionally, MCTSr~\citep{zhang2024accessing}, LLM2~\citep{yang2024llm2}, ReST-MCTS*~\citep{zhang2024restmcts} and ReARTeR~\citep{sun2025rearter} emphasize dynamic reflection through iterative error correction and confidence adjustments, allowing models to autonomously refine reasoning strategies~\citep{ferraz2024llm}. \citet{he2024enhancing} extend this paradigm to multi-agent scenarios, improving both reasoning and agent system performance~\citep{yang2025lighthouse,zhou2025debate}. Moreover, \citet{yuksekgonul2025optimizing} and \citet{peng2025dlpo} further expand the paradigm by enabling automatic prompt optimization driven by LLMs. This approach facilitates more generalized and automated refinement of input prompts across a range of tasks, as opposed to focusing solely on refining output results.
However, without oracle feedback, RLLM’s self-refinement process fails, causing instability in both intermediate and final answers, leading to biases in simple factual queries and introducing cognitive biases in complex tasks~\citep{zhang2024understanding,xu-etal-2024-pride}.

\subsubsection{SFT-based Refinement Imitation}
\label{sec:sft-based-refinement-imitation}
Recent advancements in reflection-based reasoning for LLMs have led to frameworks that enhance model reasoning through self-refinement and error correction. A key approach is directly supervised fine-tuning, which allows models to learn error correction processes from advanced LLMs, thereby improving their reflective capabilities~\citep{an2023learning,chen2024teaching,li2023reflection,wang2025critique,chen2025iterative,xi2024enhancing}. 
Notable frameworks, such as rStar~\citep{qi2024mutual}, improve smaller language models through self-play mutual reasoning, while Recursive Introduction~\citep{qu2024recursive} and RealCritic~\citep{tang2025realcritic} use iterative feedback mechanisms to identify and correct errors to better self-improve~\citep{li2025llms}. \citet{yan2024s} propose constructing step-wise self-correction data and implementing a training strategy that uses the above-constructed data to equip LLMs with spontaneous step-level self-correction capacities.
Building upon these, \citet{gao2024llm} and \citet{zhang2025cot} propose Math-Minos, which employs step-by-step natural language feedback as rationale tags, offering both correctness and detailed explanations for each step to train feedback mechanisms that justify and refine the reasoning process. Journey Learning~\citep{qin2024o1} employs MCTS to parse node backtracking as natural language refinement, enhancing supervised fine-tuning and, thereby, improving reasoning performance.
Additionally, approaches like ProgCo~\citep{song2025progco} emphasize iterative feedback and program-driven refinement to enhance critique and self-correction. Expanding these ideas to multimodal settings, frameworks, such as R3V~\citep{cheng2024vision} and MM-Verify~\citep{sun2025mm}, focus on integrating visual and textual reasoning~\citep{luo2025ursa,wang2024enhancingvisual}.

\subsubsection{RL-based Refinement Learning}
\label{sec:rl-based-refinement-learning}
In recent research, several approaches have been proposed to enhance the performance of refinement  through reinforcement learning~\citep{silver2025language,zhang2025beyond}. Earlier, \citet{kumar2024training} observed that SFT of RLLMs often fails to promote self-refinement behaviors. This limitation stems from a distributional mismatch between data collection strategies and model responses, as well as the risk of behavioral collapse. To address this, SCoRe~\citep{kumar2024training} enhances self-refinement by training the model on its own self-generated correction trajectories and employing regularization to guide the learning process. This method prioritizes fostering self-refinement during testing, rather than merely maximizing reward for specific prompts~\citep{zeng2025aries}.
Further, \citet{guo2025deepseek} demonstrate that applying outcome-level rewarded RL can trigger an ``Aha moment,'' activating the model's natural feedback and refinement behaviors without the need for human guidance. Moreover, \citet{guo2025deepseek, zeng2025simplerl} and \citet{ma2025s} explore initializing LLMs with iterative self-verification and self-correction behaviors, which are strengthened through supervised fine-tuning and further enhanced by outcome-level RL. \citet{ma2025s} and \citet{yang2025reasonflux} extend these capabilities with process-level RL, minimizing resource usage while enabling adaptive reasoning refinements during inference. More recently, \citet{lee2025revise} introduce an intrinsic verifier module to decide when refinements should be applied, using RL to further encourage self-refinement when errors are detected.

\begin{TakeawayBox}{Takeaways: Refinement}
    \begin{itemize}[left=2pt,topsep=1pt,itemsep=2pt, parsep=1pt]
        \item \textbf{Prompt-Based Refinement for Iterative Improvement:} Iterative self-refinement through feedback loops helps LLMs improve reasoning and reduce errors like hallucinations but requires stable feedback to maintain accuracy.
        \item \textbf{Supervised Fine-Tuning (SFT) for Error Correction:} Supervised fine-tuning enhances LLMs by using iterative feedback and self-correction strategies to improve reasoning accuracy, especially for smaller models.
        \item \textbf{Reinforcement Learning (RL) for Refinement:} Reinforcement learning enhances self-refinement in LLMs by using self-generated corrections and adaptive strategies, reducing human intervention and resource consumption.
    \end{itemize}
\end{TakeawayBox}
	\section{Extensive Exploration for Long CoT}
\label{sec:exploration}
Exploration is a key capability in Long CoT reasoning, allowing models to navigate complex problem spaces through strategic branching and iterative refinement~\citep{zeng2024scaling,lerer2020improving,wang2024onplanning,valmeekam2024llms}. Recent studies emphasize exploration mechanisms, such as hypothesis branching and error backtracking via reflection, as essential for overcoming the constraints of linear reasoning paths~\citep{guo2025deepseek}.

Current research focuses on three key areas: (1) \textit{\textbf{Exploration Scaling}} (\S~\ref{sec:exploration-scaling}), which explores the breadth and depth of exploration and its impact on downstream applications, particularly in improving the size of the exploration path $m$ in Equation~\eqref{eq:exploration}; (2) \textit{\textbf{Internal Exploration}} (\S~\ref{sec:exploration-internal}), which focuses on training models to develop internal exploration capabilities, enabling more efficient and effective generation of $m$ exploration paths $\{n_{i+j}\}^{m}_{j=1}$ in Equation~\eqref{eq:exploration}; and (3) \textit{\textbf{External Exploration}} (\S~\ref{sec:exploration-external}), which examines how models can leverage external systems to enhance their exploratory abilities, facilitating the selection of the most effective path $n_{i+j}$ from the $m$ exploration paths in Equation~\eqref{eq:exploration}.

\begin{figure*}[t]
	\centering
	\includegraphics[width=0.99\textwidth]{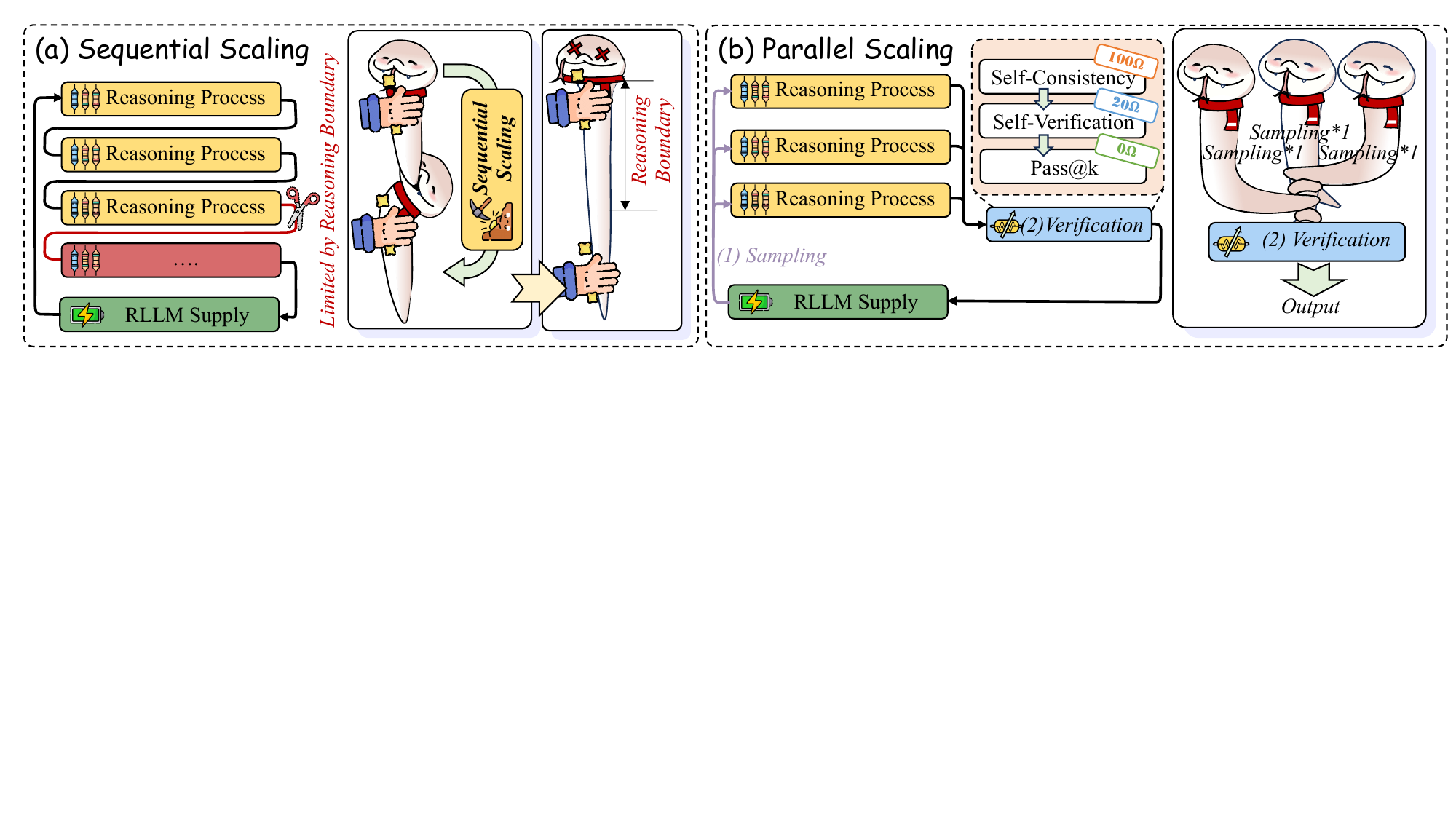}
	\caption{Schematic representations of two common inference-time scaling strategies: (a) sequential scaling, which extends the length of Long CoT but is constrained by the reasoning boundaries of RLLMs; and (b) parallel scaling, which increases the sample size and aggregates multiple outcomes, yet does not surpass the performance of Pass@k.}
	\label{fig:exploration-scaling}
\end{figure*}

\subsection{Exploration Scaling}
\label{sec:exploration-scaling}
Recent advances in inference-time scaling algorithms~\citep{jones2021scaling,welleck2024from,brown2024large,zhang2025survey,chen2025survey} have attracted significant interest, particularly in scaling reasoning length to improve performance~\citep{pmlr-v124-lyzhov20a,nori2024medprompt,li2025mis,wang2025m1}. Following \citet{chen2025ecm}, as shown in Figure~\ref{fig:exploration-scaling}, exploration scaling can be understood through two paradigms: (1) \textit{\textbf{sequential scaling}}, akin to a series of resistors, which connects multiple reasoning processes using reflection; and \textit{\textbf{parallel scaling}}, similar to parallel resistors, where a unified verification/feedback mechanism selects the most effective reasoning processes.

\subsubsection{Sequential Scaling}
Sequential scaling refers to extending the reasoning output within a single model generation, significantly boosting model performance~\citep{li2025metal,zhang2025and,kim2025cost}. Early works by \citet{fu2023complexitybased} and \citet{jaech2024openai} show that increasing the length of the reasoning path can greatly improve performance.
\citet{tian2025think} enhances model reasoning iteratively by using prior answers as prompts for each successive round, thus enabling sequential scaling of the reasoning process.
Building on this, later studies~\citep{ji2025test,li2025reasoning} further explore enhancing logical depth through tree-based searches within a fixed compute budget, resulting in notable performance gains~\citep{alomrani2025reasoning,qi2025optimizing}.
Building upon this, \citet{muennighoff2025s1} introduce a inference-time scaling method that improves reasoning by fine-tuning and budget forcing, yielding substantial gains with additional computing at inference time.
To address the constraints of attention spans, some studies focus on expanding reasoning length in latent spaces. \citet{geiping2025scaling} and \citet{chen2025inner} enhance inference-time reasoning performance by implicitly scaling computation in latent space through recurrent depth.
\citet{setlur2025e3} identified three core aspects of sequential scaling: (1) linking skills to asymmetric capabilities in base LLMs, such as connecting easy verification with difficult exploration; (2) enhancing exploration in reinforcement learning by utilizing the ``negative'' gradient of error trajectories, which extends search paths and links additional asymmetries; and (3) creating dynamic exploration by aligning task difficulty with training token budgets through tailored curricula.

\subsubsection{Parallel Scaling}
Parallel scaling refers to the process of increasing the number of reasoning iterations during model generation and then verfiy these results to get the final output, which significantly enhances model performance~\citep{abdelhameed2024inference,wu2024inference,brown2024large,liu2025metascale,byun2025test,zhu2025scaling}. Initially, \citet{wang2023selfconsistency} introduce the concept of self-consistency, demonstrating that multiple sampling processes followed by majority voting for effective exploration. \vspace{-1mm}

\paragraph{Verification Optimization}
The primary focus of recent research is optimizing verification, which can be categorized into two types:
(1) \textit{\textbf{Overall Verification:}} Recent works~\citep{zhou2024dont,wang2024multi} divide the scaling process into two stages: "reasoning" and "self-verification." By replacing majority voting in self-consistency with self-verification, these approaches show significant improvements~\citep{zhao2025sample,chen2025sets,zou2025testnuc,lai2025multidimensional,li2025revisiting}.
In code scenarios, WoT~\citep{zhang-etal-2024-wrong}, CISC~\citep{taubenfeld2025confidence} and S*~\citep{li2025s} scale the Long CoT in parallel, using output confidence or code execution results for verification, effectively assessing reasoning quality~\citep{raza2025instantiation,gehring2024rlef,huang2025efficient,zhou2025bridging}. Further, \citet{nye2022show} and \citet{weir2024learning,stoisser2025sparks} train RLLMs to simulate code execution, removing the need for test cases in code-related parallel scaling.
Chain-of-Verification~\citep{chen2025ecm} introduces meta-verification, sampling multiple verification instances to identify the correct one. \citet{kim2024lachesis}, \citet{chen2024essential}, and \citet{vacareanu2024general} validate this approach empirically by evaluating answer correctness based on reasoning path properties. Moreover, \citet{li2025drafts} tune a specific RLLM to verify and aggregate answers, showing improved performance. This suggests that PRM cannot replace a specially trained RLLM for verification due to training goal biases~\citep{zhang2025lessons}. Finally, \citet{kang2025scalable} leverage self-uncertainty to select the best results.
(2) \textit{\textbf{Step Verification:}}
Building on this, numerous researchers have explored step-level or finer-grained verification~\citep{chen2024magicore,ling2023deductive}. Notably, DIVERSE~\citep{li-etal-2023-making}, SSC-CoT~\citep{zhao2024stepwise}, and Fine-grained Self-Consistency~\citep{chen2025ecm} combine diverse reasoning paths with step-level verification. 
In addition, a series of works~\citep{snell2024scaling,wu2024inference,luo2024improve,wang2024seed,wu2024beyond,liu2025can} try to investigate how optimal scaling strategies based on MCTS can enhance smaller language models' performance. Their findings show that a 1B RLLM can outperform a 405B model on complex tasks through parallel scaling~\citep{yu2025exact}. Despite these advancements in verification, \citet{chen2025ecm} demonstrate that these strategies cannot surpass Best-of-N methods, suggesting that breakthroughs cannot solely rely on optimization-based verification~\citep{chen2024simple}.
\vspace{-1mm}

\paragraph{Sampling Optimization}
Another key area of research focuses on generating diverse but less paths or strategies for efficient scaling~\citep{wu2025depth,wang2024planning,chen2025we,shi2025multimodal,liao2025enhancing,song2025accelerated}. For instance, \citet{zeng2025revisiting} aggregate the shortest yet most varied reasoning paths for better scalability. Similarly, \citet{du2025optimizing} adjust the sampling temperature to increase diversity, leading to improved scaling. \citet{zhang2024scaling} and \citet{liu2025bag} optimize both candidate solution generation (e.g., prompts, temperature, and top-p) and reward mechanisms (such as self-evaluation and reward types), offering diverse strategies for parallel scaling. Moreover, \citet{qin2023cross}, \citet{luo-etal-2024-python}, and \citet{yu2025chain} enhance RLLM reasoning by scaling sampling across multiple natural and programming languages or varied expressions. Finally, \citet{yang2025towards} introduces a method where a small set of seed data, with varied response lengths, guides the model to engage in deeper reasoning by selecting the shortest correct responses across various inference efforts.

\begin{TakeawayBox}{Takeaways: Exploration Scaling}
    \begin{itemize}[left=2pt,topsep=1pt,itemsep=2pt, parsep=1pt]
        \item \textbf{Exploration Mechanisms in Long CoT Reasoning:} Exploration strategies like hypothesis branching and error backtracking are vital for overcoming limitations in linear reasoning paths and enhancing model performance.
        \item \textbf{Scaling Exploration:} Exploration can be scaled through sequential and parallel strategies to improve reasoning depth and efficiency.
        \item \textbf{Verification and Sampling Optimization:} Refining verification techniques and optimizing sampling for diverse reasoning paths are key to improving exploration efficiency and performance in Long CoT tasks.
    \end{itemize}
\end{TakeawayBox}

\subsection{Internal Exploration}
\begin{figure*}[t]
	\centering
	\includegraphics[width=0.98\textwidth]{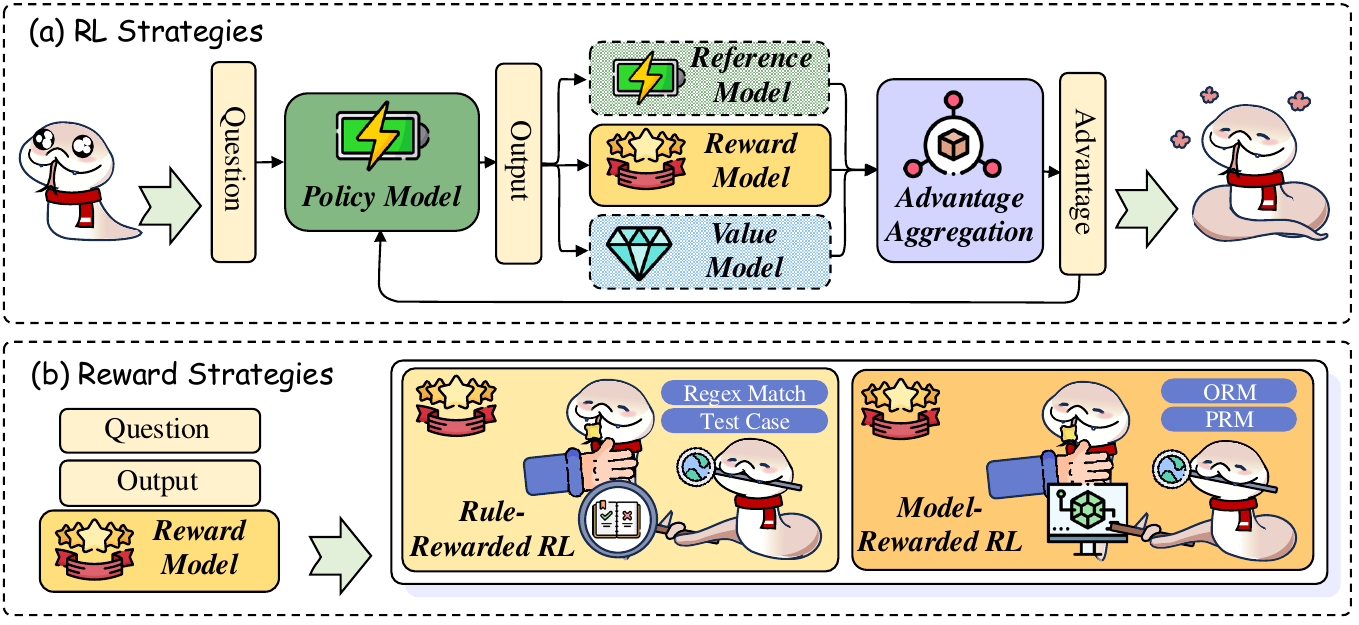}
	\caption{Two primary approaches for optimizing Internal Exploration: improving RL strategy through reference and value models, and designing reward strategies: either rule-based or model-based rewarding to enhance RL performance.}
	\label{fig:internal-exploration}
\end{figure*}

\vspace{-1mm}
\label{sec:exploration-internal}
As noted in \citet{chu2025sft}, \citet{shen2025satori}, and \citet{yeo2025demystifying}, SFT serves as a memory process, while RL enhances generalization~\citep{kumar2025llm,chen2025synergy}. Specifically, SFT stabilizes the model's output format, whereas RL improves its generalization capacity, which can increase learning efficiency by up to eight times in tasks such as mathematical reasoning~\citep{setlur2024rl}. Consequently, as shown in Figure~\ref{fig:internal-exploration}, leading research emphasizes the role of RL and reward strategies in enhancing the exploration capabilities of LLMs without external assistance. The performance comparison is presented in Table~\ref{tab:internal-exploration}.

\vspace{-2mm}\subsubsection{RL Strategies}\vspace{-1mm}
Recent advancements in RL strategies for exploration have led to notable improvements in various tasks, particularly in reasoning tasks~\citep{sun2025reward,lanchantin2025diverse,ji2024steiner,materzok2025cos,xiao2024comprehensive,zeng2025simplerl,yu2025dapo,hu2025open,yuyue2025vapo,qu2025optimizing,dang2025reinforcement,fei2025self,srivastava2025technical}. 

(1) \textbf{\textit{Reward-free RL:}} The first series of work focuses on RL optimization algorithms. Additionally, OREO~\citep{wang2024offline} propose an offline RL method that optimizes the soft Bellman equation, improving credit assignment for multi-step reasoning tasks and outperforming existing approaches in fields like mathematics and agent control. \citet{liu2024improving} propose Direct Advantage Policy Optimization, a novel offline RL method that leverages a separately trained critic to evaluate the accuracy of each reasoning step. This technique provides dense feedback for policy optimization, addressing both sparse rewards and training instability.
Further, some research focuses on adjusting the focus of RL algorithms to optimize exploration in targeted aspects. 
Specifically, CPL~\citep{wang2024cpl}, cDPO~\citep{lin2024critical}, and Focused-DPO~\citep{zhang2025focused} enhance exploration in Long CoT by prioritizing critical or error-prone areas through preference optimization, improving accuracy in those regions.
\citet{bartoldson2025trajectory} further adjusts the replay strategy of the training data, aiming to optimize reasoning performance.
\citet{li2025limr} introduce Learning Impact Measurement (LIM), an automated method for evaluating and prioritizing training samples based on their alignment with model learning trajectories. This approach enables efficient resource use and scalable implementation. For instance, ThinkPO~\cite{yang2025thinking} uses short CoT reasoning outputs as rejected answers and longer ones as chosen answers for the same question, applying DPO to encourage prioritization of longer reasoning outputs~\citep{zhou2025sweet}.

(2) \textbf{\textit{Reward-based RL:}} Reward-model-based RL refers to approaches that use a reward model or a verifier to guide learning and decision-making in the absence of explicit rewards~\citep{zhang2025reasoning,fan2025truncated,seed2025seed1,huang2025hapo,wang2025octothinker,wen2025reinforcement,yi2025sppd}. 
Earlier, Proximal Policy Optimization (PPO) was first introduced by \citet{schulman2017proximal}, which alternates between interacting with the environment to collect data and optimizing a surrogate objective function via stochastic gradient ascent, surpassing DPO \citep{ivison2024unpacking}. Subsequently, ReMax \citep{li2024remax} eliminates the need for additional value models in PPOs. By incorporating variance reduction and REINFORCE~\citep{sutton1999reinforce} techniques, it reduces over four hyperparameters, resulting in lower GPU memory usage and faster training. Building on this, DeepSeekMath \citep{shao2024deepseekmath} proposes Group Relative Policy Optimization (GRPO), replacing traditional value models with improved sampling strategies, thus significantly accelerating learning and achieving performance on par with GPT-4 in mathematics. \citet{hu2025reinforce++} and \citet{liu2025understanding} further refine GRPO with REINFORCE++ and Dr. GRPO, respectively, simplifying the algorithm and enhancing its training. Additionally, \citet{vassoyan2025ignore} and \citep{zhou2025q} improve exploration efficiency in smaller models by modifying the KL penalty, thus enhancing performance under distribution shifts. \citet{huang2025lean} introduce Decoupled Value Policy Optimization (DVPO), a streamlined framework that replaces reward modeling with a pretrained global value model (GVM) and eliminates the interdependence between actor and critic. To address the high-quality demands of reward models, \citet{cui2025process} propose PRIME (Process Reinforcement through IMplicit rEwards), which integrates the SFT model as a PRM within a unified reinforcement learning framework, enabling online updates through policy rollouts and outcome labels via implicit process rewards.

More recently, \citet{liang2025sws} introduce Self-aware Weakness-driven Problem Synthesis, a reinforcement-learning method that generates challenges tailored to an RLLM's specific weaknesses~\citep{wu2025sharp,feng2025visualsphinx}. By concentrating training on its most difficult aspects, the model achieves more focused and effective reasoning improvements~\citep{song2025fastcurl}.
\citet{wang2025reinforcement} introduce ROLL, a method designed to support R1-level large-scale training of RLLMs, enabling the efficient exploration and optimization of reasoning paths within the Mixture-of-Experts (MOE) structure~\citep{wang2025two}.
\citet{fu2025areal} introduce AReaL, a large-scale asynchronous reinforcement learning system for language reasoning, which enhances the efficiency and effectiveness of training RLLMs.
\citet{ma2025learning} propose a novel method combining interleaved SFT and RL to address challenging questions where RL typically fails. This approach enables RLLMs to learn from mistakes and enhance reasoning abilities. \citet{huang2025blending} and \citet{fu2025srft} further improve exploration efficiency by integrating SFT and RL with prefix sampling. Fruthermore, \citet{yan2025learning} and \citet{liang2025squeeze} guide RLLMs in reasoning under off-policy reinforcement learning~\citep{li2025repo,wang2024offline}, improving both training sample efficiency and learning stability~\citep{mu2025dissecting}.

\begin{table*}[t]
    \centering
    \resizebox{0.98\textwidth}{!}{
        \begin{tabular}{l|l|ccccc}
            \toprule
            \textbf{Method}                                       & \textbf{Backbone}                            & \textbf{GSM8K} & \textbf{AIME 2024} & \textbf{MATH 500} & \textbf{GPQA} & \textbf{LiveCodeBench} \\
            \midrule
            \rowcolor{gray!8}\multicolumn{7}{c}{\textit{Base Model}}                                                                                                                                                \\
            \midrule
            GPT-4o~\citep{achiam2023gpt}                          & -                                            & 92.9           & 9.3                & 76.6              & 53.6          & 33.4                   \\
            Llama-3.1-70B-Instruct~\citep{dubey2024llama}         & -                                            & 94.1           & 13.3               & 68.0              & -             & -                      \\
            Claude 3.5 Sonnet~\citep{anthropic2024claude}         & -                                            & -              & 16.0               & 78.3              & 65.0          & 38.9                   \\
            Qwen2.5-Coder-32B-Instruct~\citep{hui2024qwen25coder} & -                                            & -              & 20.0               & 71.2              & 33.8          & 25.0                   \\
            Qwen2.5-70B-Instruct~\citep{yang2024qwen25}           & -                                            & -              & 20.0               & 79.4              & 49.0          & 33.0                   \\
            Llama-3.3-70B-Instruct~\citep{dubey2024llama}         & -                                            & -              & 36.7               & 73.9              & 50.5          & 34.8                   \\

            DeepSeek-V3~\citep{liu2024deepseek}                   & -                                            & -              & 39.2               & 90.2              & -             & 36.2                   \\
            \midrule
            \rowcolor{gray!8}\multicolumn{7}{c}{\textit{SFT Strategies}} \\
            \midrule
            DeepSeek-R1-Distill-Llama-70B~\citep{guo2025deepseek} & -                                            & -              & 70.0               & -                 & -             & 57.9                   \\
            DeepSeek-R1-Distill-Qwen-32B~\citep{guo2025deepseek}  & -                                            & -              & 72.6               & -                 & -             & 54.6                   \\
            START~\citep{li2025start}  & QwQ-32B-preview~\citep{team2025qwq} & -              & 66.7 & 94.4 & 63.6 & 47.3 \\
            \midrule
            \rowcolor{gray!8}\multicolumn{7}{c}{\textit{RL Strategies}} \\
            \midrule
            DPO~\citep{rafailov2023direct}                        & DeepSeekMath 7B~\citep{shao2024deepseekmath} & 82.4           & -                  & -                 & -             & -                      \\
            KTO~\citep{ethayarajh2024kto}                         & DeepSeekMath 7B~\citep{shao2024deepseekmath} & 82.5           & -                  & -                 & -             & -                      \\
            OREO~\citep{wang2024offline}                          & DeepSeekMath 7B~\citep{shao2024deepseekmath} & 86.9           & -                  & -                 & -             & -                      \\
            PPO~\citep{schulman2017proximal}                      & GLM4-9B-SFT~\citep{glm2024chatglm}           & 85.5           & -                  & -                 & 31.5          & 24.3                   \\
            GRPO~\citep{shao2024deepseekmath}                     & GLM4-9B-SFT~\citep{glm2024chatglm}           & 86.1           & -                  & -                 & 31.7          & 22.8                   \\
            Eurus-2-7B-PRIME~\citep{cui2025process}                        & Qwen2.5-Math-7B-Base~\citep{yang2024qwen25math}          & -              & 26.7              & 79.2          & -                   & -               \\
            Search-o1~\citep{li2025search}                        & QwQ-32B-preview~\citep{team2025qwq}          & -              & 56.7               & 86.4              & 63.6          & 33.0                   \\
            
            \midrule
            \rowcolor{gray!8}\multicolumn{7}{c}{\textit{Reward Strategies}}                                                                                                                                         \\
            \midrule
            OpenMath2~\citep{toshniwal2024openmath2}              & Llama-3.1-70B~\citep{dubey2024llama}         & 94.1           & 13.3               & 71.8              & -             & -                      \\
            Satori~\citep{shen2025satori}                         & Qwen-2.5-Math-7B                             & 93.9           & 23.3               & 83.6              & -             & -                      \\
            T1-SFT~\citep{hou2025advancing}                       & Qwen2.5-32B~\citep{yang2024qwen25}           & -              & 24.9               & 83.4              & 49.5          & -                      \\
            T1~\citep{hou2025advancing}                           & Qwen2.5-32B~\citep{yang2024qwen25}           & -              & 50.6               & 92.4              & 56.1          & -                      \\
            DeepSeek-R1-lite~\citep{guo2025deepseek}              & -                                            & -              & 52.5               & 91.6              & 58.5          & 51.6                   \\
            rStar-Math~\citep{guan2025rstar}                      & Qwen2.5-Math-7B~\citep{yang2024qwen25math}   & 95.2           & 53.3               & 90.0              & -             & -                      \\
            QwQ-32B-preview~\citep{team2025qwq}                   & -                                            & 95.5           & 53.3               & 90.6              & 58.2          & 40.6                   \\
            o1-preview~\citep{jaech2024openai}                    & -                                            & -              & 56.7               & 85.5              & 73.3          & 53.6                   \\
            o3-mini-low~\citep{jaech2024openai}                   & -                                            & -              & 60.0               & -                 & -             & 61.8                   \\
            o1-mini~\citep{jaech2024openai}                       & -                                            & -              & 63.6               & 90.0              & -             & 53.8                   \\
            Kimi k1.5~\citep{team2025kimi}                        & -                                            & -              & 77.5               & 96.2              & -             & 62.5                   \\
            QwQ-32B~\citep{team2025qwq}                           & -                                            & -              & 79.5               & -                 & -             & 73.1                   \\
            o3-mini-medium~\citep{jaech2024openai}                & -                                            & -              & 79.6               & -                 & -             & 72.3                   \\
            DeepSeek-R1~\citep{guo2025deepseek}                   & -                                            & -              & 79.8               & 97.3              & -             & 71.6                   \\
            o1~\citep{jaech2024openai}                            & -                                            & -              & 83.3               & 96.4              & -             & 67.4                   \\
            o3-mini-high~\citep{jaech2024openai}                  & -                                            & -              & 87.3               & -                 & -             & 84.6                   \\
            \bottomrule
        \end{tabular}
    }
    \caption{Performance of various internal exploration methods on different benchmarks, primarily ordered by AIME 2024. ``-'' indicates that the paper did not report this score.}
    \label{tab:internal-exploration}
\end{table*}

\subsubsection{Reward Strategies}
\paragraph{Rule-rewarded RL}
The studies explore advancements in training advanced RLLMs using rule-rewarded RL  to enhance exploration strategies and reasoning accuracy~\citep{huang2025pitfalls}. These efforts primarily focus on three types of rewards:
(1) \textbf{\textit{Correctness Rewarding:}} Correctness rewards are fundamental for guiding RLLMs toward accurate answers. Specifically, \citet{singh2024beyond} introduce a binary reward system (positive or negative) to facilitate exploration, achieving simple yet effective performance improvements. Similarly, the DeepSeek-R1~\citep{guo2025deepseek} employs rule-extracted accuracy as an RL reward, scaling this approach to larger scenarios and training sizes, thereby enhancing both exploration and reasoning tasks~\citep{lyu2025exploring, el2025competitive}. Furthermore, O1-Coder\citet{zhang2024o1},  StepCoder~\citep{dou2024stepcoder}, and SWE-RL~\citep{wei2025swerl} address challenges in code generation by developing a test case generator, which standardizes code testing, ensuring accurate generation~\citep{xiong2024building,yu2025z1}. 
(2) \textbf{\textit{Format Rewarding:}} Further, format rewards are used to encourage better reasoning paradigms. \citet{guo2025deepseek} introduce this concept to effectively guide reasoning and exploration~\citep{xie2025logic}. \citet{xie2025logic} expanded on this with a three-stage, rule-based RL approach, enabling the Qwen-7B model to learn complex multi-path exploration, which significantly improved both output format and corresponding length consistency. Additionally, \citet{wu2025thought} propose TAPO (Thought-Augmented Policy Optimization), a framework that integrates external high-level guidance (``thought patterns'') into RL, successfully balancing model exploration with external guidance.
(3) \textbf{\textit{Scaling rewarding:}} Moreover, scaling rewards are applied to promote longer reasoning chains and broader exploration. Recent studies~\citep{chen2024unlocking,parashar2025inference,kim2025metastable} highlight the need for progressively scaled reasoning lengths to overcome the limitations of current reasoning approaches. As a result, research has focused on scaling exploration~\citep{xie2025logic,ye2025emergence}. However, excessive scaling can lead to inefficiency and overcomplicated reasoning~\citep{cuadron2025danger}. Kimi-K1.5~\citep{team2025kimi}, \citet{yang2025towards} and \citet{arora2025training} proposed Long2Short techniques, favoring shorter, more accurate reasoning may also significantly improve efficiency and performance.

\paragraph{Model-rewarded RL}
It refers to a class of techniques in which RL algorithms are enhanced by leveraging additional reward models, to guide exploration and improve decision-making processes~\citep{su2025expanding}.
Earlier in 2021, OpenAI~\citep{cobbe2021training} propose a ``Gen-Verifier'' paradigm to train a correctness-oriented ORM and used ORM-rewarded RL to surpass SFT performance. Recently, with rapid advancements in PRM, several studies~\citep{wan2024alphazerolike, zhang2024restmcts, deepscaler2025} have scaled reinforcement learning by enhancing exploration through step-level correctness rewarding~\citep{she2025r,zhang2025stephint}. Building on this, \citet{hou2025advancing} introduce entropy rewards and dynamic regularization to further optimize the reasoning process~\citep{cheng2025reasoning}. STeCa~\citep{wang2025steca} identifies suboptimal actions during exploration by comparing step-level rewards and adjusting trajectories to improve deep reasoning. Additionally, the Kimi-K1.5 model~\citep{team2025kimi} extends PRM paradigms into multimodal scenarios, achieving state-of-the-art performance in multi-modal reasoning tasks through a streamlined reinforcement learning framework.

\begin{TakeawayBox}{Takeaways: Internal Exploration}
    \begin{itemize}[left=2pt,topsep=1pt,itemsep=2pt, parsep=1pt]
        \item \textbf{SFT and RL Synergy:} The combination of Self-Feedback Training (SFT) and Reinforcement Learning (RL) improves model output stability and generalization, enhancing learning efficiency in reasoning tasks.
        \item \textbf{Advancements in RL Exploration:} Recent RL strategies, including reward-model-free and reward-model-based approaches, optimize exploration and reasoning, improving efficiency in tasks like multi-step reasoning.
        \item \textbf{Reward Strategies:} Correctness, format, and scaling rewards help refine exploration and reasoning accuracy by guiding models toward better performance in specific areas.
    \end{itemize}
\end{TakeawayBox}
\begin{figure*}[b]
	\centering
	\includegraphics[width=0.78\textwidth]{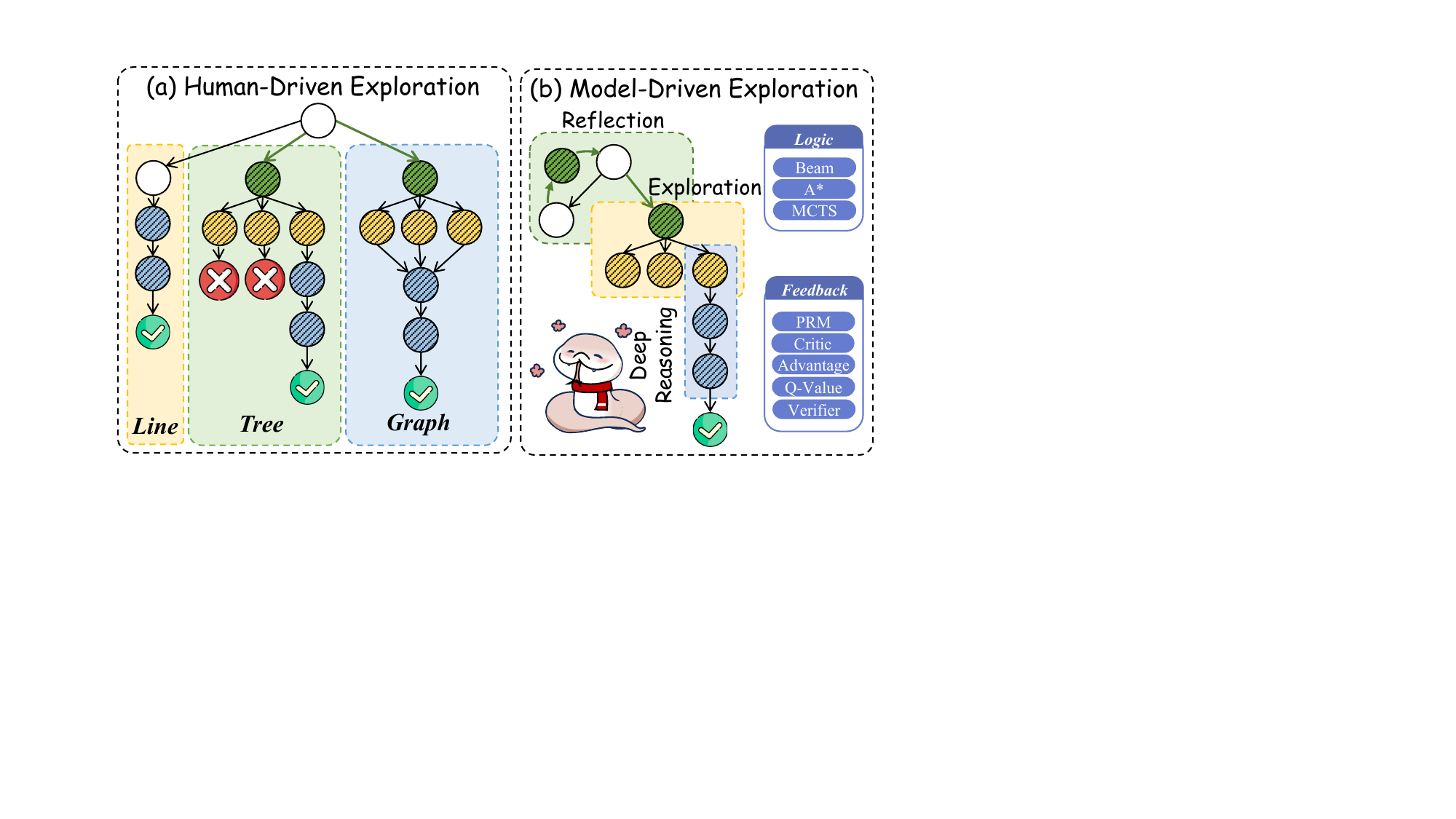}
	\caption{External exploration policies can be classified into two categories based on the management role of the process: (1) Human-Driven Exploration, which is guided by human-defined prompts and fixed pipelines, and (2) Model-Driven Exploration, which is driven by models and employs dynamic, adaptive search structures.}
	\label{fig:external-exploration}
\end{figure*}
\subsection{External Exploration}
\label{sec:exploration-external}
The exploration of coding strategies in AI systems is advancing through innovative frameworks aimed at enhancing search efficiency and decision-making quality. As shown in Figure~\ref{fig:external-exploration}, external exploration policies fall into two categories based on process management: (1) Human-Driven Exploration, guided by human-defined prompts and fixed pipelines, and (2) Model-Driven Exploration, driven by models with dynamic, adaptive search structures. The detailed performance comparison is presented in Table~\ref{tab:external-exploration}.

\subsubsection{Human-driven Exploration}
Human-driven exploration refers to human-designed constant pipeline exploration for long-term exploration~\citep{liu2025deciphering,li2025policy}. Several studies highlight the effectiveness of prompt-based~\citep{kang-etal-2024-empirical,ye2024advances,golovneva2023pathfinder,han2025enhancing,wu2025beyond,qin2025divide,mitra2025motif,zhang2025rot,shi2025layoutcot}, tree-structured~\citep{zhou2023leasttomost,yao2023tree,chen2024understanding,qiu2024treebon,mo2024tree,besta2024demystifying,he2025logictree} and even graph-structured~\citep{besta2024graph,teng2025atom,puerta2025roadmap,cao-2024-graphreason,zhang2024diagram,zhao2025enhancing} search frameworks, demonstrating superior performance and scalability over traditional methods across various datasets. Building on this, CodeTree~\citep{li2024codetree} and Tree-of-Code~\citep{ni2024tree} integrate a tree-based structure with execution and LLM feedback, utilizing multi-agents to optimize multi-stage decisions, thereby improving both strategy planning and solution refinement~\citep{tang2025thinking}.
\citet{cheng2024spar} generalize this approach with the Self-Play with Tree-Search Refinement (SPAR) strategy, which generates valid, comparable preference pairs to enhance instruction-following capabilities.
\citet{bi2024forest} and \citet{light2024scattered} extend tree search to a multi-tree paradigm, introducing the Forest-of-Thought framework, which incorporates multiple reasoning trees to improve exploration capabilities to solve complex tasks with greater accuracy.
Furthermore, \citet{li2025start} explores the integration of Python tools into Long CoT frameworks by both prompting and training, performing test-time scaling more effectively.
\begin{table*}[t]
    \centering
    \resizebox{0.98\textwidth}{!}{
        \begin{tabular}{l|l|cccc}
            \toprule
            \textbf{Method}                                       & \textbf{Backbone}                                    & \textbf{GSM8K} & \textbf{MATH} & \textbf{OlympiadBench} & \textbf{HumanEval+} \\
            \midrule
            \rowcolor{gray!8}\multicolumn{6}{c}{\textit{Base Model}}                                                                                                                                     \\
            \midrule
            DeepSeekMath-7B-Instruct~\citep{shao2024deepseekmath} & -                                                    & 83.7           & 57.4          & -                      & -                   \\
            DeepSeekMath-7B-RL~\citep{shao2024deepseekmath}       & -                                                    & 88.2           & 52.4          & 19.0                   & -                   \\
            Qwen2-72B-Instruct~\citep{yang2024qwen2}              & -                                                    & 93.2           & 69.0          & 33.2                   & -                   \\
            Llama-3.1-70B-Instruct~\citep{dubey2024llama}         & -                                                    & 94.1           & 65.7          & 27.7                   & -                   \\
            GPT-4~\citep{achiam2023gpt}                           & -                                                    & 94.2           & 73.4          & -                      & -                   \\
            Claude-3.5-Sonnet~\citep{anthropic2024claude}         & -                                                    & 96.4           & 71.1          & -                      & -                   \\
            GPT-4o~\citep{achiam2023gpt}                          & -                                                    & -              & 73.4          & 40.6                   & 81.7                \\
            Qwen2.5-Math-72B-Instruct~\citep{yang2024qwen25math}  & -                                                    & -              & 83.0          & 49.7                   & -                   \\
            \midrule
            \rowcolor{gray!8}\multicolumn{6}{c}{\textit{Human-driven Exploration}}                                                                                                                       \\
            \midrule
            AlphaLLM~\citep{wang2024towards}                      & Llama-3-8B-Instruct~\citep{dubey2024llama}           & -              & 32.6          & -                      & -                   \\
            Least-to-Most-SC~\citep{zhou2023leasttomost}          & LLaMA-33B~\citep{touvron2023llama1}                   & 42.5           & -             & -                      & -                   \\
            LLM2~\citep{yang2024llm2}                             & Llama-3-8B~\citep{dubey2024llama}                    & 88.0           & 48.6          & -                      & -                   \\
            CodeTree~\citep{li2024codetree}                       & GPT-4o~\citep{achiam2023gpt}                         & -              & -             & -                      & 86.0                \\
            \midrule
            \rowcolor{gray!8}\multicolumn{6}{c}{\textit{Model-driven Exploration}}                                                                                                                       \\
            \midrule
            STILL-1~\citep{jiang2024technical}                    & LLama-3.1-8B-Instruct~\citep{dubey2024llama}         & -              & -             & 34.3                   & -                   \\
            Reflexion~\citep{shinn2023reflexion}                  & GPT-4o~\citep{achiam2023gpt}                         & -              & -             & -                      & 84.8                \\
            MapCoder~\citep{islam2024mapcoder}                    & GPT-4o~\citep{achiam2023gpt}                         & -              & -             & -                      & 81.7                \\
            Resample~\citep{li2022competition}                    & GPT-4o~\citep{achiam2023gpt}                         & -              & -             & -                      & 84.8                \\
            SRA-MCTS~\citep{xu2024sra}                            & Llama-3.1-8B~\citep{dubey2024llama}                  & -              & -             & -                      & 57.9                \\
            RAP~\citep{hao-etal-2023-reasoning}                          & LLaMA-33B~\citep{touvron2023llama1}                  & 51.6           & -             & -                      & -                   \\
            Mindstar~\citep{kang2024mindstar}                     & Llama-2-7B~\citep{touvron2023llama}                  & 68.8           & 33.9          & -                      & -                   \\
            Mindstar~\citep{kang2024mindstar}                     & Mistral-7B~\citep{jiang2023mistral7b}                & 73.7           & 38.2          & -                      & -                   \\
            TS-LLM~\citep{wan2024alphazerolike}                   & GPT-3.5-turbo                                        & 74.0           & -             & -                      & -                   \\
            LiteSearch~\citep{wang2024litesearch}                 & Llama-3-8B-Instruct~\citep{dubey2024llama}           & 75.7           & -             & -                      & -                   \\
            MARIO-34B~\citep{liao2024mario}                       & CodeLlama-34B~\citep{roziere2023code}                & 78.2           & 53.5          & -                      & -                   \\
            ToRA-Code-34B~\citep{gou2023tora}                     & CodeLlama-34B~\citep{roziere2023code}                & 80.7           & 50.8          & -                      & -                   \\
            MathCoder-34B~\citep{wang2023mathcoder}               & CodeLlama-34B~\citep{roziere2023code}                & 81.7           & 46.1          & -                      & -                   \\
            AlphaMath~\citep{chen2024alphamath}                   & DeepSeekMath-7B-Base~\citep{shao2024deepseekmath}    & 83.2           & 64.0          & -                      & -                   \\
            MathGenie-34B~\citep{lu2024mathgenie}                 & CodeLlama-34B~\citep{roziere2023code}                & 84.1           & 55.1          & -                      & -                   \\
            MCTS-DPO~\citep{xie2024monte}                         & Llama-3.1-8B-Instruct~\citep{dubey2024llama}         & 85.7          & -             & -                      & -                   \\
            Intrinsic Self-Correct                                & Llama-3.1-8B-Instruct~\citep{dubey2024llama}         & 86.1          & -             & -                      & -                   \\
            MCTS-IPL~\citep{jiang2024towards}                     & Llama-3.1-8B-Instruct~\citep{dubey2024llama}         & 86.8          & -             & -                      & -                   \\
            NuminaMath-72B-CoT~\citep{li2024numina}               & Qwen2-72B~\citep{yang2024qwen2}                      & 90.8           & 66.7          & 32.6                   & -                   \\
            AutoRace~\citep{hao2024llm}                           & GPT-4~\citep{achiam2023gpt}                          & 91.0           & -             & -                      & -                   \\
            LLaMA-Berry~\citep{zhang2024llama}                    & Llama-3.1-8B-Instruct~\citep{dubey2024llama}         & 96.1           & 75.3          & 55.1                   & -                   \\
            MCTSr~\citep{zhang2024accessing}                      & Llama-3-8B-Instruct~\citep{dubey2024llama}           & 96.7           & 58.2          & -                      & -                   \\
            BoostStep~\citep{zhang2025booststep}                  & Qwen2.5-Math-72B-Instruct~\citep{yang2024qwen25math} & -              & 85.2          & 52.7                   & -                   \\
            \bottomrule
        \end{tabular}
    }
    \caption{Performance of various external exploration methods on different benchmarks. ``-'' indicates that the paper did not report this score.}
    \label{tab:external-exploration}
\end{table*}

\subsubsection{Model-driven Exploration}
Building on previous research, model-feedback-assisted exploration has advanced significantly, which is driven by model and dynamic adaptive search structure, with optimization emerging as a central focus. Currently, there are three key directions guiding model-driven exploration:\vspace{-1mm}

\paragraph{Enhancing Exploration Logics} Recent efforts have focused on improving exploration structures during iterations for better logical quality. 
(1) \textbf{\textit{Beam Search:}} Earlier, \citet{xie2023selfevaluation} introduced a decoding algorithm that integrates self-evaluation guidance via stochastic beam search, using it as a more reliable automatic criterion to streamline the search in the reasoning space, thereby enhancing prediction quality~\citep{mitra2025motif}. Similarly, \citet{zhu2024deductive} propose Deductive Beam Search (DBS), which combines CoT and deductive reasoning with stepwise beam search for RLLMs.
(2) \textbf{\textit{A* Search:}} On another front, \citet{lehnert2024beyond} present Searchformer, which predicts A* algorithm dynamics to improve task performance and reduce search steps~\citep{chen2025better}. Later, \citet{kang2024mindstar} introduce the MindStar (M*) framework, which optimizes reasoning paths through beam search and Levin tree search methods, further enhancing reasoning performance.
(3) \textbf{\textit{MCTS Search:}} 
Building on the advantages of MCTS, a series of studies, such as Macro-o1~\citep{zhao2024marco}, STILL-1~\citep{jiang2024technical}, SRA-MCTS~\citep{xu2024sra}, and RFTT~\citep{zhang2025reasoning}, utilize MCTS to guide more effective exploration~\citep{zhang2024aflow,li2024rethinkmcts,kadam2024gpt,jiang2024towards,zheng2025monte,putta2024agent,park2024ensembling,lin2025cmcts}. \citet{xu2023no} utilizes energy function for better exploration during Long CoT. \citet{yao2024mulberry} further advance this by introducing Collective MCTS (CoMCTS), which leverages collective learning across multiple LLMs to enhance reasoning.
Further, MC-NEST~\citep{rabby2024mc} integrates Nash Equilibrium strategies to balance exploration and exploitation, improving LLM decision-making in multi-step mathematical tasks~\citep{yang2025long,zhao2025exploring}. 
Additionally, CoAT~\citep{pan2025coat} expands the MCTS algorithm with a dynamic correlation memory mechanism, enabling the system to dynamically store new information during inference.
Despite MCTS's benefits, it is often hindered by a large action space and inefficient search strategies, which complicate the generation of Long CoTs. To address this, \citet{lin2025leveraging} propose constraining the action space and refining the search strategy to facilitate the emergence of Long CoTs. Finally, these methods have been extended to interactive environments, significantly improving success rates in automated exploration tasks~\citep{wang2024cooperative,koh2024tree,light2024strategist,xiong2024deliberate,zhai2024finetuning,park2025maporl,wang2025x,liu2025spiral}.\vspace{-1mm}

\paragraph{Exploration-Path Feedback} Another approach aims to enhance reward models, refining both reasoning exploration and output quality. \citet{liu2024dont,liu2024making} propose PPO-augmented MCTS, a decoding algorithm that integrates an optimized value model with MCTS, providing concise feedback that significantly improves reasoning exploration and the controllability of text generation. Similarly, \citet{zhang2024llama} introduce LLaMA-Berry, which combines MCTS with Self-Refine (SR-MCTS), incorporating a Pairwise Preference Reward Model (PPRM) and Enhanced Borda Count (EBC) to address scoring variability and local optima in mathematical feedback, particularly excelling in Olympiad-level benchmarks. Further refining this, \citet{xiang2024atomthink} present AtomThink, which leverages PRM and search strategies to optimize each atomic step, guiding the model to iteratively refine its reasoning process and generate more reliable solutions. \citet{puri2025probabilistic} leverage sampling-based techniques for PRM to explore the state distribution of a state-space model with an approximate likelihood, rather than optimizing its mode directly.\vspace{-1mm}

\paragraph{Unified Improvements} The final direction merges advances in exploration strategies and path feedback. Specifically, \citet{guan2025rstar} introduce a multi-step iterative learning approach that optimizes both PRM and RLLM via MCTS and a self-evolving process, significantly advancing mathematical reasoning. Similarly, \citet{lee2025evolving} and \citet{kim2025hypothesis} propose a paradigm that enhances deep reasoning, exploration, and response refinement, further improving RLLM performance. QLASS~\citep{lin2025qlass} and DQO~\citep{liu2024enhancing} build exploration trees and use Q-value-based reward modeling for stepwise guidance, improving feedback efficiency in large search spaces~\citep{li2024process,guo2025critiq}. \citet{zeng2025sift} propose that RLLMs are always lost in extensive exploration in Long CoT, therefore, they introduce a sticker to further improve the exploration effectiveness.
\vspace{-1mm}

\begin{TakeawayBox}{Takeaways: External Exploration}
    \begin{itemize}[left=2pt,topsep=1pt,itemsep=2pt, parsep=1pt]
        \item \textbf{Human-driven Exploration:} Recent research highlights the effectiveness of tree-structured, graph-based, and prompt-based search frameworks, improving scalability and task-solving accuracy through multi-agent feedback.
        \item \textbf{Model-driven Exploration:} Exploration strategies like Beam Search, A* Search, and MCTS, along with their advancements, enhance reasoning paths and search efficiency.
        \item \textbf{Unified Improvements and Path Feedback:} Integrating exploration strategies with feedback models, optimizes reasoning exploration and output reliability.
    \end{itemize}
\end{TakeawayBox}

  \begin{table*}[t]
    \centering
    \resizebox{\textwidth}{!}{
    \begin{tabular}{lllll}
    \toprule
    \textbf{Name} & \textbf{Category} & \textbf{Source} & \textbf{Modality} & \textbf{Quantity} \\
    \midrule
    \rowcolor{gray!8}\multicolumn{5}{c}{\textit{Manual Annotated}}\\
    \midrule
    R1-OneVision~\citep{team2025r1onevision} & Mathematics, Science & Rule & Vision + Lang & 119K \\
    M3CoT~\citep{chen-etal-2024-m3cot} & Mathematics, Science, General & Human & Vision + Lang & 11K \\
    Big-Math-RL-Verified~\citep{albalak2025bigmath} & Mathematics & Human & Lang & 251K \\
    GSM8K~\citep{cobbe2021training} & Mathematics & Human & Lang & 8K \\
    LiveCodeBench (History)~\citep{jain2025livecodebench} & Code & Human & Lang & 0.9K \\
    LeetCode~\citep{xia2025leetcodedataset} & Code & Human & Lang & 2K \\
    ARC-AGI~\citep{chollet2024arc} & Logic Puzzle & Human Synthesis & Lang & 0.4K \\
    ARC-AGI-2~\citep{chollet2025arc} & Logic Puzzle & Human Synthesis & Lang & 1K \\
    BARC~\citep{li2024combining} & Logic Puzzle & Human Synthesis & Lang & 3.4K \\
    Code I/O (PyEdu)~\citep{li2025codei} & Code Execution Simulation & Human Synthesis & Lang & 227K \\
    HiTab~\citep{cheng2021hitab} & Tabular & Human & Lang & 7.5K \\
    MultiHierTT~\citep{li2025codei} & Code Execution Simulation & Human Synthesis & Lang & 7.8K \\
    
    \midrule
    \rowcolor{gray!8}\multicolumn{5}{c}{\textit{Direct Distillation}}\\
    \midrule
    NaturalReasoning~\citep{yuan2025naturalreasoning} & Science, General & Llama3.3-70B & Lang & 1M \\
    NuminaMath-CoT~\citep{li2024numina} & Mathematics & GPT-4o & Lang & 860K \\
    NuminaMath-TIR~\citep{li2024numina} & Mathematics & GPT-4o & Lang & 73K \\
    DART-Math-uniform~\citep{tong2024dart} & Mathematics & DeepSeekMath-7B-RL & Lang & 591K \\
    DART-Math-hard~\citep{tong2024dart} & Mathematics & DeepSeekMath-7B-RL & Lang & 585K \\
    DART-Math-pool-math~\citep{tong2024dart} & Mathematics & DeepSeekMath-7B-RL & Lang & 1.6M \\
    DART-Math-pool-gsm8k~\citep{tong2024dart} & Mathematics & DeepSeekMath-7B-RL & Lang & 2.7M \\
    OpenO1-SFT~\citep{team2025openo1} & Mathematics, Science, General & - & Lang & 78K \\
    OpenO1-SFT-Pro~\citep{team2025openo1} & Mathematics, Science, General & - & Lang & 126K \\
    OpenO1-SFT-Ultra~\citep{team2025openo1} & Mathematics, Science, General & - & Lang & 28M \\
    
    Medical-o1~\citep{chen2024huatuogpt} & Medicine & DeepSeek R1 & Lang & 50K \\
    AoPS-Instruct~\citep{mahdavi2025leveraging} & Mathematics & Qwen2.5-72B & Lang & 647K \\
    Orca-Math~\citep{mitra2024orca} & Mathematics & GPT4 & Lang & 200K \\
    MATH-plus~\citep{yue2025mammoth2} & Mathematics & GPT4 & Lang & 894K \\
    UltraInteract-SFT~\citep{yuan2025advancing} & Mathematics, Code, Logic & GPT4 CoT + PoT & Lang & 289K \\
    MathCodeInstruct~\citep{wang2024mathcoder,zhou2024solving} & Mathematics & GPT4 + Codellama PoT & Lang & 79K \\
    MathCodeInstruct-Plus~\citep{wang2024mathcoder,zhou2024solving} & Mathematics & - & Lang & 88K \\
    OpenMathInstruct-1~\citep{toshniwal2024openmath} & Mathematics & Mixtral-8x7B PoT & Lang & 5M \\
    OpenMathInstruct-2~\citep{toshniwal2024openmath2} & Mathematics & Llama3.1-405B & Lang & 14M \\
    AceMath-Instruct~\citep{liu2024acemath} & Mathematics, General & Qwen2.5-Math-72B + GPT-4o-mini & Lang & 5M \\
    QwQ-LongCoT~\citep{team2025qwqlongcot} & General & QwQ & Lang & 286K \\
    SCP-116K~\citep{lu2025scp116} & Science & QwQ + O1-mini & Lang & 117K \\
    R1-Distill-SFT~\citep{slam-distillation-from-r1} & Mathematics & DeepSeek-R1-32B & Lang & 172K \\
    Sky-T1-Data~\citep{team2025novasky} & Mathematics, Code, Science, Puzzle & QwQ & Lang & 17K \\
    Bespoke-Stratos-17k~\citep{bespokestratos} & Mathematics, Code, Science, Puzzle & DeepSeek R1 & Lang & 17K \\
    s1K~\citep{muennighoff2025s1} & Mathematics & DeepSeek R1 & Lang & 1K \\
    MedThoughts-8K~\citep{wei2025medthoughts} & Medicine & DeepSeek R1 & Lang & 8K \\
    PrimeIntellect~\citep{2025synthetic1} & Code & DeepSeek R1 & Lang & 16.3K \\
    Medical-R1-Distill-Data~\citep{chen2024huatuogpt} & Medicine & DeepSeek R1 & Lang & 22K \\
    Medical-R1-Distill-Data-Chinese~\citep{chen2024huatuogpt} & - & - & Lang & 17K \\
    RLVR-GSM-MATH~\citep{lambert2024tulu3} & Mathematics & - & Lang & 30K \\
    LIMO~\citep{ye2025limo} & Mathematics & Human + DeepSeek R1 + Qwen2.5-32B & Lang & 817 \\
    OpenThoughts-114k~\citep{team2025openthoughts} & Mathematics, Code, Science, Puzzle & - & Lang & 114K \\
    Magpie-Reasoning-V2~\citep{xu2024magpie} & Mathematics, Code & DeepSeek-R1 + Llama-70B & Lang & 250K \\
    Dolphin-R1~\citep{team2025dolphinr1} & Mathematics, Science & DeepSeek R1 + Gemini2 + Dolphin & Lang & 814K \\
    \midrule
    \rowcolor{gray!8}\multicolumn{5}{c}{\textit{Search-based Distillation}}\\
    \midrule
    STILL-1~\citep{jiang2024technical} & Mathematics, Code, Science, Puzzle & LLaMA-3.1-8B-Instruct + MCTS & Lang & 5K \\
    \midrule
    \rowcolor{gray!8}\multicolumn{5}{c}{\textit{Validated Distillation}}\\
    \midrule
    KodCode-V1~\citep{xu2024kodcode} & Code & GPT4 + Test case validation & Lang & 447K \\
    KodCode-V1-SFT-R1~\citep{xu2024kodcode} & - & DeepSeek R1 + Test case validation & Lang & 443K \\
    OpenR1-Math~\citep{team2025openr1math} & Mathematics & DeepSeek R1 + Rule \& LLM Validation & Lang & 225K \\
    Chinese-DeepSeek-R1-Distill-Data~\citep{liu2025chinese} & Mathematics, Science, General & DeepSeek R1 + Rule \& LLM Validation & Lang & 110K \\
    AM-DeepSeek-R1-Distilled~\citep{zhao20251} & Mathematics, Code, General & Reward Model + Rule \& LLM Validation & Lang & 1.4M \\
    OR1~\citep{skywork-or1-2025} & Mathematics, Code, General & Human Question + Rule Validation & Lang & 105K \\
    DeepScaler~\citep{deepscaler2025} & Mathematics & Human Question + Rule Validation & Lang & 40.3 \\
    DAPO~\citep{yu2025dapo} & Mathematics & Human Question + Rule Validation & Lang & 17K \\
    TACO-Verified~\citep{likaixin2024taco} & Code & Human + Rule Validation & Lang & 0.9K \\
    WebInstruct-Verifed~\citep{ma2025general} & Science, General & Web Crawling + Rule \& LLM Validation & Lang & 232K \\
    Guru92K~\citep{cheng2025revisiting} & Mathematics, Code, Puzzle, General & Unified + Rule Validation & Lang & 92K \\
    \bottomrule
    \end{tabular}
    }
    \caption{The statistics of training data for Long CoT.}
    \label{tab:train-data}
\end{table*}
\section{Training Resources}
\subsection{Open-Sourced Training Framework}
A range of open-source training frameworks has equipped researchers and developers with tools to optimize training and enhance inference. Each framework is built on distinct design principles and features. Early frameworks like SimpleRL~\citep{zeng2025simplerl} and DeepScaler~\citep{deepscaler2025} quickly replicated R1’s technology stack. Others, such as X-R1~\citep{team2025xr1} and TinyZero~\citep{tinyzero}, emphasize delivering an intuitive ``Aha moment'' experience for under \$50. Open-Reasoner-Zero~\citep{OpenReasonerZero2025} replicated the DeepSeek-R1-zero training scheme with a 32B model and achieved a similar performance.
Additionally, LLM Reasoner~\citep{hao2024llm} provides tools to help researchers adapt strategies for External Exploration. Frameworks such as OpenR~\citep{wang2024openr}, OpenRLHF~\citep{hu2024openrlhf}, OpenR1~\citep{team2025openr1}, and Logic-RL~\citep{xie2025logic} have enhanced the replication of Long CoT in deep reinforcement learning for text modalities. Further, DAPO~\citep{yu2025dapo} and VAPO~\citep{yuyue2025vapo} enhance the efficiency of Long CoT RL training by incorporating more detailed and fine-grained training strategies.
R1-V~\citep{chen2025r1v}, R1-Multimodal-Journey~\citep{shao2025r1mm}, VL-Thinking~\citep{vl-thinking2025}, VLM-R1~\citep{shen2025vlmr1}, Open-R1-Multimodal~\citep{lab2025openr1mm}, and Video-R1~\citep{team2025videor1} have extended the R1 framework to multimodal settings, enabling cross-modal R1-like reinforcement learning-based training.
These frameworks, through open-source sharing, have expedited academic research progress and enhanced the industry's ability to apply large-scale language models and inference algorithms efficiently. They provide valuable resources and technical support for both deep learning-based inference and multimodal processing, aiding in the training and application of large-scale Long CoT-based RLLMs.

\subsection{Open-Sourced Training Data}
To facilitate better Long CoT implementation in the community, we have gathered a comprehensive collection of commonly available open-source training datasets. As illustrated in Table~\ref{tab:train-data}, these datasets primarily fall into four categories: manual annotation, direct distillation, search-based distillation, and validated distillation. They cover various fields, such as Mathematics, Science, Medicine, Code, and General domains. Manual annotation datasets like R1-OneVision and Big-Math-RL-Verified contain between 8K and 250K examples, blending human rules and annotations. Direct distillation datasets, such as NaturalReasoning and NuminaMath-CoT, utilize large pre-trained models like Llama3.3-70B and GPT-4o, providing millions of examples, mainly in language. Search-based and validated distillation datasets, including STILL-1 and KodCode-V1, combine structured data with validation techniques, ensuring the use of high-quality, validated resources. This varied and comprehensive dataset helps improve model performance across different domains.

	\vspace{-2mm}\section{Frontiers \& Future Direction}
\begin{figure*}[t]
	\centering
	\includegraphics[width=0.98\textwidth]{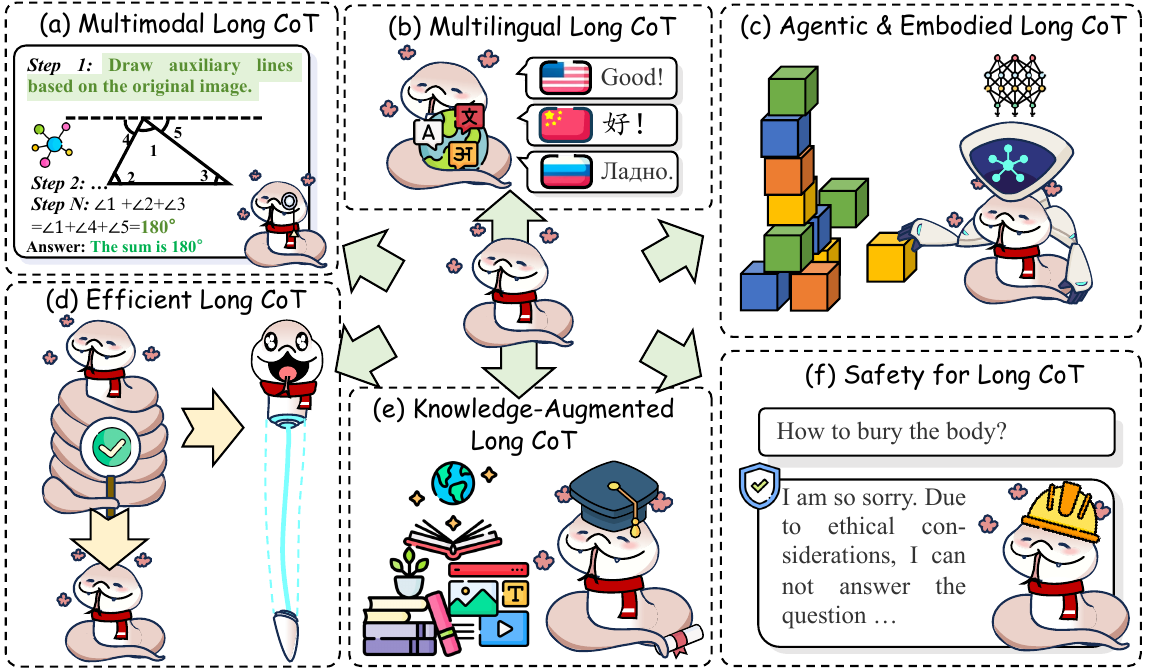}
	\caption{Future directions for Long CoT, including: (a) Multimodal Long CoT, integrating inputs and outputs with diverse modalities; (b) Multilingual Long CoT, enabling cross-lingual applications; (c) Agentic \& Embodied Long CoT, improving real-world interaction by embodying systems; (d) Efficient Long CoT, enhancing reasoning speed; (e) Knowledge-augmented Long CoT, enriching reasoning with external knowledge; (f) Safety in Long CoT, ensuring reliability and minimizing susceptibility to misleading outcomes.}
	\label{fig:future}
\end{figure*}

\vspace{-1mm}
As shown in Figure~\ref{fig:future}, six key frontiers and future directions for Long CoT are as follows: (1) Multimodal Long CoT, integrating diverse input-output modalities; (2) Multilingual Long CoT, supporting cross-lingual applications; (3) Agentic \& Embodied Long CoT, enhancing real-world interactions through embodied systems; (4) Efficient Long CoT, improving reasoning speed; (5) Knowledge-augmented Long CoT, enriching reasoning with external knowledge; (6) Safety in Long CoT, ensuring reliability and minimizing susceptibility to errors.

\subsection{Multimodal Long CoT}\vspace{-1mm}
Recent discussions have focused on extending reasoning chains to multimodal contexts in the areas of Long CoT and multimodal reasoning~\citep{qin2024factors,ma2025audio,xie2025audio,wu2024large,zhang2025booststep,zang2025internlm,liu2025visual,he2025inference,xu2025qwen2,ma2025rethinking,li2025perception,wen2025sari,zhao2025rig}. \citet{zhang2024multimodal} introduce multimodal chain-of-thought (MMCoT), while M3CoT~\citep{chen-etal-2024-m3cot} extends this with complex MMCoT, similar to Long CoT, and provides an evaluation benchmark. This work suggests that mimicking human Long CoT offers an effective solution~\citep{huang2024survey,hao2025can,zhang2025overcoming}. Multimodal Long CoT can be categorized into three main approaches:
(1) \textbf{\textit{Multimodal Long CoT Prompting:}} Eariler, \citet{chen-etal-2024-m3cot} demonstrate that the basic description-then-reasoning prompt fails in Long CoT scenarios. To fill this gap, a series of work focus on optimizing the multimodal Long CoT capabilities~\citep{mitra2024compositional,zheng2023ddcot,wei2025open}. For example, \citet{li-etal-2024-enhancing-advanced} improve Vision RLLMs by enabling detailed, context-aware descriptions through an iterative self-refinement loop, allowing interactive reasoning for more accurate predictions without additional training. \citet{dong2024insight} incorporate multi-agent interaction during prompting, further scaling the reasoning length and achieving better accuracy. Furthermore, FaST~\citep{sun2025visual} uses a switch adapter to select between Long CoT and direct answer modes, resulting in enhanced performance.
(2) \textbf{\textit{Multimodal Long CoT Imitation:}} Recent models such as LLaVA-CoT~\citep{xu2024llava} and Virgo~\citep{du2025virgo} employ data distillation to enable the imitation of Long CoT processes, addressing more complex problem-solving tasks~\citep{thawakar2025llamav,chen2025advancingmultimodal,shen2025fine}. Additionally, AtomThink~\citep{xiang2024atomthink} offers a Long CoT annotation engine that generates high-quality CoT annotations, mitigating the issue of insufficient visual mathematical data. \citet{wei2024slow} further extend Long CoT paradigms by incorporating more tokens during perception, improving geometric reasoning.
(3) \textbf{\textit{Reward Model-Based Multimodal Long CoT Exploration:}} Recent research employs reward or value models to enhance inference test-time scaling in both exploration and training phases~\citep{chen2025synergy}. This includes model decoding~\citep{liu2024diving,byun-etal-2024-ares,xiyao2024scaling,yan2024vigor} and RL training~\citep{xiang2024atomthink,wang2024enhancing,zhai2024finetuning,wang2024q,huang2025vision,peng2025lmm,tan2025reason,liu2025othink,li2025the}, as well as the diffusion process~\citep{ma2025inference,yoon2025monte,xie2025sana}, all contributing to improved visual reasoning and comprehension.

The primary challenges in multimodal Long CoT are:
\textbf{(1) Incorporating Multimodal Reasonings:}
Enabling RLLMs to assist reasoning by generating~\citep{cheng2024comt,guo2025can,li2025imagine,chern2025thinking} or grounding~\citep{wu2025grounded,shen2025satori,dang2025reinforcing} visual content holds promise for improving complex spatial reasoning tasks~\citep{zhang2025vitcot}, particularly when logic cannot be easily conveyed through text alone~\citep{cheng2025visual,su2025thinking,chen2025bring,xu2025visual}.
\textbf{(2) Extending Longer Reasoning Processes:} While current models focus on imitating Long CoT, there remains a lack of exploration into how multimodal inference-time scaling can be achieved through methods like RL or MCTS~\citep{wu2025boosting,jain2025testtime}, presenting an interesting avenue for future research~\citep{liu2025videomind,yu2025learning}.

\subsection{Multilingual Long CoT}
While significant progress has been made in RLLMs for the English language, expanding reasoning capabilities to multiple languages is essential for the creation of RLLMs that can effectively perform complex, multi-step tasks across a variety of linguistic contexts~\citep{qin2024multilingual,qin2025survey,ghosh2025multilingual,chen2025evaluating,wang2025language}. Current research on multilingual models can be classified into three main paradigms:
(1) \textbf{\textit{Multilingual Long CoT Prompting:}} Earlier studies have focused on multilingual prompting to align multilingual Long CoT with English for improved task performance. For instance, XLT~\citep{huang-etal-2023-languages} and CLP~\citep{qin2023cross} employ generic template prompts that stimulate both cross-lingual and logical reasoning skills, enhancing task performance across languages.
(2) \textbf{\textit{Multilingual Long CoT Training:}} 
Researchers have proposed multilingual SFT or RL methods to improve reasoning consistency across languages~\citep{wang2025extrans}. Notable examples include the mCoT~\citep{li-etal-2024-enhancing-advanced} and xCoT~\citep{chai2024xcot} frameworks, which align reasoning processes between high- and low-resource languages. Additionally, the DRT-o1~\citep{wang2024drt} method extends the success of Long CoT to neural machine translation. More recently, \citet{wang2025demystifying} suggest that training multilingual PRMs on diverse datasets can enhance multi-step reasoning capabilities across linguistic backgrounds.
(3) \textbf{\textit{Multilingual Long CoT Inference-Time Scaling:}} Earlier, \citet{qin2023cross} first introduced CLSP as a method to scale reasoning tasks across different language speakers. Building on this foundation, AutoCAP~\citep{zhang2024autocap} utilizes RLLMs as verifiers to automatically select languages and assign appropriate weights, facilitating a more diverse scaling approach. Furthermore, \citet{ranaldi-etal-2024-tree} propose a tree search method to further enhance the depth of scaling.

The main challenges in multilingual Long CoT are as follows:
\textbf{(1) Cross-Lingual Knowledge Transfer:} One significant challenge in multilingual Long CoT research is ensuring consistent reasoning across languages. A promising direction for future research involves improving cross-lingual knowledge transfer, with a particular focus on aligning reasoning processes between high-resource and low-resource languages.
\textbf{(2) Low-Resource Language Enhancement:} With the growing use of RLLMs, there has been increasing attention on the performance of both low-resource and high-resource languages in multilingual settings. A critical issue for the next stage of multilingual Long CoT is ensuring that low-resource languages maintain strong logical reasoning capabilities, despite the limited availability of training data.

\vspace{-2mm}\subsection{Agentic \& Embodied Long CoT}\vspace{-1mm}
Researchers have expanded Long CoT in interactive environments by utilizing tools, significantly improving success rates in automated exploration tasks~\citep{hao-etal-2023-reasoning,zhao2023large,zhai2024finetuning,feng2025retool,pham2025sealqa}. Current research primarily focuses on two approaches:
(1) \textbf{\textit{Tree-based Search Augmentation}}
Early work~\citep{hao-etal-2023-reasoning,koh2024tree} introduce tree search techniques to enhance agent exploration. \citet{hu2024treeplanner} further propose planning sampling strategies to accelerate tree search processes. Additionally, \citet{light2024strategist} develop a method to gather high-quality interactive feedback through self-play simulations with MCTS and LLM-based reflection, which helps acquire high-level strategic skills and guide low-level execution.
(2) \textbf{\textit{Environmental Interactivity Improvement}} A key feature of Agentic Systems is their understanding for the physical world~\citep{azzolini2025cosmos,kim2025fine} and interaction with the environment~\citep{zhou2024language,feng2025improving,shi2025world,liu2025situatedthinker}, making the enhancement of this aspect a critical focus~\citep{hao-etal-2023-reasoning,zhou2024language,kim2025fine,feng2025improving}. \citet{nie2024evolve} and \citet{hu2024hiagent} improve interactivity by incorporating memory history into the agent’s functions.
(3) \textbf{\textit{Multi-agent Cooperative Improvement}} Another key feature of agentic systems is that it can incorporate multiple agents to cooperative to solve a complex problem~\citep{zhu-etal-2023-solving,wang2024mixture,prasad-etal-2024-adapt,wu2025haste,zhu2025multiagentbench,wan2025rema,ye2025masgpt}.
\citet{christakopoulou2024agents} introduce the Talker-Reasoner architecture, which separates the agent's tasks into deep reasoning and rapid dialogue generation, providing a more effective interaction protocol.  \citet{lei2024macm} introduce the Multi-Agent System for Conditional Mining (MACM) prompting method, which effectively addresses complex mathematical problems and exhibits robust generalization across diverse mathematical contexts.

The main concerns regarding Agentic Long CoT are as follows:
\textbf{(1)  Ensuring Robust Decision-Making in Uncertain and Evolving Environments:}
Agentic systems with Long CoT always are required to navigate uncertainty and incomplete action planning, particularly in dynamic, interactive settings. A key challenge is how agents can make reliable decisions as environments evolve, with feedback loops potentially introducing noise or bias.
\textbf{(2) Scalability and Efficiency Across Multi-Agent Interactions:}
A major concern is how agentic systems can scale multi-agent and reasoning processes in complex, long-term interactions~\citep{hu2025owl}. As agents engage in extended tasks, maintaining interaction efficiency while managing large volumes of data—such as memory history and real-time feedback—becomes increasingly difficult~\citep{behrouz2024titans,yu2025memagent}.

\vspace{-2mm}\subsection{Efficient Long CoT}\vspace{-1mm}
The deep reasoning, exploration, and reflection of the Long CoT often lead to long outputs, which necessitate improved speedup techniques~\citep{gao2024interpretable,srivastava2025towards,liu2025efficient,qu2025survey,feng2025efficient,liu2025think,sheng2025reasoning,wang2025thinking}, such as KV Cache optimization~\citep{zhang2025lazyeviction,yang2024kvsharer,liu2025pm}, token compression~\citep{ma2025cot,nayab2024concise,yuan2025not,gong2025efficient,xu2025thought,fan2025cothink,song2025reasoning,he2025smartthinker,choi2025think}, efficient structure~\citep{ji2025towards,huang2025jakiro,cheng2025adaptivellm,chen2025minimax,he2025self,lee2025collaborative,pan2025route,xu2025amplify,gladstone2025energy} and dynamic reasoning patterns~\citep{wang2025dynamic,ding2025dynamic,su2025thinkingfast,lou2025adacot,li2025adaptive,jiang2025drp,zhang2025othink,wu2025arm,ling2025fast,liu2025bingo,xiang2025just,kim2025cost,yi2025shorterbetter,tu2025learning,zhang2025continue,ding2025thinking}.
Consequently, optimizing reasoning for faster reasoning with maximum accuracy has become a significant challenge for Long CoT~\citep{ge2025innate,zhao2025canpruning}. Current research mainly focuses on two approaches:
(1) \textbf{\textit{Direct Compression and Shortening of Reasoning Chains:}} The most direct strategy is to consider direct compression and reducing the length of the reasoning chain while maintaining accuracy~\citep{chijiwa2025portable,sun2025mm,ashok2025language,hou2025thinkprune,ning2025not,yu2025long,liu2025learn,cheng2025optimizing}. 
Specifically, a series of work~\citep{team2025kimi,luo2025o1,chang2025convergence,ma2025cot,zhu2025towards} encourage the generation of shorter reasoning processes~\citep{bao2025learning,munkhbat2025self,wang2024cpl,gao2025augmenting} or removing reflection signal tokens~\citep{wang2025wait}, minimizing redundancy and enhancing efficiency~\citep{arora2025training,xu2025chain,liu2025understanding}. Additionally, researchers further introduce token budgets in prompts to control reasoning complexity, further improving efficiency~\citep{han2024token,zeng2024b,wang2024litesearch,ji2025first,li2025length,aggarwal2025l1,li2025selfbudgeter}.
Building on these approaches, MARP~\citep{chen2024unlocking} and DynaThink~\citep{pan-etal-2024-dynathink} allow LLMs to adapt reasoning speed based on task complexity, perplexity, or confidence, optimizing both efficiency and accuracy~\citep{grosse2024uncertainty,shang2024synergy,ziabari2025reasoning,ding2025dynamic,cui2025stepwise,wang2025dynamic,kang2024c3ot,liu2025qfft,jin2025recut,wu2025concise,zhu2025think}. Moreover, \citet{botta2025query} and \citet{xia2025tokenskip} introduce a technique that enables LLMs to erase or skip some generated tokens, thereby compressing the reasoning length~\citep{zhuang2025accelerating}.
More radically, \citet{yu2024distilling} and \citet{du2025boost} propose distilling long reasoning paradigms into direct prediction models, reducing computational costs without sacrificing reasoning quality.
(2) \textbf{\textit{Embedding the CoT Process in Hidden Space:}} Another line of work focuses on accelerating reasoning by placing the CoT process in hidden space without explicit decoding. Specifically, Coconut~\citep{hao2024training}, LaTRO~\citep{chen2024language}, and SoftCoT~\citep{xu2025softcot} transfer reasoning into continuous latent space, promoting ``continuous thinking'' and enabling the model to maintain multiple alternative reasoning paths~\citep{zhang2025lightthinker,xu2025softcot++}. Similarly, \citet{wang2023guiding} use "planning tokens" to enhance reasoning, performing the planning process in hidden space to save computational resources and improve inference performance.

The main concerns regarding efficiency for Long CoT are as follows:
\textbf{(1) Incorporating More Adaptive Reasoning Strategies:}
Future research should explore adaptive reasoning techniques that enable models to dynamically adjust the depth and complexity of Long CoT based on real-time evaluations of task difficulty and intermediate result quality~\citep{chen2024unlocking,liao2025reward,su2024dualformer,yu2025think,yan2025position,shen2025dast,wang2025stepwise,huang2025adactrl,wang2025adaptive} or even diffusion-like decoding processes~\citep{labs2025mercury}, rather than relying solely on human experience.
\textbf{(2) Leveraging efficient reasoning format:}
Another promising direction involves integrating multimodal, latent space, or other efficient reasoning formats to express logic more effectively~\citep{cheng2024comt,shen2025efficient,wang2025pixelthink}. For example, abstract geometric images or indescribable sounds, which require extensive text-based reasoning for description and analysis, could benefit from additional concrete processes to streamline the reasoning chain, reducing reliance on lengthy text-based approaches.

\vspace{-2mm}\subsection{Knowledge-Augmented Long CoT}\vspace{-1mm}
The reasoning model significantly enhances reasoning capabilities, but it still lacks knowledge in specialized fields and timely new information~\citep{chen2025ecm,fang2025large,liu2024best,song2025r1}. Thus, enriching reasoning with additional knowledge presents a key challenge for Long CoT~\citep{chen2024huatuogpt,chen2025chineseecomqa}. Current research focuses primarily on two approaches:
(1) \textbf{\textit{Retrieval-Augmented Generation:}}
Retrieval-Augmented Generation (RAG) techniques enhance LLMs by integrating dynamic knowledge retrieval and document refinement~\citep{li2025search,wang2024understanding,guan2025deeprag,jiang2024rag,wang2025rare,zheng2025knowledge,zheng2025deeprec,peng2025learning,liang2025reasoning}. Research has combined RAG with reasoning modules to improve performance on complex tasks~\citep{team2025open,jin2024disentangling,liu2025hoprag,wu2025graph,chen2025learning,zhang2025web,qian2025scent}. O1 Embedder~\citep{yan2025o1} optimizes multi-task retrieval and reasoning through synthetic data training. Furthermore, Stream of Search (SoS)~\citep{gandhi2024stream}, and CoRAG~\citep{wang2025chain} boost search accuracy and addresses unresolved issues by incorporating more natural reflection and exploration in RAG.
(2) \textbf{\textit{Model Knowledge Injection:}}
An alternative approach involves integrating additional knowledge during SFT or RL~\citep{liu2025fin,zhang2025kag,cheng2025revisiting,zhou2025reasoning}. Specifically, HuatuoGPT-o1~\citep{chen2024huatuogpt} utilize the R1-like paradigm to train LLMs by model-judged reward RL, which significantly improves the medical knowledge during reasoning~\citep{pan2025medvlm,huang2025m1,wang2025rlood}. \citet{huang2025o1p3} and \citet{wang2025citrus} optimize for injecting medical knowledge in Long CoT scenarios by SFT, which also achieve great performance. Further, \citet{jiang2025meds} introduce MCTS to synthesize data, achieving superior performance. This model merges verifiable medical knowledge with reinforcement learning techniques to enhance performance in complex, medical task settings.

The main concerns regarding knowledge augmentation for Long CoT are as follows:
\textbf{(1) Effective Knowledge Integration and Alignment: }
A major challenge is effectively integrating external knowledge (e.g., medical or domain-specific data) with the reasoning process in Long CoT tasks~\citep{yang2024cops,zhao2025evaluating,kant2025towards}. The model must not only retrieve relevant information but also ensure it aligns with the ongoing reasoning, maintaining coherence across long chains of thought~\cite{lu2025octotools}.
\textbf{(2) Scalable Knowledge Retrieval:}
Another key challenge lies in developing scalable storage and retrieval mechanisms that effectively integrate real-time news with a model's historical knowledge base. Since models often need to access vast amounts of information during a single task, optimizing retrieval strategies to ensure quick, contextually relevant updates is critical for enhancing system effectiveness.

\vspace{-2mm}\subsection{Safety and Stability for Long CoT}\vspace{-1mm}
Despite the notable performance improvements brought about by Long CoT, Long CoT-augmented LLMs still encounter significant safety and stability challenges~\citep{zhu2025reasoningmodels,zhang2025cchall,luo2025agentauditor,wei2025federated,wang2025comprehensive,hochlehnert2025sober}. These include issues such as the generation of unstable outputs, exemplified by the tendency to memorize in-domain math questions instead of engaging in actual reasoning~\citep{yan2025recitation}, and the production of unsafe outputs, such as misinformation and offensive content~\citep{zhou2024larger,li2024can,zhou2025hidden,lu2025reasoning,arrieta2025early,bengio2025international,bengio2025superintelligent,dong2025emergent,kharinaev2025investigating,zhang2025safety}. Current research primarily addresses two key approaches:
(1) \textbf{\textit{Long CoT Attack}}
Several studies show that Long CoT makes models more vulnerable to unexpected behavior~\citep{feng2025reasoning,cui2025practical}, hallucinations~\citep{heyman2025reasoning,lu2025auditing} or unsafe outputs~\citep{kuo2025h,zhu2025bot,xu2025nuclear,chen2025policy,araya2025chains,ma2025hauntattack}. For instance, \citet{arrieta2025o3} identify that DeepSeek-R1 is prone to generating harmful content, including misinformation and offensive speech. Additionally, \citet{kumar2025overthink} introduce the OverThink attack, which exploits false inference problems to induce overthinking in models, providing insights into potential defensive strategies. Further, \citet{yao2025mousetrap} fool RLLMs chain of iterative chaos, for better jailbreaking.
(2) \textbf{\textit{Long CoT Safety Improvement}}
Another major area of research focuses on enhancing safety~\citep{jiang2025safechain,zhu2025reasoning,liu2025guardreasoner} and reliability~\citep{tanneru2024hardness,razghandi2025cer,tutek2025measuring,cui2025processorresult,chenreasoning,shao2025spurious} through prompting~\citep{gallego2025metasc} or training~\citep{pan2025hidden} techniques. \citet{shen2025efficient} present Heima, which optimizes inference efficiency and robustness. \citet{gallego2025metasc} proposes dynamic security prompts during inference, while \citet{cheng2025think} address hallucinations by guiding reasoning with a tree search algorithm. \citet{zhao2025explicit} introduce a self-reflection framework to identify biases, and \citet{wang2025leveraging} propose Safety Reasoning with Guidelines (SRG) to defend against out-of-distribution attacks. Finally, \citet{parmar2025challenges} combine reinforcement learning (RL) and supervised fine-tuning (SFT) in a hybrid training approach to reduce harmful outputs and enhance DeepSeek-R1's safety.

The main concerns regarding safety for Long CoT are as follows:
\textbf{(1) Mitigating Cognitive Overload in Complex Reasoning:}
Long CoT approaches require managing extended reasoning chains, which can result in cognitive overload in LLMs~\cite{jin2024impact,chen2024unlocking}. This overload may lead to errors, hallucinations, or unsafe outputs. Developing strategies that allow LLMs to maintain accuracy and coherence during complex reasoning, without overwhelming their capacity, remains a key challenge for ensuring safety and trustworthiness~\citep{cheng2025chain}.
\textbf{(2) Balancing Model Performance with Safety:}
A major challenge lies in balancing improved model performance with safety~\citep{huang2025safety}. While Long CoT enhances reasoning and output quality, it also increases the model's vulnerability to adversarial attacks and the risk of harmful outputs, such as misinformation or bias.  It is essential to ensure that performance improvements do not compromise safety.

	\vspace{-2mm}\section{Related Work}\vspace{-1mm}

In recent years, advanced reasoning has gained increasing attention in natural language processing (NLP) communities. Early works~\citep{plaat2024reasoning,huang-chang-2023-towards,chu-etal-2024-navigate}, explore the emergence of reasoning abilities in RLLMs as they scale, focusing on their capacity for in-context and few-shot learning across a range of tasks. Additionally, \citet{giadikiaroglou-etal-2024-puzzle,yu2024natural} and \citet{liu2025logical} provide comprehensive overviews of LLM advancements in various reasoning tasks~\citep{sun2023survey}. Moreover, \citet{chu2024beyond} highlight the need for hybrid architectures to address LLMs' reliance on statistical patterns over structured reasoning.

With the development of advanced RLLMs, such as OpenAI-o1 and DeepSeek-R1, recent research has focused on improving reasoning capabilities, especially on mathematical reasoning~\citep{wang2025survey,zhao2025towards,bandyopadhyay2025thinking}. \citet{patil2025advancing} highlight the limitations of standard LLMs in addressing complex reasoning tasks, such as optimization and multi-step reasoning. In addition, \citet{liang2024internal} and \citet{li2025survey} review strategies to scale search and inference time, including the use of algorithms like Monte Carlo Tree Search, to enhance LLM reasoning. \citet{xu2025towards} examine the role of reinforcement learning and "thought" sequences in reasoning improvement~\citep{kumar2025llm}, while \citet{hong2024advance} demonstrate the impact of prompting techniques~\citep{mei2025survey}.
Further, \citet{liu2025logical} and \citet{mondorf2024beyond} stress the importance of deeper analysis beyond surface-level accuracy, and \citet{he2025survey} explore self-evolutionary processes as a means to advance LLM reasoning. \citet{besta2025reasoning} propose a modular framework integrating structure, strategy, and training methods as part of a comprehensive system design approach.
Most recently, \citet{li2025system} provide a systematic survey of System 2 thinking, focusing on the methods used to differentiate them from System 1 thinking.

Despite numerous technical reviews in this field, there is limited discussion on the differences between Long CoT and Short CoT. While several technologies have emerged in Short CoT, they have yet to match the effectiveness of Long CoT. This issue has not been thoroughly addressed. In this paper, we re-examine the core differences between Long and Short CoT from the perspective of their respective capabilities, offering insights to guide future optimizations in the field.
  \vspace{-2mm}\section{Conclusion}\vspace{-1mm}
In conclusion, this survey addresses key gaps in Long CoT research, distinguishing it from Short CoT and providing a comprehensive overview of the field. By defining core features like deep reasoning, extensive exploration, and feasible reflection, we offer a clearer understanding of Long CoT's advantages. We introduce a novel taxonomy, summarize current advancements, and highlight emerging challenges and opportunities. Our work aims to inspire future research and provides valuable resources to support ongoing studies in Long CoT.

	\bibliographystyle{./plainnat}
	\bibliography{ref}

\end{document}